\documentclass{article}

 \usepackage[main, final]{neurips_2026}

\usepackage[utf8]{inputenc} 
\usepackage[T1]{fontenc}    
\usepackage[colorlinks=true,linkcolor=purple,anchorcolor=black,citecolor=blue,filecolor=black,menucolor=black,runcolor=black,urlcolor=blue]{hyperref}
\usepackage[most]{tcolorbox}
\usepackage{url}            
\usepackage{booktabs}       
\usepackage{amsfonts}       
\usepackage{nicefrac}       
\usepackage{microtype}      
\usepackage{amsmath}
\usepackage{amssymb}
\usepackage{mathtools}
\usepackage{amsthm}
\usepackage{caption}
\usepackage{bbm}
\usepackage{tikz}
\usetikzlibrary{shadows, calc, positioning}
\usepackage{pifont}
\usepackage{url}
\usepackage{afterpage}
\usepackage{adjustbox}
\usepackage{fnpct}
\usepackage{listings}
\usepackage{placeins}
\usepackage{nicefrac}
\usepackage{xurl}
\usepackage{enumitem}
\usepackage{tabularx}
\usepackage{nicefrac}
\usepackage{makecell}
\usepackage{xspace}
\usepackage{graphbox}
\usepackage{pifont}
\usepackage{enumitem}
\usepackage{colortbl}
\usepackage{multirow}
\usepackage{diagbox}
\usepackage{array}
\usepackage{wrapfig}
\usepackage{subcaption}
\usepackage{color}
\usepackage{stfloats}
\usepackage{diagbox}
\usepackage{afterpage}
\usepackage{etoolbox}
\usepackage{float}
\usepackage[capitalize,noabbrev]{cleveref}
\usepackage[dvipsnames]{xcolor}

\usepackage{booktabs}
\usepackage{multirow}

\newcommand{\norm}[1]{\left\|#1\right\|}
\newtheorem{lemma}{Lemma} 

\def\L{\mathcal{L}}

\def\min{\mathop{\rm min}\nolimits}
\def\max{\mathop{\rm max}\nolimits}

\definecolor{Gray}{gray}{0.85}
\definecolor{NewGray}{gray}{.45} 
\definecolor{BlueGray}{rgb}{0.92, 0.92, 1}
\definecolor{RedGray}{rgb}{0.98, 0.9, 0.78}
    \definecolor{LightCyan}{rgb}{0.88,1,1}
\definecolor{lightgreen}{rgb}{0.8, 0.99, 0.85}
\definecolor{DarkGreen}{rgb}{.1, .75, .1}
\definecolor{DarkRed}{rgb}{.95, .0, .1}
\definecolor{GrayGreen}{rgb}{0.8,0.92,0.8}
\definecolor{GrayRed}{rgb}{0.85,0.7,0.7}
\definecolor{bluey}{rgb}{0.65,0.75,0.95}
\definecolor{lightblue}{rgb}{0.7,0.9,1.0}
\definecolor{lowcolor}{rgb}{0.9,0.9,1.0}
\definecolor{highcolor}{rgb}{0.8,0.9,1.0}
\definecolor{beige}{rgb}{1,0.95,0.85}
\definecolor{lightorange}{RGB}{255, 224, 178}

\usepackage{xcolor}

\definecolor{lightgreen}{RGB}{220, 255, 220}
\definecolor{lightred}{RGB}{255, 223, 223}
\definecolor{lightgray}{RGB}{240, 240, 240}
\definecolor{lightblue}{RGB}{220, 230, 250}

\newcommand{\gbox}[2]{\colorbox{lightgreen}{\parbox{#1}{#2}}}
\newcommand{\rbox}[2]{\colorbox{lightred}{\parbox{#1}{#2}}}
\newcommand{\bbox}[2]{\colorbox{lightblue}{\parbox{#1}{#2}}}
\newcommand{\graybox}[2]{\colorbox{lightgray}{\parbox{#1}{#2}}}
\newcommand{\obox}[2]{\colorbox{lightorange}{\parbox{#1}{#2}}}

\usepackage{xcolor,pifont}
\newcommand*\colourcheck[1]{%
  \expandafter\newcommand\csname #1check\endcsname{\textcolor{#1}{\ding{52}}}%
}
\newcommand*\colourcross[1]{%
  \expandafter\newcommand\csname #1cross\endcsname{\textcolor{#1}{\ding{56}}}%
}
\colourcheck{blue}
\colourcheck{DarkGreen}
\colourcross{DarkRed}

\newcommand{\tofu}{TOFU\xspace}

\newcommand{\rouge}{ROUGE\xspace}

\newcommand{\llmj}{LLM-Judge\xspace}
\newcommand{\mistral}{Mistral-7B\xspace}
\newcommand{\phipara}{Phi-3.5-mini-instruct\xspace}
\newcommand{\phirw}{Phi-3 Mini-4K-Instruct (3.8B)\xspace}
\newcommand{\qwen}{Qwen2.5-3B-Instruct\xspace}
\newcommand{\llama}{Llama-3.2-3B-Instruct\xspace}
\newcommand{\gemini}{Gemini-2.5-Flash}
\newcommand{\rwku}{RWKU\xspace}
\newcommand{\lpf}{LPF\xspace}
\newcommand{\ours}{JensUn\xspace}
\newcommand{\jsdiv}{JSD\xspace}
\newcommand{\ga}{GradAscent\xspace}
\newcommand{\gd}{GradDiff\xspace}
\newcommand{\simnpo}{SimNPO\xspace}
\newcommand{\ourfloss}{$\mathcal{L}_{\mathcal{F}}^{\text{\jsdiv}}$\xspace}

\usepackage{pifont}
\newcommand{\RNum}[1]{\uppercase\expandafter{\romannumeral #1\relax}}

\newcolumntype{C}[1]{>{\centering\arraybackslash}p{#1}}
\newcolumntype{L}[1]{>{\raggedright\arraybackslash}p{#1}}
\newcolumntype{R}[1]{>{\raggedleft\arraybackslash}p{#1}}
\newlength\newl
\newlength\newlc

\newlength\colwidth

\makeatletter
\def\blfootnote{\xdef\@thefnmark{}\@footnotetext}
\makeatother

\newcommand{\mh}[1]{\textcolor{black}{#1}}
\newcommand{\mhb}[1]{\textcolor{black}{#1}}
\newcommand{\mhc}[1]{\textcolor{black}{#1}}
\newcommand{\nds}[1]{{\color{black}#1}}
\newcommand{\maxm}[1]{\textcolor{black}{#1}}
\newcommand{\ap}[1]{\textcolor{black}{#1}}

\newcommand{\partialrowcolorredsm}[4]{
  \cellcolor{lightred} #1 & \cellcolor{lightred} #2 & \cellcolor{lightred} #3 & \cellcolor{lightred} #4
}
\newcommand{\partialrowcolorbluesm}[4]{%
  \cellcolor{lightblue} #1 & \cellcolor{lightblue} #2 & \cellcolor{lightblue} #3 & \cellcolor{lightblue} #4
}

\title{Unlearning That Lasts: Utility-Preserving, Robust, and Almost Irreversible Forgetting in LLMs}

\author{%
  Naman Deep Singh$^{1,2}$\thanks{Correspondence: \href{mailto:naman-deep.singh@uni-tuebingen.de}{\textit{naman-deep.singh@uni-tuebingen.de}}} \And Maximilian Müller$^{1,2}$ \And Amit Peleg$^{1,2}$ \AND Francesco Croce$^{3,4}$ \And Matthias Hein$^{1,2}$ \AND 
  \normalfont{T\"ubingen AI Center$^{1}$} \hspace{3pt} \normalfont{University of T\"ubingen$^{2}$} \hspace{3pt} \normalfont{ELLIS Institute Finland$^{3}$} \hspace{3pt} \normalfont{Aalto University$^{4}$} \\
}

\begin{document}

\maketitle

\begin{abstract}
Unlearning in large language models (LLMs) requires precisely removing specific information from a pre-trained model, e.g. to delete private data or harmful knowledge acquired during pre-training. However, existing unlearning methods often fall short under thorough evaluation. This paper makes three contributions. First, we introduce JensUn, a novel unlearning approach which leverages the Jensen-Shannon Divergence (JSD) as the training objective for both forget and retain sets.
 We provide theoretical arguments why JSD outperforms KL-Divergence, and show in extensive experiments that JensUn achieves the best forget-utility trade-off and demonstrates strong resilience to benign relearning, outperforming existing methods.
Second, for a precise unlearning evaluation, we introduce LPF, a curated dataset of less prominent facts that provides a realistic unlearning scenario. Third, to comprehensively test unlearning methods, we 
\textit{(i)} employ an LLM as semantic judge instead of the standard \rouge score, and \textit{(ii)} evaluate worst-case unlearning over various paraphrases and input formats.
Our improved evaluation reveals that many unlearning methods are less effective than previously thought.
\end{abstract}

\section{Introduction}

Training large language models (LLMs) on massive data scraped from the internet yields impressive performance but comes with serious safety concerns, including the risk of exposing private information~\citep{nasr2023scalable}, violating copyrights~\citep{jang2023knowledge, karamolegkou2023copyright}, and amplifying harmful content~\citep{huang2024position, lu2022quark, barrett2023identifying, wen2023unveiling}. 
To prevent acquisition of undesired knowledge, one could selectively remove or adjust problematic samples in the training data and then re-train LLMs from scratch.
Since this \mh{is infeasible in practice}, recent works have explored more efficient alternatives, such as model editing and machine unlearning.
In contrast to re-training, these approaches aim to update a pre-trained LLM to change or remove the internal knowledge encoded in its parameters.
While model editing is used to update the model for a specific piece of existing information~\citep{meng2022locating, ilharcoediting}, \textit{machine unlearning} aims to remove entire concepts from the model~\citep{liu2025rethinking}, like dangerous information~\citep{li2024wmdp,  barrett2023identifying} and sensitive personal data~\citep{nasr2023scalable}, or tries to make the model adhere to the right to be forgotten~\citep{zhang2024right}.
Given its practical relevance in these high-stakes scenarios, many approaches to machine unlearning have appeared~\citep{jang2023knowledge, rafailov2023direct, fan2024simplicity, li2024wmdp}.
However, evaluating their effectiveness is a delicate task, since it has to be determined
if the relevant information has been truly forgotten, or if the model simply suppresses it at a superficial level without actually removing it  \citep{hu2024jogging, thaker2024position,Wang25effectiveEval} and it can be re-introduced by fine-tuning on new data~\citep{hu2024jogging}.

\begin{figure*}
    \centering
    \includegraphics[width=0.98\linewidth]{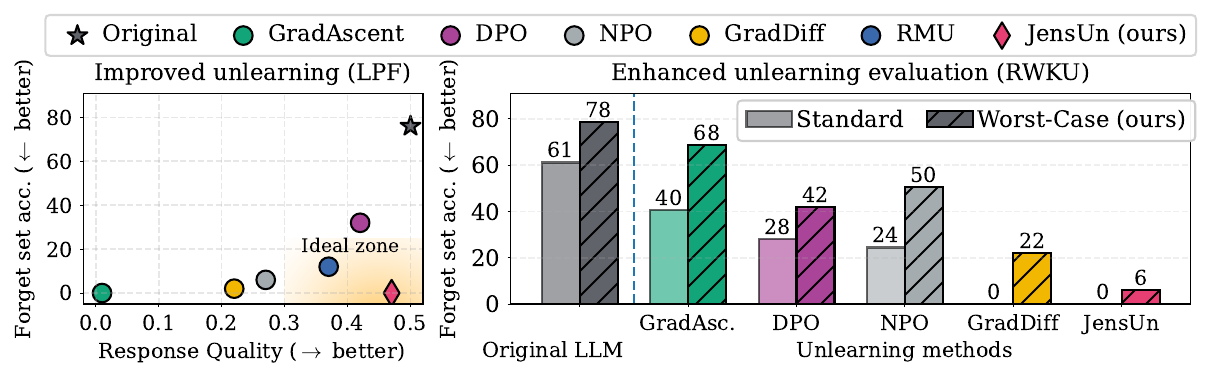} 
    \caption{\textbf{Our \ours yields the best trade-off between unlearning quality (forget set accuracy) and utility of the LLM.
    }
\textit{(left)} Our unlearning method \ours achieves on our \lpf dataset an optimal worst-case forget set accuracy of 0\%  while maintaining high response quality (win rate of the unlearned vs the original model on AlpacaEval, see Section \ref{sec:improve-utility-evaluation}), the most similar to the original \llama pre-trained model. \textit{(right)} Our novel worst-case evaluation using \mhc{9} paraphrases of the query on \rwku reveals that using single question-answer evaluations overestimates unlearning quality: our worst-case evaluation drastically increases forget set accuracy for the fine-tuned LLMs across different unlearning methods as well as the original model (\phirw). The code and the associated LPF dataset can be found at \url{https://github.com/nmndeep/JensUn-Unlearning}.}
    \label{fig:teaser}
\end{figure*}

In this work, we propose a \textit{new unlearning method based on the Jensen-Shannon Divergence}, termed \textbf{\ours}. LLMs unlearned with \ours demonstrate a better forget-utility trade-off than the state-of-the-art baselines (see left plot in~\cref{fig:teaser}). In fact, our models attain the best unlearning quality (under our proposed strong worst-case evaluation) while preserving the highest utility on average across different utility metrics, LLMs, and unlearning datasets. To further test whether the results persist at a larger scale, we also test the unlearning performance on 20 targets simultaneously from the \rwku dataset. Moreover, 
\ours is the only method that is robust to
\textit{relearning}~\citep{lucki2024adversarial, hu2024jogging}. After 400 unlearning steps, \ours shows little recovery of the initially forgotten information when fine-tuned on disjoint topics or on the retain set used for unlearning. We extend the relearning study beyond disjoint topics and the retain set. We give the attacker partial access to the forget set and also test whether quantization~\citep{zhang2025quantization} exposes forgotten knowledge. Under both attacks, \ours reveals little or no held-out knowledge.

Furthermore, we also critically examine 
current unlearning evaluation protocols. We show that \rouge scores \citep{lin-2004-rouge}, commonly used to measure unlearning quality in popular benchmarks \citep{mainitofu, shi2024muse, jinrwku}, may fail to measure the correctness of answers to factual questions (
\cref{fig:rogue-ex}
). 
To address this, we \textit{replace \rouge with capable LLMs as semantic judges}.
In contrast to the ROUGE score, those have high agreement with human judges. 
Moreover, we evaluate the robustness of unlearning towards query variations using paraphrased versions of the queries from the forget set.
Following \cite{thaker2024position}, we also augment each query with in-context samples from a set of non-unlearnt questions.
We argue that one should report the \textit{worst-case evaluation over all such variations}:
unlearning is considered successful \textbf{only} if the LLM cannot correctly answer \textbf{any} of the reformulated questions.
To rigorously test removal of factual knowledge,
we additionally collect \textit{a new, high quality unlearning dataset} with non-dichotomous queries, named Less Prominent Facts (\lpf). 
Testing existing unlearning methods (on both \lpf and \rwku \citep{jinrwku}) with our worst-case evaluation reveals significantly lower unlearning quality,
see~\cref{fig:teaser} (right).
%
In summary,
\begin{itemize}[left=0mm, topsep=0mm, parsep=0pt, itemsep=4pt]
\item We propose \textbf{JensUn}, a novel unlearning method using the \textbf{Jensen-Shannon Divergence} (JSD) as training objective. We show theoretically that, unlike established losses such as Kullback-Leibler divergence, JSD yields balanced forget and retain dynamics, leading to smaller changes to the original model and thus better utility preservation.
Extensive experiments \nds{across varied model sizes} show that JensUn achieves a superior forget-utility trade-off and is resilient to relearning.
    \item 
    Building on previous work, we establish a more \textbf{rigorous evaluation framework} that addresses 
    \mh{flaws} in current protocols by replacing \rouge with a capable \textbf{LLM as a semantic judge} and adopting a \textbf{worst-case evaluation} over multiple paraphrases and input formats.
    \item We introduce \textbf{LPF} (Less Prominent Facts), a new, high-quality dataset of non-dichotomous queries curated to provide a more realistic and challenging benchmark for factual unlearning.
\end{itemize}

\section{Related Work}
\label{sec:related_work}

LLM unlearning aims to remove specific information (individual facts or concepts), represented by a forget set, from a pre-trained model while trying to preserve its overall utility leveraging a retain set.

\textbf{Unlearning methods.} 
Several unlearning methods have been proposed in literature.
Gradient Ascent~\citep{jang2023knowledge}, for instance, 
maximizes the cross-entropy loss on the forget set to remove its influence. This simple solution unlearns effectively but makes the resulting LLM unusable on nominal open-ended tasks. Hence, in Gradient Difference (\gd)~\citep{liu2022continual, mainitofu}, the cross entropy loss on the retain set is minimized in addition. 
Methods based on preference optimization like DPO~\citep{rafailov2023direct}, NPO~\citep{zhangnegative} and \simnpo~\citep{fan2024simplicity} are also commonly used for unlearning, as well as simple solutions like Rejection Tuning (RT)~\citep{ishibashi2023knowledge, mainitofu} and In-Context Unlearning (ICU)~\citep{pawelczyk2024context}. 
Similar to the model editing works~\citep{meng2022locating, ilharcoediting}, RMU~\citep{li2024wmdp} tries to work at the internal representation level across select layers for unlearning. A comparison of the loss functions of \ours, NPO, SimNPO, GradAscent, and GradDiff can be found in~\cref{app:loss-comp}. Another line of works uses distillation-based methods to define desired outputs after unlearning: UnDIAL~\citep{dong2025undial} modifies teacher logits before self-distillation, while DUET~\citep{zhong2026duet} distills refusal behavior from a prompt-steered teacher. Wang et al.~\citep{wang2025gradient} use the G-effect to analyze unlearning across examples, steps, and layers. In comparison, our token-level analysis derives closed-form gradient norms for KL and JSD to explain their different forget and retain dynamics.

\textbf{Unlearning benchmarks.} Existing unlearning benchmarks differ in evaluation set sizes, types, and concepts. \tofu~\citep{mainitofu} uses information about fictitious authors, while WHP~\citep{eldan2023s} employs Harry Potter as the topic with question-answer (QA) queries. MUSE~\citep{shi2024muse} utilizes News and Books corpora, assessing unlearning via verbatim completion, QA, and membership inference attacks (MIA)~\citep{murakonda2021quantifying, ye2022enhanced} for privacy. WMDP~\citep{li2024wmdp} focuses on unlearning harmful concepts using multiple choice questions (MCQs). Beyond forget set evaluation, \rwku~\citep{jinrwku} measures LLM abilities including reasoning~\citep{suzgun2023challenging}, truthfulness~\citep{lin2022truthfulqa}, factuality~\citep{joshi2017triviaqa}, repetitiveness~\citep{alpacaeval} and general knowledge~\citep{hendryckstest2021}.
\maxm{While these benchmarks mostly rely on \rouge scores,~\citep{liu2024revisiting} and~\citep{wei2025dollmsreallyforget} used LLMs to judge the effectiveness of unlearning methods,} 
\mh{ as we do in this paper.}


\textbf{Relearning.} LLMs, after unlearning, can revert to their pre-trained state when 
fine-tuned on data disjoint from the forget set~\citep{lucki2024adversarial, hu2024jogging}. This so-called ``benign relearning'' implies information suppression, not eradication, posing a challenge for LLM deployment. While combining unlearning with Sharpness Aware Minimization (SAM)~\citep{foretsharpness}  partially mitigates this phenomenon \citep{fantowards}, 
we identify contexts where relearning still persists. Our \ours unlearning approach (introduced in the next section) demonstrates better resistance to benign relearning than competitors.
Quantization provides a different test of whether knowledge was removed or merely hidden. Zhang et al.~\citep{zhang2025quantization} show that lowering weight precision after unlearning can recover knowledge for several existing methods. We therefore evaluate \ours at 8-bit and 4-bit precision (see~\cref{app:stronger-recovery}).

\section{
Unlearning via Jensen-Shannon Divergence
}



\textbf{Background.}
The goal of LLM unlearning is to delete knowledge about certain facts or concepts given by a  forget set ($\mathcal{D_F}$), while preserving the utility of the LLM, in particular of related but different facts or concepts in a retain set ($\mathcal{D_R}$).
The forget set is given by $\mathcal{D_F} = \{(x, y)_i\}_{i=1}^{N_\mathcal{F}}$, where $N_\mathcal{F}$ is the number of samples and $(x, y)$ can be QA pairs or paragraphs. The objective is to unlearn the ground truth\footnote{
In practice, one might also want to unlearn an ``incorrect'' output of an LLM. 
} $y$ associated with the input $x$. Both $x$ and $y$ are sequences of tokens and we denote by $y_t$ the $t$-th token in sequence $y$ and by $|y|$ its length. 
Most unlearning methods minimize an objective of the form
%
\begin{align}
\mathcal{L}_{\text{unlearning}}(\theta) = \lambda_\mathcal{F} \L_\mathcal{F}(\theta, \mathcal{D_F}) + \lambda_\mathcal{R} \L_\mathcal{R}(\theta, \mathcal{D_R}), \label{eq:unlearn-obj}
\end{align}
where $\theta$ are the model parameters, $\L_\mathcal{F}$ is the forget set loss, $\L_\mathcal{R}$ the retain set loss, and $\lambda_\mathcal{F}$ and $\lambda_\mathcal{R}$ are tunable hyper-parameters.
The unlearning methods discussed in \cref{sec:related_work} 
fit into this framework, and differ in their choice of $\L_\mathcal{F}$, $\L_\mathcal{R}$. Methods like GradAscent, GradDiff, and RMU aim to move away from the output of the original model on the forget set, while Rejection Tuning instead outputs a 
\mh{fixed string} like ``I don't know''. For the first class of methods the output on the forget set is not well-defined and thus the LLM tends to output random tokens.
\mh{Replacing existing information with potential hallucinations of the LLM is undesired in practice. Thus, we use a fixed output string.}
The choice of the loss functions of existing unlearning methods is discussed in~\cref{app:loss-comp}.

\subsection{Unlearning via \ours}


The Jensen-Shannon Divergence (\jsdiv), 
\begin{align*} \text{\jsdiv}(P \parallel Q) = \frac{1}{2} D_{\text{KL}}(P \parallel M) + \frac{1}{2} D_{\text{KL}}(Q \parallel M),
\end{align*}
measures the distance between two distributions $P$ and $Q$,
%
%
where $M = \frac{1}{2}(P + Q)$ and $D_{\text{KL}}$ is the Kullback-Leibler (KL) Divergence. 
Unlike other losses, e.g. the plain KL-divergence, the \jsdiv is bounded and symmetric. \jsdiv-based losses have been shown to be effective for stabilizing training in GANs~\citep{goodfellow2014generative}, training with noisy labels~\citep{englesson2021generalized}, and semantic segmentation~\citep{croce2024towards}. We show below that, due to its properties, \jsdiv is ideal for unlearning.

\textbf{Forget loss.}
For the forget-loss term, we propose minimizing the \jsdiv between the model output and a fixed target string, e.g. 
\mh{we use} ``No idea'', actively trying to replace the model's answer with a refusal target. 
For each input $(x,y) \in \mathcal{D_F}$, we construct a unique refusal target, $y^\textrm{target}$, by repeating the refusal string and truncating it to match the length of the original sequence $|y|$. 
Denoting by $\delta_{y^\textrm{target}_t}$ the one-hot distribution of the token $y^\textrm{target}_t$ over the vocabulary size, the forget loss $\L_\mathcal{F}^{\text{\jsdiv}}(\theta, \mathcal{D_F})$ is defined as
\begin{align}
    \label{eq:our-forget}
    \L_\mathcal{F}^{\text{\jsdiv}} \!= \!
    \frac{1}{N_\mathcal{F}} \sum_{(x,y) \in \mathcal{D}_F}
 \sum_{t=1}^{|y^{\text{target}}|}\! \text{\jsdiv}\left( p_{\theta}(\cdot | x, y^{\text{target}}_{< t}) \!  \parallel\! \delta_{y^\textrm{target}_t}\right).
\end{align}
\textbf{Retain loss.}
For the retain set $\mathcal{D_R} = \{(x, y)_i\}_{i=1}^{N_\mathcal{R}}$
, the unlearnt model should yield the same output distribution as the base model parameterized by $\theta_\textrm{ref}$.
Thus, we minimize the \jsdiv of these two distributions,
\begin{align}
\!\!\!\L_\mathcal{R}^{\text{\jsdiv}} \!=\! \frac{1}{N_\mathcal{R}} \!\sum_{(x,y) \in \mathcal{D}_R}\sum_{t = 1}^{|y|} \text{\jsdiv}\big( p_{\theta}(\cdot | x, y_{< t})\! \parallel \!p_{\theta_\textrm{ref}} (\cdot| x, y_{< t})\big).\!
\end{align}
Unlike other methods, we do not learn the retain answers but only aim to reproduce the original model's output. The overall \ours objective is the \ap{weighted} sum of forget and retain losses as in \eqref{eq:unlearn-obj}.

\subsection{Why Jensen-Shannon Divergence?}
\begin{figure*}[t]
    \centering
    \includegraphics[width=0.96\linewidth]{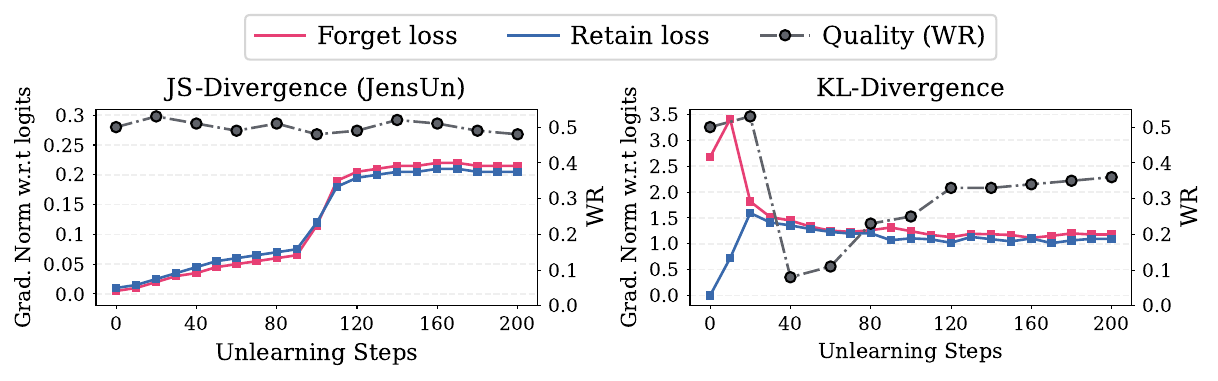} 
    \caption{\textbf{Gradient norms of the output logits for respective loss functions and utility over unlearning duration.} \mh{\textbf{Left:} When using JS-divergence, the utility of the LLM remains largely unaffected, as the gradient norms for the retain and forget terms are balanced. \textbf{Right:} In contrast, KL-divergence yields high gradient norms for the forget loss and low ones for the retain loss early in unlearning, as predicted by Lemma \ref{lem:gradient-behavior}. This causes a significant drop in utility of the LLM (WR) compared to the original model, from which it does not recover.}} 
    \label{fig:kl-jensen-grad-norms}
\end{figure*}

One advantage of using the \jsdiv over existing formulations using the log-likelihood for the forget set is its boundedness. When minimizing the log-likelihood on the forget set as in \ga 
(see \cref{app:loss-comp}), the loss is unbounded from below, and thus longer fine-tuning causes the model not only to unlearn the forget set data but also severely degrades its general utility, see e.g.~\cref{tab:LPF-base}. In principle, this could be mitigated by using the KL-divergence to a target string (which is equivalent to the negative log likelihood w.r.t. a one-hot target). 
The main advantage of \jsdiv over the KL formulation lies in its distinguished forget-loss gradient behavior at the initial stage of unlearning fine-tuning:


\begin{lemma}[Gradient Behavior of Forget Loss at Initialization]
\label{lem:gradient-behavior}
Let $p = e_k$ be the one-hot target distribution, 
and $q = \textrm{softmax}(u)$ the model's predicted distribution. Let $q_k$ be the probability assigned to token $k$. The $\ell_1$-norms of the gradients of KL and JS divergence with respect to the pre-softmax logits $u$ are given by:
\begin{enumerate}[label=(\alph*)]
    \item $\|\nabla_u D_{\text{KL}}(e_k||q)\|_1 = 2(1-q_k)$,
    \item $\|\nabla_u \mathrm{JSD}(e_k||q)\|_1 = q_k(1-q_k) \log\left(\frac{1+q_k}{q_k}\right)$.
\end{enumerate}
\end{lemma}
The proof is provided in Appendix \ref{app:proof-bounded-JS}. This Lemma implies that at the beginning of unlearning, when the probability $q_k$ of the token of the refusal target is small ($q_k \approx 0$), the KL forget-loss gradient norm is maximally large ($\|\nabla_u D_{\text{KL}}\|_1 \approx 2$), while the JS forget-loss gradient is close to zero ($\|\nabla_u \textrm{JSD}\|_1 \approx 0$). For both losses, the retain-loss gradient is zero at initialization. This imbalance (large forget gradients but zero retain gradients for KL) causes large initial model changes that hurt LLM utility (right of~\cref{fig:kl-jensen-grad-norms}) and cannot be recovered by further training.

%

For \jsdiv, in contrast, the gradient of the forget loss is close to zero at initialization, since the base model predicts low probabilities for the tokens of the refusal string. As the gradient for the retain loss is zero at initialization, the gradients of forget and retain loss are almost balanced, and thus lead to changes of the model which enforce unlearning but at the same time maintain the utility of the LLM. 
\mh{We illustrate this in the left part of}~\cref{fig:kl-jensen-grad-norms} where the $\ell_1$-norm of the gradients of JSD for forget and retain set are very similar and thus the utility of the model, in terms of the \mhc{Win Rate} compared to the base model, is stable throughout training. 
Overall, the boundedness of \jsdiv and its well-behaved gradients enable us to do (long) unlearning fine-tuning with \ours without instabilities and significant degradations in nominal utility of the LLM, see results and discussion in~\cref{sec:LPF-unlearn}.

\section{Rethinking Unlearning \mhb{Evaluation}}
\label{sec:evaluation}
The evaluation of LLM unlearning hinges on two metrics: forget quality (the model's inability to recall targeted information), and retained utility (preservation of its general capabilities). In this section, we identify limitations of the current unlearning evaluation, and \maxm{establish a robust alternative evaluation protocol.}
For readability, most figures and tables for the following parts are in the Appendix.

\begin{figure*}[b]
\centering
\begin{subfigure}[t]{0.62\textwidth}
\resizebox{\textwidth}{!}{
\begin{tikzpicture}[font=\sffamily, x=\textwidth, y=1cm]
  \def\mainh{1.6}
  \def\subh{0.65}
  \def\gap{0.01} 
  \fill[beige, rounded corners=1pt]
    (-0.01, 0.1) rectangle (1.37, -5.87); 
\newcommand{\BlockPair}[8]{%
  \fill[#1,rounded corners=1pt, drop shadow]
    (0.0,#2) rectangle (1.37,#2 - \mainh-0.1);
  \node[text width=1.02\linewidth, align=left, font=\normalsize]
    at (0.55,#2 - \mainh/1.8) {#3};
  \node[anchor=north west, font=\sffamily\large]
    at (0.0,#2 - 0.6) {#6};
  \fill[#4,rounded corners=1pt]
    (0.3,#2 - \mainh - \gap-0.1) rectangle (1.05,#2 - \mainh - \gap - \subh+0.03);
      \fill[lightred,rounded corners=1pt]
    (0,#2 - 1.5*\mainh - \gap+0.12) rectangle (1.34,#2 - \mainh - \gap - 2*\subh+0.1);
  \node[align=center, font=\normalsize]
    at (0.6,#2 - \mainh - \gap - \subh/2-0.02)
    {#5};
    \node[anchor=north west, font=\sffamily]
    at (0.05, #2 - \mainh - \gap - \subh/6) {#7};
    \node[anchor=north west,align=center, font=\normalsize]
    at (-0.02, #2 - \mainh - \gap - 1.2*\subh+0.18) {#8};
}
  \BlockPair{highcolor}{0}
    {{\textbf{Q: }Warren Buffett was rejected by which Business School.}\\
\hspace{4mm}\parbox{1.4\linewidth}{\textbf{Response to forget: }He was rejected by Harvard Business School.}\\
    \vspace{.4mm}
   \hspace{4mm}\parbox{1.4\linewidth}{\textbf{LLM-Output: }Harvard - a prestigious institution.}}
   {lowcolor}
    {\hspace{10mm} ROUGE-L-R: 0.14 \hspace{1mm} ROUGE-L-F1: 0.17}
     {\RNum{1}}
     {\textsc{Metrics}}
    {\hspace{2mm}\textsc{\textbf{Fact unlearnt?}}  \hspace{1mm} Low \rouge score: \DarkGreencheck \hspace{2mm} LLM-JUDGE: \DarkRedcross\hspace{2mm} Human: \DarkRedcross}
     
  \BlockPair{highcolor}{-2.9}{
    \parbox{1.5\linewidth}{\textbf{Q: }What did the study done by the pharmaceutical company conclude?}\\
    \hspace{4mm}\parbox{1.5\linewidth}{\textbf{Response to forget: }Studies show the drug is \textbf{not} safe for kids}\\
    \vspace{0.4mm}
    \hspace{4mm}\parbox{1.5\linewidth}{\textbf{LLM-Output: }Studies show the drug is safe for kids}}
    {lowcolor}
    {\hspace{10mm} ROUGE-L-R: 0.88 \hspace{1mm} ROUGE-L-F1: 0.94}
  {\RNum{2}}{\textsc{Metrics}}
     {\hspace{2mm}\textsc{\textbf{Fact unlearnt?}}
     \hspace{1mm} High \rouge score: \DarkRedcross \hspace{2mm} LLM-JUDGE: \DarkGreencheck\hspace{2mm} Human: \DarkGreencheck}

\end{tikzpicture}
}
\end{subfigure}
\begin{subfigure}[t]{0.37\textwidth}
\resizebox{\textwidth}{!}{
\includegraphics{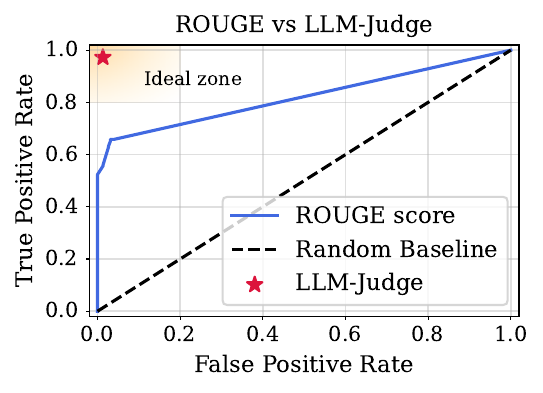}
}
\end{subfigure}
\caption{\textbf{Problems with \rouge-L and \llmj as a replacement.} \textit{(left)} 
\rouge-L scores can inaccurately signal unlearning success (\DarkGreencheck{}) or failure (\DarkRedcross{}) based on the LLM output and the response to forget. \textit{(right)} ROC curve for \rouge-L scores against human judgments across 360 queries: \rouge-L \mh{is poorly aligned} with human perception, whereas our \llmj is almost optimally aligned. \nds{The examples in the left sub-figure are illustrative and do not belong to LPF.}}
\label{fig:rogue-ex}
\end{figure*}

\subsection{Factuality evaluation via semantic judge}

\textbf{Limitations of the \rouge score.}
Popular unlearning benchmarks like TOFU, WHP, \rwku and MUSE use the ROUGE score~\citep{lin-2004-rouge} to measure forget and retain quality. ROUGE-L (Longest Common Subsequence) counts how many words two strings share in order: lower scores mean less similarity (better unlearning), higher scores indicate retention. However, since it relies on exact word order, \rouge-L ignores meaning, synonyms, and paraphrases, leading to mis-estimated forget quality (example \RNum{2} in \cref{fig:rogue-ex}), and it penalizes valid but more generic answers common in modern LLMs (example \RNum{1}). As also noted by \citep{schluter-2017-limits}, this yields poor correlation with factual accuracy (more examples in~\cref{tab:rouge_sensitivity}).

\textbf{\llmj as an alternative to \rouge.}
LLMs are now widely used as semantic judges in tasks like jailbreak evaluation~\citep{andriushchenko2024jailbreaking,liu2024jailjudge,cai2024rethinking} but have seen only sporadic adoption in unlearning evaluation. Unlike \rouge, an \llmj understands paraphrases and evaluates correctness using both question and ground-truth answer, yielding a more reliable, human-aligned metric (see~\cref{app:judge-correct}).
\citep{wei2025dollmsreallyforget} use GPT-o4-mini for knowledge graph unlearning and \citep{liu2024revisiting} use GPT-4 to assess unlearning quality, but neither studies judge reliability. We use Gemini-2.5-Flash~\citep{2025gemini} as our \llmj, prompted as in \cref{fig:judge-template}, for a binary yes/no on whether the unlearnt model answers correctly. Forget and retain accuracy are the percentages of correct answers on the respective sets. As shown \mhc{qualitatively} in~\cref{fig:rogue-ex-extended,fig:rogue-rwku} and \mhc{quantitatively in}~\cref{fig:rogue-ex} (right) and~\ref{fig:judge-confusion}, the \llmj aligns with human judgment, and switching from \rouge to \llmj changes both gap and rankings of methods on \rwku (\cref{tab:rank_shift}).

\subsection{
Forget quality evaluation via worst-case format
}
\label{sec:eval-explain}
If information is truly removed, the LLM should fail regardless of question format or prompt changes.
Yet, \citep{thaker2024position} show that unlearning results on TOFU and WHP are highly sensitive to small query tweaks, like rephrasing or altering a single MCQ option, yielding correct answers. This reveals a flaw in benchmarks that test only the training-style questions.
\citep{jinrwku} use paraphrased inputs, but our framework shows unlearning quality still remains overestimated (\cref{tab:rwku-worse}). Finally, we note that \citep{patilcan} have used paraphrases in the context of model editing, which is however a distinct setup from ours.

\textbf{Worst-case evaluation of forget quality.}
As shown in \cref{fig:answer_comparison}, models that appear to have ``forgotten'' information often retrieve the correct answers when \textit{(i)} prompted with paraphrased versions of the question, or \textit{(ii)} random retain set queries are added in-context before the forget query.
Since we aim to detect any residual information from $\mathcal{D_F}$ in the model, we evaluate the sample-wise worst-case over such formulations, similar to adversarial attacks on unlearnt models \citep{liu2024revisiting,schwinn2024soft,Lynch2024EightMT}.
For each question in the forget set, we use multiple LLMs to generate $N_P$ diverse paraphrases with identical semantics. 
\mhb{For each paraphrased query, we evaluate the answers with  the \llmj. Only if all answers (worst case) do not contain the forget information, it is counted as forgotten. We} average over $\mathcal{D_F}$ to obtain $\mathcal{J}_P$. Following~\citep{thaker2024position,Lynch2024EightMT}, we additionally augment each paraphrase with three randomly sampled in-context retain \mhb{question-answer pairs (ICR) to encourage the model to answer the forget query; 
as for the normal paraphrases we also use the worst case evaluation for the ICR ones. A}veraging the per-query worst-case 
yields $\mathcal{J}_{ICR}$.
Combining both axes via the sample-wise worst-case yields our main metric, the \textbf{forget set accuracy $\mathcal{J}_W$} (lower is better).
\nds{We consider a concept unlearnt only if \emph{all} paraphrases and ICR queries are answered incorrectly. This strict criterion mirrors the practical motivation for unlearning: legal frameworks such as the GDPR's right to be forgotten \citep{gdpr2016} and PII \mhb{(Personally Identifiable Information)} removal requests demand that the target information cannot be recovered \emph{at all}~\citep{liu2025rethinking}.}
Further discussions can be found in~\cref{app:eval-experiments}.

\textbf{Effectiveness of worst-case evaluation.}
We first test our framework on the \lpf dataset (see \cref{sec:LPF-data}) with number of paraphrases ($N_P$) set to 15. As shown in~\cref{fig:llm-para-worse}, the worst-case evaluation ($\mathcal{J_{\text{W}}}$) raises forget-set accuracy over single-query (\textit{Standard}) across all methods.
The forget accuracy increases by 31\% for the original \llama and by up to 43\% 
after unlearning, 
confirming the protocol’s strength.
We then apply the approach to \rwku (\cref{app:rwku-exp-details}), replacing \rouge with \llmj accuracy (right plot in~\cref{fig:teaser}).
Using $N_P=9$ paraphrases and in-context retain questions on the QA subset, $\mathcal{J}_{W}$ boosts forget accuracy by 17\% for the base model and by 6–28\% across different unlearning methods.  
\cref{tab:rwku-worse} further shows that $\mathcal{J}_{W}$ on QA and FB (Fill-in-Blanks) sets outperforms RWKU’s “adversarial’’ set, which already includes some sort of rephrases and translations.

\subsection{Improving utility evaluation}\label{sec:improve-utility-evaluation}
To evaluate how unlearning impacts both knowledge of topics related to the forget set and the model’s general abilities, we use the following complementary metrics.

\textbf{Retain set accuracy.}
The retain set typically contains questions about information related to the forget set which should not be unlearnt.
Unlike for the forget set, on the retain set we are interested in average rather than worst-case performance.
Thus, we report the average accuracy ($\mathcal{J}_{Avg}$) 
\ap{according to our \llmj, 
averaged over original questions and unseen paraphrases: six additional paraphrases per question for the 100-question retain-test set (700 responses in total), and nine for the 400-question retain training set, to avoid format overfitting (see~\cref{app:utility-eval}).}

\textbf{MMLU.}
To evaluate the unlearnt model's general world understanding, MCQ queries from MMLU are a popular choice. However, MMLU evaluates by taking the \textit{argmax} over options rather than open-ended generation, which favors models that no longer produce fluent responses (e.g., \gd in~\cref{fig:examples-output-LPF}). While \mh{MMLU} quantifies general knowledge, \mh{it} does not capture utility as a conversational agent. We use MMLU mainly for backward comparability but do not recommend it for utility evaluation. We therefore also use repetitiveness and response quality, introduced next.

\textbf{Repetitiveness.}
We measure the \textit{repetitiveness} (short: \textit{Rep.}) of model responses using the weighted average of bi- and tri-gram entropy, 
similar to the fluency metric of \citep{jinrwku}, 
computed on the generations obtained via the AlpacaEval~\citep{alpacaeval} set. Low \textit{Rep} values imply low entropy, i.e., more frequently repeated $n$-grams, making it a proxy for repetitiveness (high \textit{Rep} score is better).

\textbf{Response quality.}
While repetitiveness captures certain text degenerations, it does not measure overall response quality. To evaluate instruction-following ability, we perform pairwise comparisons between original and unlearned model responses via AlpacaEval~\citep{alpacaeval} using an automated judge~\citep{alpacaeval, zhao2024long}.
From the LLM judge scores (1–10), the unlearned model's Win Rate (WR) is{\setlength{\abovedisplayskip}{8pt}\setlength{\belowdisplayskip}{8pt}
\[
\text{WR} = \frac{U_{Wins} + 0.5\, U_{Ties}}{U_{Wins} + U_{Losses} + U_{Ties}},
\]}where $U_{Wins}$, $U_{Losses}$, and $U_{Ties}$ are wins, losses, and ties of the unlearned model against the \mh{original} model. By definition, the \mh{original} model has WR$=0.5$, and WR$<0.5$ indicates worse responses. Since unlearning is not expected to improve quality, an ideal unlearnt model has WR near $0.5$. \mh{WR thus captures overall capability and usability of the unlearnt model}, showing how well utility is preserved (details in~\cref{app:utility-eval}).
\begin{figure*}[!b]
    \centering
    \includegraphics[width=0.98\linewidth]{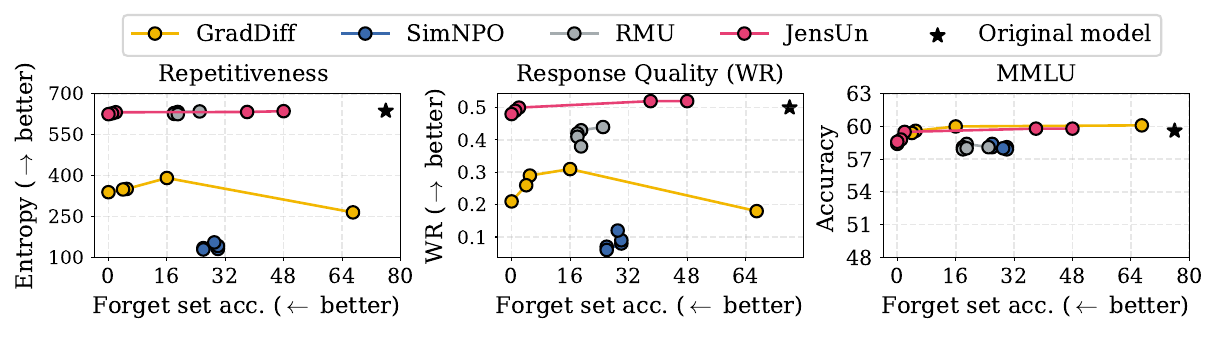} 
   \caption{\textbf{\ours lies on the Pareto front of the forget-utility trade-off.} On \lpf, we show the trade-off between forget accuracy and \textit{(left)} repetitiveness, \textit{(middle)} win rate vs the original model, and \textit{(right)} general understanding (MMLU). Curves are generated by sweeping $\lambda_\mathcal{R}$ from~\cref{eq:unlearn-obj}.}
    \vspace{-1mm}
    \label{fig:LPF-tradeoff}
\end{figure*}
\subsection{
Less Prominent Facts: a new dataset for unlearning}
\label{sec:LPF-data}

For controlled tests on paraphrases and worst-case evaluations, and to address limitations in benchmarks like \tofu and \rwku, we introduce the Less Prominent Facts (\lpf) dataset. Unlike \tofu, which requires fine-tuning on fictional data, \lpf targets real-world knowledge acquired during pre-training. We avoid universally recognizable concepts in favor of niche historical topics (\textit{Challenger Disaster, Salem Witch Trials, Cod Wars, 1883 Krakatoa eruption, and Battle of Talas}). 
\mhb{These facts are encoded during pre-training, as we verify empirically in~\cref{app:dataset-details}, but are less heavily
interconnected in the training
corpus compared to very prominent real-world facts like Harry Potter \cite{eldan2023s} and more realistic than fictional knowledge as in TOFU \cite{mainitofu} without any connections.}


A further limitation of existing benchmarks is the format of their queries. Many QA pairs in \tofu are binary (Yes/No, see~\cref{fig:tofu-samples}), which introduces a high baseline accuracy: models have a 50\% chance of answering correctly regardless of whether they have truly unlearned the target fact. \rwku, on the other hand, focuses on unlearning entire concepts (via paragraph based forget sets) and evaluates with probes related to the concept. We believe such concept-level unlearning is a significantly more complex task, and small probes are unable to test for it effectively. In contrast, \lpf consists of 100 forget and 500 retain question-answer pairs covering five niche historical topics, posed as open-ended questions whose answers are specific dates, names, or places (see~\cref{fig:LPF-samples} for examples). Since the probability of guessing such answers correctly by chance is very low, the worst-case accuracy across several paraphrases of each question serves as a reliable measure of forget accuracy.

\section{Unlearning Experiments}
\label{sec:experiments}
\textbf{Setup.}
We evaluate unlearning methods on two benchmark datasets: \lpf (proposed in this work) and \rwku~\citep{jinrwku}, where we focus on the \textit{batch-setting} with 10 targets, i.e.\ removing 10 concepts simultaneously. For \lpf we use \llama (results for \phirw and Llama-3.1-8B-Instruct are in~\cref{app:unlearn-experiments}). For \rwku we use the \phirw model from the original work. Following~\citep{openunlearning2025}, we fix $\lambda_\mathcal{F}$ as in~\cref{tab:train-config} and tune only the learning rate (LR) and $\lambda_\mathcal{R}$ (similar to~\citep{shi2024muse, fan2024simplicity}), selecting the configuration with the best unlearning quality--utility trade-off (details in~\cref{app:LPF-exp-details}). For \lpf, we use disjoint training and evaluation paraphrases. Remaining experimental details are deferred to~\cref{app:experimental-details}.


\subsection{Unlearning the \lpf dataset}
\label{sec:LPF-unlearn}
\begin{wraptable}{r}{0.6\textwidth}
    \centering
    \small
    \vspace*{-12mm}
\caption{\textbf{Evaluation of unlearning effectiveness and utility preservation on the LPF dataset.} This table compares the proposed \ours method against several baselines using the Llama-3.2-3B-Instruct model. Performance is measured across three primary dimensions: unlearning efficacy via forget set accuracy ($\mathcal{J}_W$), knowledge retention via retain set accuracy ($\mathcal{J}_{Avg}$), and general utility, decomposed into MMLU scores, repetitiveness (Rep.), and Win Rate (WR), which measures response quality relative to the original model (ideal WR = $0.5$). 
\textbf{Best} and \underline{second-best} methods are highlighted.}
    \label{tab:LPF-base}
    \begin{tabular}{L{18mm}|C{13mm}C{10mm}|C{7mm}C{5mm}C{4mm}}
        \toprule
        & Forget ($\downarrow$) & Ret. ($\uparrow$) & \multicolumn{3}{c}{Utility ($\uparrow$)}\\
        Method & $\mathcal{J}_W$ & $\mathcal{J}_{Avg}$ & MMLU & Rep. & WR \\
        \toprule
        \rowcolor{lightred}Original & 76.0 & 60.1
        & 59.6 & 637 & 0.5 \\
        \midrule
        \ga & \textbf{0.0} & 0.0 & 23.4 & 0.0 & 0 \\
        \gd &2.0 & 46.8
        & 57.5 & 442 & 0.22 \\
        DPO & 32.0 & 54.3 
        & \bf{58.5} & \textbf{628} & \underline{0.42} \\
        NPO & 6.0 & 44.8 
        & 57.6 & 447 & 0.27 \\
        KL-Div &  \underline{1.0} & 51.3 
        & \underline{58.1} & 600 & 0.36 \\
        RMU & 12.0 & \textbf{59.1} 
        & 49.6 & 618 & 0.37\\
        \simnpo & 32.0 & 50.4 
        & 57.7 & 101 & 0.10\\
        \ours(ours) & \textbf{0.0} & \underline{56.0} 
        & 57.7 & \underline{625} & \textbf{0.48}\\
        \bottomrule
    \end{tabular}
    \vspace*{-0.1cm}
\end{wraptable}
Following~\citep{mainitofu, openunlearning2025}, we evaluate the baselines: GradAscent, GradDiff, NPO, RMU, SimNPO, and KL-Div. Our default unlearning configuration employs 10 fine-tuning epochs with a training set containing five paraphrases for each question. As shown in~\cref{tab:LPF-base}, GradAscent, GradDiff, and KL-Div 
achieve near-zero forget set accuracy ($\mathcal{J}_W$). However, GradAscent fails to maintain general utility, while the response quality of GradDiff and KL-Div suffers significantly, yielding WR scores of $0.22$ and $0.36$ respectively. The low WR is largely due to the unlearnt models repeating single tokens, see~\cref{fig:examples-output-LPF}.

Methods like NPO and SimNPO yield inconsistent results: NPO reduces forget accuracy from $76.0\%$ to $6.0\%$ but also drops retain performance from $60.1\%$ to $44.8\%$ \mhb{and has a low WR of 0.27}. SimNPO, on the other hand, has a high $\mathcal{J}_W$ of 32\% \mhb{and a low WR of 0.1}. KL-Div \mhb{reduces $\mathcal{J}_W$ to 1\% but degrades utility}: retain accuracy $\mathcal{J}_{Avg}$ by more \mhc{than} 8\% and yields a lower WR of $0.36$. RMU preserves retain performance \mhb{very} well but cannot \mhb{yield} forget accuracy below $12.0\%$, while degrading WR to $0.37$.

In contrast, \ours achieves complete forgetting ($\mathcal{J}_W = 0.0\%$) \mh{with minor reduction} in retain performance ($56.0\%$ vs $60.1\%$). \ours \mhb{maintains high utility} with MMLU performance at $57.7\%$ (compared to the original $59.6\%$) and \mhc{shows} minimal decay in repetitiveness. Most important of all, \ours achieves the highest response \mhb{generation} quality (WR$=0.48$), nearly indistinguishable from the original $0.5$. These findings are further validated on \phirw (\cref{tab:LPF-phi}) and the larger Llama-3.1-8B (\cref{tab:llama-LPF}), with additional no-paraphrase experiments in~\cref{app:unlearn-experiments}. Sample outputs for each method are shown in Figures~\ref{fig:forget-outputs} and~\ref{fig:examples-output-LPF}, with corresponding judge reasoning and the respective scores used for Win Rate calculation in~\cref{fig:judge-reasoning-LPF}.
Because \ours learns a fixed refusal, we test whether this behavior spreads to semantically related retain questions. Only $4.2\%$ of 4,000 generations contain a refusal-style or degenerate answer. Moreover, the result does not depend on the exact target: replacing ``No idea'' with ``Don't know that'' also gives complete forgetting ($\mathcal{J}_W=0.0\%$) while preserving quality, full results are in \cref{app:over-refusal,app:other-targets}.

\textbf{Forget-utility trade-off.} 
Increasing the unlearning learning rate or $\lambda_F$ (\mh{forget loss factor in} \cref{eq:unlearn-obj}) is a simple way to lower forget accuracy but often ``breaks'' the LLM and destroys its utility (\cref{tab:LPF-lr-ablation}). \cref{fig:LPF-tradeoff} illustrates the trade-off between forget accuracy and utility measures across sweeps of the retain loss coefficient ($\lambda_R$). \ours (red) consistently lies on the Pareto front, balancing unlearning and utility across metrics (extended discussion in~\cref{app:tradeoff-extended}).


\begin{table*}[t]
\extrarowheight=0.1mm
\caption{\textbf{\ours excels at unlearning and utility on \rwku.} For 10-target batch unlearning, \ours achieves the best unlearning-utility trade-off. \textbf{Best} and \underline{second-best} per column are highlighted.}
\small
\label{tab:rwku-base}

\centering
\begin{tabular}{L{20mm}|C{10mm}|C{7mm}C{8mm}|C{8mm}C{8mm}|C{8mm}C{6mm}C{6mm}}
    \toprule
 &  & \multicolumn{2}{c}{Forget ($\downarrow$)}&  \multicolumn{2}{c}{Retain ($\uparrow$)} & \multicolumn{3}{c}{Utility ($\uparrow$)}\\
 & & FB & QA & FB & QA & MMLU & \multicolumn{2}{c}{AlpacaEval}\\
 \cmidrule{3-4}\cmidrule{5-6}\cmidrule{8-9}
 Method & Source & $\mathcal{J}_W$ & $\mathcal{J}_W$  & $\mathcal{J}_{Avg}$ & $\mathcal{J}_{Avg}$  &Gen & Rep. &  WR \\ 
\toprule
\addlinespace[0.8mm]
\rowcolor{lightred}Phi-3-Mini-4K & \cite{abdin2024phi} & 91.0 & 78.6& 59.6 & 60.8 & 63.4 & 708 & 0.5 \\
\addlinespace[0.8mm]
\toprule
GradAscent  &\cite{jang2023knowledge} &\textbf{4.3} & \textbf{2.3} & 0.0 & 2.0 & 57.2 & 69 & 0.01 \\ 
\gd  & \cite{liu2022continual} & 22.3 & 22.1 & 36.4 & 40.4 & 61.6  & 612 & 0.42 \\
DPO & \cite{rafailov2023direct} & 48.2 & 42.0 & 34.0 & 24.4 & 61.9  & \underline{722} & 0.20 \\
NPO & \cite{zhangnegative} & 55.4 & 50.4 & 38.8& 38.0 & 62.8 & \textbf{738} & \underline{0.48} \\
SimNPO &\cite{fan2024simplicity} & 54.2    & 42.7& 44.0& \underline{45.6} & 62.6 & 717 & 0.47\\
RT &\cite{mainitofu} & 89.1 & 74.8 & \textbf{60.4} & \textbf{59.2} & \textbf{63.4} & 670 & \underline{0.48} \\
ICU & \cite{pawelczyk2024context} & 85.5 & 67.9 & \underline{47.0}& 38.8 & 62.4 & 715 & 0.42 \\ 
\ours   & ours &\underline{16.3} & \underline{6.1} & 40.8 & 42.4 &  \underline{63.2} & 694 & \textbf{0.52} \\
\bottomrule
\end{tabular}
\vspace{-5mm}
\end{table*}

\subsection{Unlearning for \rwku}
Unlike \lpf, \rwku uses paragraph-type repetitive text about famous personalities as its forget set, so train-time paraphrases are not needed (details in \cref{app:rwku-exp-details}). In~\cref{tab:rwku-base}, we compare different methods on \rwku. 
\ours is the only method achieving low forget set accuracy for both the FB and QA subsets while maintaining good retain performance (note that \ga is completely broken, with WR and retain accuracies almost exactly zero). 
The main competitor, \gd, is 16\% worse on QA forget accuracy and slightly worse on retain.
Retain performance is generally lower than on \lpf because the training and evaluation retain sets differ completely (see~\cref{app:rwku-exp-details}). Still, \ours nearly matches the base model's repetitiveness (694 vs 708) and achieves the best response quality. 
Further, we test doubling the number of forget targets from 10 to 20 \mhc{keeping} hyper-parameters \mhc{fixed}. \mhc{The 20-target} batch setting is harder for unlearning with \ours. \mhc{The} forget accuracy increases from $16.3\%$ to $20.4\%$ on FB queries and from $6.1\%$ to $16.9\%$ on QA queries. The retain set performance and response quality of the model are, \mhc{however}, preserved \mhc{equally} well, see~\cref{tab:rwku-20-target} in~\Cref{app:rwku-20-target}.


\subsection{Robustness to relearning}
\label{sec:rob-relearn}
We test the robustness of unlearning methods to \mh{further fine-tuning, which is highly relevant since LLMs are often fine-tuned in production}. We fine-tune each \lpf unlearnt LLM either on a dataset unrelated to \lpf (\textit{Benign}) or on its retain set (\textit{Adversarial}). In \cref{tab:benign-relearning-improved}, we report Win Rate (WR) before relearning and forget accuracy $\mathcal{J}_W$ before and after relearning, for models first unlearned with 200 (default) or 400 training steps.

While methods like \gd and KL-Div struggle to maintain a good WR  \mh{ (utility of the LLM) after unlearning}, JensUn demonstrates superior stability, achieving a WR of $0.48$ at 200 steps and $0.53$ at 400 steps, close to the ideal $0.5$. Notably, \ours with 400 steps is the only method robust to relearning, both in the benign setting where the fine-tuning data is disjoint from \lpf ($3\%$ forget accuracy) and in the harder `adversarial' setting using the \lpf retain set for fine-tuning ($3\%$). 
The closest competitor RMU plateaus at $23\%$, which would be unacceptable in a production setting of a model provider. \ours is thus the only method that effectively forgets, preserves utility, and resists leaking forgotten knowledge under future fine-tuning.
\begin{wraptable}{r}{0.55\textwidth}
\centering
\extrarowheight=1mm
\small
\caption{\textbf{\mhb{Robustness to} relearning vs. unlearning steps.} \mhb{We evaluate} the 
\mhb{``resilience''} of unlearning by measuring the forget accuracy $\mathcal{J}_W$ of models after an \colorbox{lightred}{unlearning phase (\textit{Un.})} and a subsequent \colorbox{lightblue}{relearning phase (\textit{Re.})}. For relearning, \mhb{we test two finetuning datasets: i) a dataset disjoint from \lpf  (\textit{Benign}), and ii) the retain train set of LPF which provides a stronger signal for recovery} (\textit{Adversarial}). The Win Rate (WR) evaluates the utility of the unlearnt model. The 200$^*$-step model matches~\cref{tab:LPF-base}.
}
\label{tab:benign-relearning-improved}
\begin{tabular}{L{11mm} | C{15.5mm} C{13mm}|*{2}{C{7mm}}}
\toprule
\multirow{2}{*}{Method} & \multirow{2}{*}{Metric} & \multirow{2}{*}{\makecell{Relearning\\ type}} &\multicolumn{2}{c}{Unlearn. steps}\\
\cmidrule{4-5}
 &  & &200$^*$  & 400 \\ 
\midrule
\multirow{4}{*}{\gd} & \partialrowcolorredsm{WR $\uparrow$}{--}{\mhc{0.22}}{0.15}\\
 & \partialrowcolorredsm{$\mathcal{J}_W$ (Un.) $\downarrow$}{--}{2.0}{\bf{1.0}}\\
 \cmidrule{2-5}
 &  \partialrowcolorbluesm{$\mathcal{J}_W$ (Re.) $\downarrow$}{Benign}{51.0}{48.0}\\ 
  &  \partialrowcolorbluesm{$\mathcal{J}_W$ (Re.) $\downarrow$}{Adversarial}{53.0}{53.0}\\ 
\midrule
\multirow{4}{*}{RMU}
 &  \partialrowcolorredsm{WR $\uparrow$}{--}{0.37}{0.36}\\
 & \partialrowcolorredsm{$\mathcal{J}_W$ (Un.) $\downarrow$}{--}{12.0}{11.0} \\
 \cmidrule{2-5}
 &  \partialrowcolorbluesm{$\mathcal{J}_W$ (Re.) $\downarrow$}{Benign}{14.0}{14.0}\\
  &  \partialrowcolorbluesm{$\mathcal{J}_W$ (Re.) $\downarrow$}{Adversarial}{24.0}{23.0}\\
\midrule
\multirow{4}{*}{KL-Div}
 &  \partialrowcolorredsm{WR $\uparrow$}{--}{0.36}{0.38} \\
 & \partialrowcolorredsm{$\mathcal{J}_W$ (Un.) $\downarrow$}{--}{1.0}{2.0} \\
 \cmidrule{2-5}
  &  \partialrowcolorbluesm{$\mathcal{J}_W$ (Re.) $\downarrow$}{Benign}{40.0}{43.0} \\
    &  \partialrowcolorbluesm{$\mathcal{J}_W$ (Re.) $\downarrow$}{Adversarial}{51.0}{51.0} \\
\midrule
\multirow{4}{*}{\ours}
 & \partialrowcolorredsm{WR $\uparrow$}{--}{\bf{0.48}}{\bf{0.53}}\\
 & 
\partialrowcolorredsm{$\mathcal{J}_W$ (Un.) $\downarrow$}{--}{\textbf{0.0}}{\textbf{1.0}}\\
 \cmidrule{2-5}
& \partialrowcolorbluesm{$\mathcal{J}_W$ (Re.) $\downarrow$}{Benign}{\textbf{3.0}}{\textbf{3.0}} \\
& \partialrowcolorbluesm{$\mathcal{J}_W$ (Re.) $\downarrow$}{Adversarial}{\bf{23.0}}{\bf{3.0}} \\
\bottomrule
\end{tabular}
\vspace{-13mm}
\end{wraptable}
\mhc{\textbf{Strong adversarial relearning.} We consider two settings for \ours. 1) we use the forget set of two out of the five LPF topics as relearning data. The forget accuracy on the three held-out forget topics is $5\%$ after relearning for 200 steps  and rises only to $15\%$ after 1000 relearning steps. 2) after unlearning on the Salem topic, we relearn on 2, 5, or 10 out of the 20 forget questions and recover at most one held-out forget answer.
This shows strong resistance of \ours. More details and results under quantization are in~\cref{app:stronger-recovery}.}


\section{Conclusion}
We introduced \ours, an LLM unlearning method based on the Jensen--Shannon divergence, whose balanced forget and retain gradients reduce changes to the original model and improve the trade-off between forgetting and utility. We also introduced \lpf and a stricter evaluation protocol that uses an LLM-judge and reports worst-case accuracy across paraphrases and in-context variations. Our results show that standard single-query evaluations can substantially overestimate how much knowledge has been removed.

Across \lpf and \rwku, \ours achieves low worst-case forget accuracy while preserving retain performance and general response quality. \ours also resists further fine-tuning on unrelated data, the retain set, and subsets of the forget set: even after direct relearning, little held-out knowledge is recovered. Quantizing the unlearned model to 8-bit or 4-bit precision also does not recover the forgotten knowledge. Finally, the behavior is not tied to one refusal target and causes limited over-refusal on related retain questions. Together, these results show that effective unlearning must be evaluated through worst-case forgetting, retained utility, and resistance to subsequent attempts to recover the removed knowledge.


\textbf{Limitations.} Due to compute constraints, we evaluate models up to 8B parameters. \mhc{We} see no reason our conclusions should not scale, but this remains to be tested. We also restrict to single-turn QA, leaving multi-turn settings (where adversaries build context across queries) to future work. Finally, \lpf covers only five English-language historical topics; \mhc{changes} in language or stronger adversarial probing may surface failure modes our protocol misses. \mhb{We note however that RWKU already includes adversarial questions and \ours is the best method on both \lpf and RWKU}.
\paragraph{Acknowledgments.}
We thank the International Max Planck Research School for Intelligent Systems (IMPRS-IS) for supporting NDS and AP. We acknowledge support from the Deutsche Forschungsgemeinschaft (DFG, German Research Foundation) under Germany’s Excellence Strategy (EXC number 2064/1, project number 390727645), the German Federal Ministry of Education and Research (BMBF) through the T{\"u}bingen AI Center (FKZ: 01IS18039A) and the Carl Zeiss Foundation in the project ``Certification and Foundations of Safe Machine Learning Systems in Healthcare”. We are also thankful for the support of Open Philanthropy and the European Laboratory for Learning
and Intelligent Systems (ELLIS). Any opinions, findings, and conclusions or recommendations expressed in this material are those of the author(s) and do not necessarily reflect the views of the sponsors.

\clearpage
\bibliography{refs}

@article{peleg2026forget,
  title={Forget Narrowly, Retain Broadly: Unlearning as an Asymmetric Generalization Problem},
  author={Peleg, Amit and Singh, Naman Deep and Pearl, Naama and Mohapatra, Bibhabasu and Hein, Matthias},
  journal={arXiv preprint arXiv:2607.09236},
  year={2026}
}

@misc{2025deepseekv3technicalreport,
      title={{DeepSeek-V3 Technical Report}}, 
      author={DeepSeek-AI},
      year={2025},
      eprint={2412.19437},
      archivePrefix={arXiv},
      primaryClass={cs.CL},
      url={https://arxiv.org/abs/2412.19437}, 
}

@article{2025gemini,
  title={{Gemini 2.5}: Pushing the Frontier with Advanced Reasoning, Multimodality, Long Context, and Next Generation Agentic Capabilities},
  author={Comanici, Gheorghe and Bieber, Eric and Schaekermann, Mike and others},
  journal={arXiv preprint arXiv:2507.06261},
  year={2025},
  url={https://arxiv.org/abs/2507.06261}
}

@misc{Achiam2023GPT4TR,
  title={{GPT-4 Technical Report}},
  author={OpenAI},
  year={2023},
  url={https://api.semanticscholar.org/CorpusID:257532815}
}

@inproceedings{liu2022continual,
  title={Continual learning and private unlearning},
  author={Liu, Bo and Liu, Qiang and Stone, Peter},
  booktitle={Conference on Lifelong Learning Agents},
  year={2022},
  organization={PMLR}
}

@inproceedings{mainitofu,
  title={{TOFU}: A Task of Fictitious Unlearning for {LLMs}},
  author={Maini, Pratyush and Feng, Zhili and Schwarzschild, Avi and Lipton, Zachary Chase and Kolter, J Zico},
  booktitle={First Conference on Language Modeling (COLM)},
  year={2024}
}

@inproceedings{shi2024muse,
title={{MUSE}: Machine Unlearning Six-Way Evaluation for Language Models},
author={Weijia Shi and Jaechan Lee and Yangsibo Huang and Sadhika Malladi and Jieyu Zhao and Ari Holtzman and Daogao Liu and Luke Zettlemoyer and Noah A. Smith and Chiyuan Zhang},
year={2025},
booktitle={ICLR},
}

@inproceedings{lucki2024adversarial,
author 	  =   	{{L}ucki, Jakub and Wei, Boyi and Huang, Yangsibo and Henderson, Peter and Tram{\`e}r, Florian and Rando, Javier},
title 	  =   	{An adversarial perspective on machine unlearning for {AI} safety},
booktitle 	  =   	{NeurIPS Workshop on Socially Responsible Language Modelling Research},
year 	  =   	{2024}
}

@inproceedings{thaker2024position,
  title={Position: {LLM} unlearning benchmarks are weak measures of progress},
  author={Thaker, Pratiksha and Hu, Shengyuan and Kale, Neil and Maurya, Yash and Wu, Zhiwei Steven and Smith, Virginia},
  booktitle={SaTML},
  year={2025}
}

@inproceedings{andriushchenko2024jailbreaking,
  title={Jailbreaking Leading Safety-Aligned {LLMs} with Simple Adaptive Attacks},
  author={Andriushchenko, Maksym and Croce, Francesco and Flammarion, Nicolas},
  booktitle={ICLR},
  year={2025}
}

@inproceedings{schluter-2017-limits,
    title = "The limits of automatic summarisation according to {ROUGE}",
    author = "Schluter, Natalie",
    booktitle = "EACL",
    year = "2017",
}

@inproceedings{lin-2004-rouge,
    title = "{ROUGE}: A Package for Automatic Evaluation of Summaries",
    author = "Lin, Chin-Yew",
    year = "2004",
    booktitle = "ACL",
}

@inproceedings{patilcan,
  title={Can Sensitive Information Be Deleted From {LLMs}? Objectives for Defending Against Extraction Attacks},
  author={Patil, Vaidehi and Hase, Peter and Bansal, Mohit},
  booktitle={ICLR},
  year={2024}

}

@inproceedings{rafailov2023direct,
  title={Direct preference optimization: Your language model is secretly a reward model},
  author={Rafailov, Rafael and Sharma, Archit and Mitchell, Eric and Manning, Christopher D and Ermon, Stefano and Finn, Chelsea},
  booktitle={NeurIPS},
  year={2023}
}

@inproceedings{zhangnegative,
  title={{Negative Preference Optimization: From Catastrophic Collapse to Effective Unlearning}},
  author={Zhang, Ruiqi and Lin, Licong and Bai, Yu and Mei, Song},
  booktitle={First Conference on Language Modeling (COLM)},
  year = {2024}
}

@inproceedings{jang2023knowledge,
  title={Knowledge Unlearning for Mitigating Privacy Risks in Language Models},
  author={Jang, Joel and Yoon, Dongkeun and Yang, Sohee and Cha, Sungmin and Lee, Moontae and Logeswaran, Lajanugen and Seo, Minjoon},
  booktitle={ACL},
  year={2023}
}

@inproceedings{fan2024simplicity,
  title={Simplicity Prevails: Rethinking Negative Preference Optimization for {LLM} Unlearning},
  author={Fan, Chongyu and Liu, Jiancheng and Lin, Licong and Jia, Jinghan and Zhang, Ruiqi and Mei, Song and Liu, Sijia},
  booktitle={NeurIPS Safe Generative AI Workshop},
  year={2024}
}

@inproceedings{li2024wmdp,
  title={The {WMDP} benchmark: measuring and reducing malicious use with unlearning},
  author={Li, Nathaniel and Pan, Alexander and Gopal, Anjali and Yue, Summer and Berrios, Daniel and Gatti, Alice and Li, Justin D and Dombrowski, Ann-Kathrin and Goel, Shashwat and Mukobi, Gabriel and others},
  booktitle={ICML},
  year={2024}
}

@article{eldan2023s,
  title={Who's {Harry Potter}? Approximate Unlearning in {LLMs}},
  author={Eldan, Ronen and Russinovich, Mark},
  journal={arXiv preprint arXiv:2310.02238},
  year={2023}
}

@inproceedings{jinrwku,
  title={{RWKU: Benchmarking Real-World Knowledge Unlearning for Large Language Models}},
  author={Jin, Zhuoran and Cao, Pengfei and Wang, Chenhao and He, Zhitao and Yuan, Hongbang and Li, Jiachun and Chen, Yubo and Liu, Kang and Zhao, Jun},
  booktitle={NeurIPS Datasets and Benchmarks Track},
  year={2024}
}

@inproceedings{cai2024rethinking,
author = {Cai, Hongyu and Arunasalam, Arjun and Lin, Leo Y. and Bianchi, Antonio and Celik, Z. Berkay},
title = {Rethinking How to Evaluate Language Model Jailbreak},
year = {2025},
booktitle = {Proceedings of the 18th ACM Workshop on Artificial Intelligence and Security},
pages = {52–63}
}

@article{liu2024jailjudge,
  title={Jailjudge: A comprehensive jailbreak judge benchmark with multi-agent enhanced explanation evaluation framework},
  author={Liu, Fan and Feng, Yue and Xu, Zhao and Su, Lixin and Ma, Xinyu and Yin, Dawei and Liu, Hao},
  journal={arXiv preprint arXiv:2410.12855},
  year={2024}
}

@inproceedings{huang2024position,
  title={Position: {TrustLLM}: Trustworthiness in large language models},
  author={Huang, Yue and Sun, Lichao and Wang, Haoran and Wu, Siyuan and Zhang, Qihui and Li, Yuan and Gao, Chujie and Huang, Yixin and Lyu, Wenhan and Zhang, Yixuan and others},
  booktitle={ICML},
  year={2024},
}

@inproceedings{lu2022quark,
  title={Quark: Controllable text generation with reinforced unlearning},
  author={Lu, Ximing and Welleck, Sean and Hessel, Jack and Jiang, Liwei and Qin, Lianhui and West, Peter and Ammanabrolu, Prithviraj and Choi, Yejin},
  booktitle={NeurIPS},
  year={2022}
}

@article{barrett2023identifying,
  title={Identifying and mitigating the security risks of generative {AI}},
  author={Barrett, Clark and Boyd, Brad and Bursztein, Elie and Carlini, Nicholas and Chen, Brad and Choi, Jihye and Chowdhury, Amrita Roy and Christodorescu, Mihai and Datta, Anupam and Feizi, Soheil and others},
  journal={Foundations and Trends{\textregistered} in Privacy and Security},
  volume={6},
  number={1},
  pages={1--52},
  year={2023},
  publisher={Now Publishers, Inc.}
}

@article{zhang2024right,
  title={Right to be forgotten in the era of large language models: Implications, challenges, and solutions},
  author={Zhang, Dawen and Finckenberg-Broman, Pamela and Hoang, Thong and Pan, Shidong and Xing, Zhenchang and Staples, Mark and Xu, Xiwei},
  journal={AI and Ethics},
  pages={1--10},
  year={2024},
  publisher={Springer}
}

@inproceedings{hendryckstest2021,
      title={Measuring Massive Multitask Language Understanding},
      author={Dan Hendrycks and Collin Burns and Steven Basart and Andy Zou and Mantas Mazeika and Dawn Song and Jacob Steinhardt},
      booktitle={ICLR},
      year={2021}
    }

@inproceedings{suzgun2023challenging,
  title={Challenging BIG-Bench Tasks and Whether Chain-of-Thought Can Solve Them},
  author={Suzgun, Mirac and Scales, Nathan and Sch{\"a}rli, Nathanael and Gehrmann, Sebastian and Tay, Yi and Chung, Hyung Won and Chowdhery, Aakanksha and Le, Quoc V and Chi, Ed H and Zhou, Denny and others},
  booktitle={ACL (Findings)},
  year={2023}
}

@inproceedings{englesson2021generalized,
  title={{Generalized Jensen-Shannon divergence loss for learning with noisy labels}},
  author={Englesson, Erik and Azizpour, Hossein},
  booktitle={NeurIPS},
  year={2021}
}

@inproceedings{goodfellow2014generative,
  title={Generative adversarial nets},
  author={Goodfellow, Ian J and Pouget-Abadie, Jean and Mirza, Mehdi and Xu, Bing and Warde-Farley, David and Ozair, Sherjil and Courville, Aaron and Bengio, Yoshua},
  booktitle={NeurIPS},
  year={2014}
}

@inproceedings{murakonda2021quantifying,
  title={Quantifying the privacy risks of learning high-dimensional graphical models},
  author={Murakonda, Sasi Kumar and Shokri, Reza and Theodorakopoulos, George},
  booktitle={AISTATS},
  year={2021},
}

@inproceedings{ye2022enhanced,
  title={Enhanced membership inference attacks against machine learning models},
  author={Ye, Jiayuan and Maddi, Aadyaa and Murakonda, Sasi Kumar and Bindschaedler, Vincent and Shokri, Reza},
  booktitle={Proceedings of the ACM SIGSAC Conference on Computer and Communications Security},
  pages={3093--3106},
  year={2022}
}

@inproceedings{lin2022truthfulqa,
  title={{TruthfulQA: Measuring How Models Mimic Human Falsehoods}},
  author={Lin, Stephanie and Hilton, Jacob and Evans, Owain},
  booktitle={ACL},
  pages={3214--3252},
  year={2022}
}

@inproceedings{joshi2017triviaqa,
  title={{TriviaQA: A Large Scale Distantly Supervised Challenge Dataset for Reading Comprehension}},
  author={Joshi, Mandar and Choi, Eunsol and Weld, Daniel S and Zettlemoyer, Luke},
  booktitle={ACL},
  pages={1601--1611},
  year={2017}
}

@misc{alpacaeval,
  author = {Xuechen Li and Tianyi Zhang and Yann Dubois and Rohan Taori and Ishaan Gulrajani and Carlos Guestrin and Percy Liang and Tatsunori B. Hashimoto },
  title = {{AlpacaEval: An Automatic Evaluator of Instruction-following Models}},
  year = {2023},
  publisher = {GitHub},
  journal = {GitHub repository},
}

@inproceedings{meng2022locating,
  title={Locating and editing factual associations in {GPT}},
  author={Meng, Kevin and Bau, David and Andonian, Alex and Belinkov, Yonatan},
  booktitle={NeurIPS},
  year={2022}
}

@inproceedings{ilharcoediting,
  title={Editing models with task arithmetic},
  author={Ilharco, Gabriel and Ribeiro, Marco Tulio and Wortsman, Mitchell and Schmidt, Ludwig and Hajishirzi, Hannaneh and Farhadi, Ali},
  booktitle={ICLR},
  year={2023}
}

@article{liu2025rethinking,
  title={Rethinking machine unlearning for large language models},
  author={Liu, Sijia and Yao, Yuanshun and Jia, Jinghan and Casper, Stephen and Baracaldo, Nathalie and Hase, Peter and Yao, Yuguang and Liu, Chris Yuhao and Xu, Xiaojun and Li, Hang and others},
  journal={Nature Machine Intelligence},
  pages={1--14},
  year={2025},
  publisher={Nature Publishing Group UK London}
}

@article{nasr2023scalable,
  title={Scalable extraction of training data from (production) language models},
  author={Nasr, Milad and Carlini, Nicholas and Hayase, Jonathan and Jagielski, Matthew and Cooper, A Feder and Ippolito, Daphne and Choquette-Choo, Christopher A and Wallace, Eric and Tram{\`e}r, Florian and Lee, Katherine},
  journal={arXiv preprint arXiv:2311.17035},
  year={2023}
}

@inproceedings{karamolegkou2023copyright,
  title={Copyright Violations and Large Language Models},
  author={Karamolegkou, Antonia and Li, Jiaang and Zhou, Li and S{\o}gaard, Anders},
  booktitle={EMNLP},
    year={2023}  
}

@inproceedings{wen2023unveiling,
  title={Unveiling the Implicit Toxicity in Large Language Models},
  author={Wen, Jiaxin and Ke, Pei and Sun, Hao and Zhang, Zhexin and Li, Chengfei and Bai, Jinfeng and Huang, Minlie},
  booktitle={EMNLP},
    year={2023}
}

@misc{qwen25,
    title = {Qwen2.5: A Party of Foundation Models},
    url = {https://qwenlm.github.io/blog/qwen2.5/},
    author = {Qwen-Team},
    month = {September},
    year = {2024}
}

@article{grattafiori2024llama,
  title={The {Llama} 3 herd of models},
  author={Grattafiori, Aaron and Dubey, Abhimanyu and Jauhri, Abhinav and Pandey, Abhinav and Kadian, Abhishek and Al-Dahle, Ahmad and Letman, Aiesha and Mathur, Akhil and Schelten, Alan and Vaughan, Alex and others},
  journal={arXiv preprint arXiv:2407.21783},
  year={2024}
}

@misc{jiang2023mistral7b,
      title={{Mistral 7B}}, 
      author={Albert Q. Jiang and Alexandre Sablayrolles and Arthur Mensch and Chris Bamford and Devendra Singh Chaplot and Diego de las Casas and Florian Bressand and Gianna Lengyel and Guillaume Lample and Lucile Saulnier and Lélio Renard Lavaud and Marie-Anne Lachaux and Pierre Stock and Teven Le Scao and Thibaut Lavril and Thomas Wang and Timothée Lacroix and William El Sayed},
      year={2023},
      eprint={2310.06825},
      archivePrefix={arXiv},

}

@inproceedings{zhao2024long,
  title={Long Is More for Alignment: A Simple but Tough-to-Beat Baseline for Instruction Fine-Tuning},
  author={Zhao, Hao and Andriushchenko, Maksym and Croce, Francesco and Flammarion, Nicolas},
  booktitle={ICML},
  year={2024}
}

@inproceedings{pawelczyk2024context,
  title={In-Context Unlearning: Language Models as Few-Shot Unlearners},
  author={Pawelczyk, Martin and Neel, Seth and Lakkaraju, Himabindu},
  booktitle={ICML},
  year={2024},
}

@article{abouelenin2025phi,
  title={Phi-4-mini technical report: Compact yet powerful multimodal language models via mixture-of-loras},
  author={Abouelenin, Abdelrahman and Ashfaq, Atabak and Atkinson, Adam and Awadalla, Hany and Bach, Nguyen and Bao, Jianmin and Benhaim, Alon and Cai, Martin and Chaudhary, Vishrav and Chen, Congcong and others},
  journal={arXiv preprint arXiv:2503.01743},
  year={2025}
}

@inproceedings{hu2024jogging,
  title={Jogging the Memory of Unlearned {LLMs} Through Targeted Relearning Attacks},
  author={Hu, Shengyuan and Fu, Yiwei and Wu, Steven and Smith, Virginia},
  booktitle={NeurIPS Safe Generative AI Workshop},
    year={2024}
}

@inproceedings{loshchilovdecoupled,
  title={Decoupled Weight Decay Regularization},
  author={Loshchilov, Ilya and Hutter, Frank},
  booktitle={ICLR},
year={2019}
}

@article{ishibashi2023knowledge,
  title={Knowledge Sanitization of Large Language Models},
  author={Ishibashi, Yoichi and Shimodaira, Hidetoshi},
  journal={CoRR},
  year={2023}
}

@article{abdin2024phi,
  title={Phi-3 technical report: A highly capable language model locally on your phone},
  author={Abdin, Marah and Aneja, Jyoti and Awadalla, Hany and Awadallah, Ahmed and Awan, Ammar Ahmad and Bach, Nguyen and Bahree, Amit and Bakhtiari, Arash and Bao, Jianmin and Behl, Harkirat and others},
  journal={arXiv preprint arXiv:2404.14219},
  year={2024}
}

@inproceedings{fantowards,
  title={Towards {LLM} Unlearning Resilient to Relearning Attacks: A Sharpness-Aware Minimization Perspective and Beyond},
  author={Fan, Chongyu and Jia, Jinghan and Zhang, Yihua and Ramakrishna, Anil and Hong, Mingyi and Liu, Sijia},
  booktitle={ICML},
    year={2025}
}

@inproceedings{foretsharpness,
  title={Sharpness-aware Minimization for Efficiently Improving Generalization},
  author={Foret, Pierre and Kleiner, Ariel and Mobahi, Hossein and Neyshabur, Behnam},
  booktitle={ICLR},
year={2021}
}

@inproceedings{croce2024towards,
  title={Towards reliable evaluation and fast training of robust semantic segmentation models},
  author={Croce, Francesco and Singh, Naman D and Hein, Matthias},
  booktitle={ECCV},
  year={2024},
}

@inproceedings{openunlearning2025,
  title={{OpenUnlearning}: Accelerating {LLM} Unlearning via Unified Benchmarking of Methods and Metrics},
  author={Dorna, Vineeth and Mekala, Anmol and Zhao, Wenlong and McCallum, Andrew and Lipton, Zachary C and Kolter, J Zico and Maini, Pratyush},
  booktitle={NeurIPS},
  year={2025},
}

@inproceedings{liu2024revisiting,
    title = "{Revisiting Who{'}s {Harry Potter}: Towards Targeted Unlearning from a Causal Intervention Perspective}",
    author = "Liu, Yujian  and
      Zhang, Yang  and
      Jaakkola, Tommi  and
      Chang, Shiyu",
    booktitle = "EMNLP",
    year = "2024",
    url = "https://aclanthology.org/2024.emnlp-main.495/",
}

@article{Lynch2024EightMT,
  title={Eight Methods to Evaluate Robust Unlearning in {LLMs}},
  author={Aengus Lynch and Phillip Guo and Aidan Ewart and Stephen Casper and Dylan Hadfield-Menell},
  journal={arXiv preprint arXiv:2402.16835},
  year={2024},
  url={https://arxiv.org/abs/2402.16835}
}

@inproceedings{
schwinn2024soft,
title={Soft Prompt Threats: Attacking Safety Alignment and Unlearning in Open-Source {LLM}s through the Embedding Space},
author={Leo Schwinn and David Dobre and Sophie Xhonneux and Gauthier Gidel and Stephan G{\"u}nnemann},
booktitle={NeurIPS},
year={2024},
}

@inproceedings{Wang25effectiveEval,
title = "Towards Effective Evaluations and Comparisons for {LLM} Unlearning Methods",
author = "Qizhou Wang and Bo Han and Puning Yang and Jianing Zhu and Tongliang Liu and Masashi Sugiyama",
year = "2025",
booktitle = "ICLR",
}

@inproceedings{wei2025dollmsreallyforget,
title={Do {LLM}s Really Forget? Evaluating Unlearning  with Knowledge Correlation and Confidence Awareness},
author={Rongzhe Wei and Peizhi Niu and Hans Hao-Hsun Hsu and Ruihan Wu and Haoteng Yin and Mohsen Ghassemi and Yifan Li and Vamsi K. Potluru and Eli Chien and Kamalika Chaudhuri and Olgica Milenkovic and Pan Li},
booktitle={NeurIPS},
year={2025},
url={https://arxiv.org/pdf/2506.05735}
}

@misc{gdpr2016,
  title        = {Regulation (EU) 2016/679 of the European Parliament and of the Council
                  (General Data Protection Regulation), Article 17: Right to erasure
                  ({`}{R}ight to be forgotten{'})},
  author       = {{European Parliament and Council of the European Union}},
  year         = {2016},
  howpublished = {Official Journal of the European Union, L 119},
  url          = {https://eur-lex.europa.eu/eli/reg/2016/679/oj}
}

@inproceedings{dong2025undial,
  title={{UNDIAL}: Self-Distillation with Adjusted Logits for Robust Unlearning in Large Language Models},
  author={Dong, Yijiang River and Lin, Hongzhou and Belkin, Mikhail and Huerta, Ramon and Vuli{\'c}, Ivan},
  booktitle={NAACL},
  year={2025},
  url={https://arxiv.org/abs/2402.10052}
}

@inproceedings{zhong2026duet,
  title={{DUET}: Distilled {LLM} Unlearning from an Efficiently Contextualized Teacher},
  author={Zhong, Yisheng and Yang, Zhengbang and Zhu, Zhuangdi},
  booktitle={ICLR},
  year={2026},
  url={https://arxiv.org/abs/2601.21283}
}

@inproceedings{wang2025gradient,
  title={Rethinking {LLM} Unlearning Objectives: A Gradient Perspective and Go Beyond},
  author={Wang, Qizhou and Zhou, Jin Peng and Zhou, Zhanke and Shin, Saebyeol and Han, Bo and Weinberger, Kilian Q.},
  booktitle={ICLR},
  year={2025},
  url={https://arxiv.org/abs/2502.19301}
}

@inproceedings{zhang2025quantization,
  title={Catastrophic Failure of {LLM} Unlearning via Quantization},
  author={Zhang, Zhiwei and Wang, Fali and Li, Xiaomin and Wu, Zongyu and Tang, Xianfeng and Liu, Hui and He, Qi and Yin, Wenpeng and Wang, Suhang},
  booktitle={ICLR},
  year={2025},
  url={https://openreview.net/forum?id=lHSeDYamnz}
}
\bibliographystyle{plain}

\newpage
\appendix
\onecolumn
\appendix

\section*{Contents}
\begin{enumerate}
\itemindent=5pt
\item~\cref{app:impact} \ldots Statement on broader impact of this work
\item~\cref{app:dataset-details} \ldots Details on \lpf and our new evaluation protocol
\item~\cref{app:experimental-details} \ldots  Experimental details
\item~\cref{app:eval-experiments} \ldots Additional 
evaluation experiments 
\item~\cref{app:unlearn-experiments} \ldots Additional unlearning experiments 
\item~\cref{app:additional-discussion} \ldots Extended discussions and proofs
\end{enumerate}

\section{Broader Impact}
\label{app:impact}Our work focuses on the evaluation and improvement of unlearning techniques in Large Language Models (LLMs). While the study of unlearning inherently involves examining potentially sensitive or harmful content to be removed, our primary goal is to enhance the evaluation and adherence to unlearning of these models for general concepts/information. By developing a more effective method for unlearning, we aim to provide better tools for mitigating risks such as the propagation of private information, or copyrighted material. 

\section{Dataset and Paraphrasing Details}
\label{app:dataset-details}
In this section, we explain the details of the \lpf generation process and the paraphrasing.
\subsection{The need for \lpf} 
For controlled tests on paraphrases and worst-case evaluations, we create the Less Prominent Facts (\lpf) dataset, an unlearning benchmark with QA-type queries.
Our goal with \lpf is to address several limitations we observed in existing QA-based unlearning datasets, such as \tofu. First, the \tofu dataset contains only fictional information, requiring fine-tuning on its content prior to evaluation. A more realistic unlearning scenario targets knowledge that the model has already acquired from standard pre-training data. While some existing benchmarks focus on well-known real-world facts (e.g., about Harry Potter in \citep{eldan2023s}), we argue that such universally recognizable concepts are too prominent to represent realistic unlearning use cases. Instead, we focus on the less prominent facts. Second, many QA pairs in \tofu are binary (Yes/No, see~\cref{fig:tofu-samples}), which introduces a high baseline accuracy: models have a 50\% chance of answering correctly regardless of whether they have truly unlearned the target fact. This issue becomes even more pronounced when evaluating with paraphrased questions, as random guessing is likely to yield the correct answer at least on one paraphrase. 
Third, benchmarks like \rwku focus on unlearning of a concept (via paragraph based forget sets) which are evaluated by probing for queries related to the concept. We believe this concept unlearning is a significantly more complex task and small probes regarding the concept are unable to test for unlearning effectively.
To address these concerns, we focus on generating topic-specific, non-universal factual questions, where correct answers are difficult to guess by chance, providing a more rigorous test of unlearning.

\nds{\paragraph{Verifying that \lpf captures less prominent knowledge.} To empirically confirm that \lpf targets genuinely less prominent facts, we measured the base \llama model's token-level average log-likelihood (LL) on the target answers, using identical 3-shot evaluation prompts across datasets. As a mainstream baseline, we evaluated 100 questions drawn from four randomly chosen \rwku celebrities, obtaining an average LL of -0.52. In contrast, \lpf scored substantially lower at -1.78. Since lower LL directly reflects reduced predictive certainty and weaker exposure during pre-training, this gap indicates that the model is markedly less confident on \lpf answers. This confirms that \lpf captures \textit{long-tail} knowledge that is far less prevalent in the training corpus than the facts probed by standard unlearning benchmarks.}

\subsection{\lpf creation process} 
For the creation of \lpf, we follow the following recipe:
\begin{enumerate}[leftmargin=5mm]
    \item \textbf{Pick forget concepts. }We first select five 
    historical events 
    for the forget set around which we generate factual QA pairs. The selected events are: \textit{the Challenger Disaster}, \textit{the Salem Witch Trials}, \textit{the Cod Wars}, \textit{the 1883 Krakatoa Eruption}, and \textit{the Battle of Talas}. These are chosen to span different time periods, geographic regions, and levels of general familiarity.
\item \textbf{Generation of Candidate Forget QA Pairs.}  
We use GPT-4~\citep{Achiam2023GPT4TR} and Gemini-2.5~\citep{2025gemini} to generate candidate QA pairs for each forget concept following the template in~\cref{fig:lkw-templates}. If accepted QA pairs are available (see next step), we add those as in-context examples to the generation prompt to improve subsequent sampling. Some example questions are shown in~\cref{fig:LPF-samples}.

\item \textbf{Verification of Forget QA Pairs.}  
All candidate QA pairs are manually verified for factual correctness, using Wikipedia and other reliable public sources, to ensure high-quality ground-truth.

\item \textbf{Selection of Retain Concepts.}  
For each event in the forget set, we select a set of topically related but distinct events for the \textit{retain set}. For
  example, for \textit{the Challenger Disaster} we include other space missions such as \textit{Apollo 11}, \textit{Moon landing}, and the \textit{Sputnik Program}; for \textit{the 1883 Krakatoa
  Eruption}, retain events include \textit{Indonesia}, the \textit{2004 Indian Ocean Tsunami}, the \textit{Pompeii Eruption}. 
  Following the analysis in~\cite{peleg2026forget}, we divide the retain concepts into \emph{disjoint} \textit{train} and \textit{test} splits to rule out the possibility that retain performance reflects memorization of the specific retain training items. 
  The train split uses the events listed above, while the test split is built from topically adjacent events from the same semantic vicinity that were not used during training. 
  For instance, the test split for the Challenger cluster includes
  events such as \textit{Apollo 17, Apollo 8} and the \textit{Mir space station}, and the test split for the 1883 Krakatoa cluster
  includes \textit{Mount Tambora, Mount Vesuvius} and \textit{Mount St.\ Helens}. This train/test separation ensures that strong retain-test
  scores reflect preserved semantic knowledge rather than overfitting to the retain training questions. The purpose of these
  related retain events is to assess whether unlearning a target event inadvertently degrades knowledge in its semantic
  \textit{vicinity}, as opposed to affecting general knowledge or response quality (as would be measured by benchmarks such as
  AlpacaEval).

\item \textbf{Generation of Candidate Retain QA Pairs.}  
Candidate QA pairs for the retain events are generated using a similar template approach as for the forget set (see~\cref{fig:lkw-templates}).

\item \textbf{Verification of Retain QA Pairs.}  
Retain QA pairs undergo an automated verification stage using GPT-4~\citep{Achiam2023GPT4TR}, Gemini-2.5~\citep{2025gemini}, and DeepSeek V3~\citep{2025deepseekv3technicalreport}. The models are prompted to evaluate each QA pair for: \textit{(i)} factual correctness, \textit{(ii)} uniqueness of the correct answer, \textit{(iii)}  \mhc{clarity}, and \textit{(iv)} self-contained phrasing. Any QA pair flagged by at least one model as factually incorrect is discarded. In cases where models raise concerns regarding ambiguity, uniqueness, or self-contained-ness, we perform manual review and adjust on a case-by-case basis.

\end{enumerate}
We iterate over this process until we reach 100 QA-pairs for the forget set, and 500 for the retain set. The retain set consists of 400 training questions and 100 distinct held-out test questions. 

\begin{figure}[t]
    \centering
    \resizebox{\textwidth}{!}{
\begin{tcolorbox}[colframe=black, colback=lightgray!20, coltitle=black, sharp corners=south, boxrule=0.3mm, width=16cm, enlarge left by=0mm, enlarge right by=0mm, top=2mm, bottom=3mm]
\textsc{Sample questions, respective answers from the forget set of \lpf}

\begin{tcolorbox}[colframe=black, colback=lightblue!10, sharp corners=all, boxrule=0.3mm, top=1mm, bottom=1mm, left=1mm, right=1mm, width=\linewidth]
\textbf{Question:}\colorbox{lightgray}{After how many seconds of flight did the Space Shuttle Challenger break apart?}\\
\textbf{Answer:} \colorbox{lightblue}{73s}\\
\end{tcolorbox}
\begin{tcolorbox}[colframe=black, colback=lightblue!10, sharp corners=all, boxrule=0.3mm, top=1mm, bottom=1mm, left=1mm, right=1mm, width=\linewidth]
\textbf{Question:}\colorbox{lightgray}{Who was the first person executed in the Salem Witch Trials?}\\
\textbf{Answer:} \colorbox{lightblue}{Bridget Bishop}\\
\end{tcolorbox}
\begin{tcolorbox}[colframe=black, colback=lightblue!10, sharp corners=all, boxrule=0.3mm, top=1mm, bottom=1mm, left=1mm, right=1mm, width=\linewidth]
\textbf{Question:}\colorbox{lightgray}{Which specific volcanic mountain exploded to cause the 1883 Krakatoa Eruption?}\\
\textbf{Answer:} \colorbox{lightblue}{Perboewatan}\\
\end{tcolorbox}

\begin{tcolorbox}[colframe=black, colback=lightblue!10, sharp corners=all, boxrule=0.3mm, top=1mm, bottom=1mm, left=1mm, right=1mm, width=\linewidth]
\textbf{Question:}\colorbox{lightgray}{Which international agreement influenced Iceland's eventual 200-mile fishing limit?}\\
\textbf{Answer:} \colorbox{lightblue}{United Nations Convention on the Law of the Sea (UNCLOS)}\\
\end{tcolorbox}

\begin{tcolorbox}[colframe=black, colback=lightblue!10, sharp corners=all, boxrule=0.3mm, top=1mm, bottom=1mm, left=1mm, right=1mm, width=\linewidth]
\textbf{Question:}\colorbox{lightgray}{Which battle marked the end of Tang military expansion into Central Asia?}\\
\textbf{Answer:} \colorbox{lightblue}{Battle of Talas}\\
\end{tcolorbox}
\end{tcolorbox}
}
\caption{\textbf{Sample questions from the \lpf forget set.} The questions come from one of the five topics described in detail in~\cref{app:dataset-details}.}
    \label{fig:LPF-samples}
\end{figure}

\begin{figure}[t]
    \centering
    \resizebox{\textwidth}{!}{
\begin{tcolorbox}[colframe=black, colback=lightgray!20, coltitle=black, sharp corners=south, boxrule=0.3mm, width=16cm, enlarge left by=0mm, enlarge right by=0mm, top=2mm, bottom=3mm]
\textsc{Sample dichotomous questions from the forget set of \tofu, the correct, plausible answer: YES}

\begin{tcolorbox}[colframe=black, colback=lightblue!10, sharp corners=all, boxrule=0.3mm, top=1mm, bottom=1mm, left=1mm, right=1mm, width=\linewidth]
\textbf{Question:}\colorbox{lightgray}{Has Takashi Nakamura received international recognition for his works?
}\\
\vspace{1mm}
\textbf{Question:}\colorbox{lightgray}{Are Kalkidan Abera's books available in other languages?
}\\
\vspace{1mm}
\textbf{Question:}\colorbox{lightgray}{Does Aysha Al-Hashim have any book series in her portfolio?
}\\
\vspace{1mm}
\textbf{Question:}\graybox{0.85\textwidth}{Are Edward Patrick Sullivan's novels, 'Nell: A Tale of Emerald Isle' and 'In Night's Silence, the Stars Will Be Our Lamps' reflective of his Irish genre preference?}\\
\end{tcolorbox}
\end{tcolorbox}
}
\caption{\textbf{Sample dichotomous questions from the \tofu forget set.} Selected dichotomous questions from the \tofu forget set, where a binary Yes/No answer suffices, making it fairly easy for a LLM to guess without reflecting true unlearning quality.}
    \label{fig:tofu-samples}
\end{figure}

\begin{figure}[htbp]
    \centering
    \resizebox{\textwidth}{!}{

\begin{tcolorbox}[colframe=black, colback=lightgray!20, coltitle=black, sharp corners=south, boxrule=0.3mm, width=16cm, enlarge left by=0mm, enlarge right by=0mm, top=2mm, bottom=3mm]

\textsc{Template for generating Forget set questions for \lpf dataset}

\begin{tcolorbox}[colframe=black, colback=lightblue!10, sharp corners=all, boxrule=0.3mm, top=1mm, bottom=1mm, left=1mm, right=1mm, width=\linewidth]
... \textit{(list of accepted forget questions about {\{forget-concept\}})}\\...\\... \\
\{Create/Add\} 15 highly specific question-answer pairs about the \textcolor{blue}{\{forget-concept\}} \{to the list\}. The questions and answers should be self-contained and not need any reference, i.e. every question should clearly indicate that it is about the \textcolor{blue}{\{forget-concept\}}. The answer should be short, either one word, or at most a few.:
\end{tcolorbox}
\vspace{2mm} 
\textsc{Template for generating Retain set questions for \lpf dataset}

\begin{tcolorbox}[colframe=black, colback=lightblue!10, sharp corners=all, boxrule=0.3mm, top=1mm, bottom=1mm, left=1mm, right=1mm, width=\linewidth]
... \textit{(list of accepted forget questions)}\\...\\... \\
Instead of the \textcolor{blue}{\{forget-concept\}}, create 20 highly specific question-answer pairs about the \textcolor{blue}{\{retain-concept-related-to-forget-concept\}} that are similar in style to the ones in the list above, but are {NOT} about the \textcolor{blue}{\{forget-concept\}}. The questions and answers should be self-contained and not need any reference, i.e. every question should clearly indicate that it is about \textcolor{blue}{\{retain-concept-related-to-forget-concept\}}. The answer should be short, either one word, or at most a few.
\end{tcolorbox}
\end{tcolorbox}
}
    \caption{\textbf{Query templates used to generate \lpf sets.} The following queries were used to generate the forget and retain set queries for the \lpf dataset.}
    \label{fig:lkw-templates}
\end{figure}

\subsection{Generation of paraphrases}
\label{app:para-llm}
As an important part of our proposed evaluation is creating diverse paraphrases of test queries, we use three different LLMs for this purpose. Specifically, we use \qwen~\citep{qwen25}, \phipara~\citep{abdin2024phi} and \mistral~\citep{jiang2023mistral7b} models to generate 5 paraphrases for each forget set question in \lpf using the template in~\cref{fig:evaljudge-phi}. For evaluation on the 400-question retain training set of \lpf, we generate 3 paraphrases from each model, giving nine additional paraphrases per question. For the 100-question held-out retain-test set, we use six additional paraphrases per question. Including the original question, these evaluations comprise 4,000 and 700 responses, respectively. Different to the evaluation paraphrases, we generate train time paraphrases for \lpf using the phi-4-mini-instruct model. This makes our test-time paraphrases disjoint of the ones used for training.

\begin{figure}[!hb]
    \centering
        \resizebox{\textwidth}{!}{
    \begin{tcolorbox}[colframe=black, colback=lightgray!20, coltitle=black, sharp corners=south, boxrule=0.3mm, width=16cm, enlarge left by=0mm, enlarge right by=0mm, top=2mm, bottom=3mm]
        \textsc{Template for generating paraphrased queries}
    \begin{tcolorbox}[colframe=black, colback=lightgray!20, coltitle=black, sharp corners=south, boxrule=0.3mm, width=15cm, enlarge left by=0mm, enlarge right by=0mm, top=2mm, bottom=3mm]{"role": "system", "content": "You are a helpful AI assistant."
        "role": "user", "content": "You are a good paraphraser. I will give you a sentence which is a question, I need you to paraphrase it for me. Generate {\textcolor{blue}{N}} grammatically correct and unique paraphrases. Make sure the output are questions again. Make sure the meaning of paraphrases remains the same as original question and that no new information is added. The output should be an enumerated list of questions. 
            Question: \{\}"}
\end{tcolorbox}
\end{tcolorbox}}

\caption{\textbf{Paraphrased query generation template.} We use this template to paraphrase questions from both \lpf and \rwku datasets. The same template is used to query \mistral, \phipara, and \qwen models with different values of {\textcolor{blue}{N}} for forget and retain sets.}
\label{fig:evaljudge-phi}
\end{figure}

For \rwku, since there is an ``adversarial'' (AA) set already that has some paraphrases, we only generate 3 paraphrases from each model for both the FB and QA subsets of their Forget set. Since paraphrasing is not crucial for neighbor/retain sets (we are doing an average case evaluation in this case), we do not paraphrase for the neighbor set and instead use the original FB and QA from \rwku. We show the effectiveness of using the paraphrased queries instead of the original ones for \rwku in~\cref{fig:answer_comparison}. We see that for some queries, the paraphrased query elicits the correct response from the LLM previously not possible with the original question, highlighting the effectiveness of paraphrased queries.

\begin{figure}[htbp]
    \centering
    \resizebox{\textwidth}{!}{

\begin{tcolorbox}[colframe=black, colback=lightgray!20, coltitle=black, sharp corners=south, boxrule=0.3mm, width=16cm, enlarge left by=0mm, enlarge right by=0mm, top=2mm, bottom=3mm]

\textsc{Sample questions, paraphrases and respective answers from the forget set of \rwku}

\begin{tcolorbox}[colframe=black, colback=lightblue!10, sharp corners=all, boxrule=0.3mm, top=1mm, bottom=1mm, left=1mm, right=1mm, width=\linewidth]
\textbf{Original Question (OQ):}\colorbox{lightgray}{Bruce Lee was born on November 27, \underline{\hspace{2cm}} }\\
\textbf{Ground Truth (GT) Answer:} \colorbox{lightgray}{1940}\\
\textbf{Answer to OQ:} \colorbox{lightred}{1882 Explanation: Bruce Lee was born on November 27, 1882, in the Lee family}\\
\textbf{Paraphrased Question (PQ):} \colorbox{lightblue}{When was Bruce Lee born, on November 27, \underline{\hspace{2cm}}?}\\
\textbf{Answer to PQ:} \colorbox{lightgreen}{November 27, 1940. Bruce Lee was born on November 27, 1940.}
\end{tcolorbox}
\vspace{2mm} 

\begin{tcolorbox}[colframe=black, colback=lightblue!10, sharp corners=all, boxrule=0.3mm, top=1mm, bottom=1mm, left=1mm, right=1mm, width=\linewidth]
 \textbf{Original Question (OQ):} \graybox{0.7\textwidth}{Bruce Lee starred in the movie Enter the Dragon which was released in \underline{\hspace{2cm}} ?}\\
\textbf{Ground Truth (GT) Answer:} \colorbox{lightgray}{1973}\\
\textbf{Answer to OQ:} \gbox{0.7\textwidth}{1973 Explanation: Enter the Dragon, featuring Bruce Lee, was released in 1973. It is a}\\
\textbf{Paraphrased Question (PQ):} \bbox{0.7\textwidth}{
What year was the film Enter the Dragon released, featuring Bruce Lee?}\\
\textbf{Answer to PQ:} \gbox{0.7\textwidth}{Enter the Dragon was released on July 26, 1973. It starred Bruce Lee, although he passed away before the movie}
\end{tcolorbox}

\vspace{2mm} 

\begin{tcolorbox}[colframe=black, colback=lightblue!10, sharp corners=all, boxrule=0.3mm, top=1mm, bottom=1mm, left=1mm, right=1mm, width=\linewidth]
 \textbf{Original Question (OQ):} \graybox{0.7\textwidth}{One of Bruce Lee's early martial arts experiences was training in \underline{\hspace{2cm}} under Ip Man.}\\
\textbf{Ground Truth (GT) Answer:} \colorbox{lightgray}{Wing Chun}\\
\textbf{Answer to OQ:} \rbox{0.7\textwidth}{One of Bruce Lee's early martial arts experiences was training in martial arts, specifically in the martial art of Chinese martial arts,}\\
\textbf{Paraphrased Question (PQ):} \bbox{0.7\textwidth}{In what martial art did Bruce Lee train under Ip Man's tutelage during his formative years?}\\
\textbf{Answer to PQ:} \gbox{0.7\textwidth}{Bruce Lee studied the martial art discipline of Wing Chun intently under Ip Man's guidance in his early career.}
\end{tcolorbox}

\end{tcolorbox}
}
    \caption{\textbf{Sample questions with our generated paraphrases for the \rwku FB forget set where the paraphrased question gets the LLM to output the correct answer.} The original questions are paraphrased either with \mistral, \qwen, or \phipara. Colored boxes depict: \colorbox{lightblue}{paraphrased question}, \colorbox{lightgreen}{correct answer w.r.t GT}, and \colorbox{lightred}{answer incorrect w.r.t GT}.}
    \label{fig:answer_comparison}
\end{figure}

\subsection{Correctness of the \llmj}
\label{app:judge-correct}

For all \llmj based evaluations we use ~\gemini\footnote{Model: gemini-2.5-flash-preview-09-2025}, which we found particularly effective.
Given the question, the LLM's output and the ground-truth answer, we query the \llmj to solicit a Yes/No response. The model should respond \textit{Yes} when the LLM output is equivalent to the ground-truth given the question at hand, and \textit{No} otherwise. 

Since the \llmj is an LLM, always controlling its response is hard and sometimes it responds with something other than Yes/No, for the template in~\cref{fig:judge-template}. Other times, the call to \gemini{} API is unsuccessful. For \rwku across 5 models this total error rate is 1.2\%$\pm$0.4 for the retain set and 1.1\%$\pm$0.2 for the forget set on average. Hence, for all \rwku evaluations we remove these unique 1.5\% samples from both the retain and forget sets. Judge's reasoning and scoring are visualized for some methods in~\cref{fig:judge-reasoning-LPF}.

To assess the reliability of our \llmj, we conducted a human study on the \lpf dataset. Six evaluators were shown the question, ground-truth answer, LLM output, and the judge's verdict, and asked whether the verdict was correct. 

\begin{wrapfigure}{r}{0.5\textwidth}
\centering
\vspace{-5mm}
\includegraphics[width=.9\linewidth]{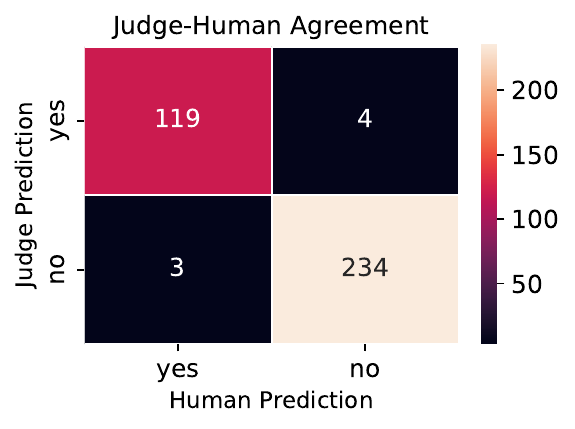}
\caption{\textbf{\llmj is highly aligned with human evaluators.} According to 6 human evaluators, for all methods from~\cref{tab:LPF-base} on random queries from the forget set, the \llmj shows high agreement with humans.}
\label{fig:judge-confusion}
\vspace{-10mm}
\end{wrapfigure}
Across 360 samples, the resulting confusion matrix (\cref{fig:judge-confusion}) shows that the \llmj is well-aligned with human judgments: \nds{annotators reached unanimous agreement on 96.7\% of samples (all six giving the same verdict), with an average pairwise agreement of 97.8\%.}

\section{Experimental Details}
\label{app:experimental-details}
\subsection{
Models and Compute}
For all unlearning experiments on \lpf, we use predominantly the \llama~\citep{grattafiori2024llama} model. For the best methods, we also unlearn the \phirw~\citep{abdin2024phi}  and Llama-3.1-8B-Instruct models. For \rwku, we use the \phirw~\citep{abdin2024phi} from their original setup. 
To generate the training time paraphrases used for \lpf, we use the Phi-4-Mini-Instruct~\citep{abouelenin2025phi} model.  

\textbf{Compute requirements.} All experiments were conducted on an internal node of Nvidia A100 40G GPUs. A single \ours unlearning run on \lpf takes approximately 15 minutes on 2$\times$NVIDIA A100 (40GB) GPUs, while \rwku runs take 45 to 60 minutes due to the larger forget and retain set sizes. Total compute for the main \lpf experiments (\cref{tab:LPF-base}, sweeping all baselines and hyperparameters) was approximately 70-90 GPU-hours; the corresponding \rwku experiments consumed roughly 60 GPU-hours. Additional GPU time was spent on the relearning experiments (\cref{tab:benign-relearning-improved}), and the larger \texttt{Llama-3.1-8B-Instruct} runs. We use mixed-precision (bf16) training and gradient accumulation throughout. All experiments fit comfortably within a single 8-GPU node.
 
\textbf{Computational overhead.}
Computing the retain loss requires the full vocabulary distribution of a frozen reference model, so a naive implementation of \ours holds two copies of the LLM in GPU memory during fine-tuning, which can hinder scaling to large models. However, since the reference distribution is fixed, the reference logits over the retain set can be pre-computed once and cached on disk, after which fine-tuning only requires the active model in memory. The overhead is thus a one-time pre-computation cost rather than a persistent memory burden, keeping \ours practical at larger scales.

\subsection{\lpf experiments}
\label{app:LPF-exp-details}
We build on top of the code-base from~\cite{openunlearning2025} for \lpf experiments and the base unlearning duration of 10 epochs is chosen from there. For~\cref{tab:LPF-base}, we train with 5 paraphrases for 10 epochs. The training-time paraphrases are generated with the same prompt (\cref{fig:evaljudge-phi}) as used for test-time paraphrases but with Phi-4-mini-instruct model. In this way we ensure that test-time paraphrases are disjoint of the ones seen during training. The baseline methods cover all types of unlearning algorithms including \ga, \gd, preference optimization based (NPO, \simnpo) and layer-wise editing (RMU) and (KL-Div). We train all methods with per GPU effective batch-size of 16, AdamW~\citep{loshchilovdecoupled} optimizer, weight decay 
of 1e-2, cosine schedule peaking at 10\% of total steps. We also test unlearning without any paraphrases for 60 epochs (\cref{tab:LPF-longer-train}).

Specific parameters used for each unlearning method are listed in~\cref{tab:train-config}. The initial LR search (\cref{tab:LPF-lr-ablation}) served as the starting point for the subsequent $\lambda_\mathcal{R}$ sweep (\cref{tab:LPF-tradeoff}), guiding the choice of settings for the unlearning quality-utility tradeoff. The default values of $\lambda_\mathcal{F}$ for each method were taken from~\citep{openunlearning2025}. For evaluation we report the worst-case $\mathcal{J}_W$ and average-case  $\mathcal{J}_{Avg}$ \llmj accuracy for the forget and retain sets respectively. Since the ground-truth answers for \lpf are either one word or short phrases, we restrict the output length of the LLM at evaluation time to a maximum of 50 tokens.

\subsection{\rwku experiments}
\label{app:rwku-exp-details}

For \rwku, we adapt the original code-base\footnote{\href{https://github.com/jinzhuoran/RWKU}{https://github.com/jinzhuoran/RWKU}} and use the \phirw model. \rwku has 100 forget targets (famous people that the pre-trained LLM already knows about), and for each target the forget set consists of several paragraph based descriptions, unlike the QA based for \lpf. Since each target has several of these paragraphs, there is a lot of paraphrased text for each target already in the respective forget sets. Hence, for \rwku, we unlearn with the batch-setting on 10 targets for 5 and 10 epochs without any further paraphrasing.  All methods were fine-tuned with AdamW optimizer, with a cosine schedule peaking at 20 steps, the same setup as in the original code-base. Also at inference, all parameters like temperature, sampling, number of output tokens etc., are set to the default values from \rwku.

The evaluation RWKU retain sets, which are QA/FB type queries, cannot be used directly during training. This is due to a data type mismatch: the training data (the forget set) consists of paragraphs, while the evaluation data (the retain set) is composed of QA/FB queries. This mismatch also means that the two losses in \ours would operate on different output token lengths. This could specifically be problematic for methods like \simnpo, \gd and \ours. For methods like ICU, DPO, NPO, \rwku has pre-defined retain set templates that are used as $\mathcal{D_R}$. Hence, for \simnpo, \gd and \ours, we define a train-time retain set ($\mathcal{D_R}$) by combining text from 10 targets disjoint of the forget set. This means that the retain set at train-time is not the same as the default one used by \rwku for evaluation. 

As baselines we take all non-LORA unlearning methods from the original work, and the results are in~\cref{tab:rwku-base}. Specific parameters used for each unlearning method are listed in~\cref{tab:train-config}. We use the same loss coefficients as for LPF, except for SimNPO (\(\lambda_R=0.125\), \(\lambda_F=1\)) and JensUn (\(\lambda_R=\lambda_F=1\)); see Table 5. For methods like ICU, RT we use the default parameters from \rwku. 
\citep{jinrwku} also use MIA attacks and other utility based metrics, and these can be found in~\cref{tab:rwku-big} along with optimal LR selection.
We also scale the best unlearning methods from the 5 epoch setup to 10 epochs in~\cref{tab:rwku-long}.

\paragraph{Twenty-target batch setting.}
\label{app:rwku-20-target}
We repeat the \rwku batch experiment with 20 forget targets. We use the 10-target \ours hyperparameters without retuning. The larger forget set makes unlearning harder, especially on QA, while retain and general utility remain comparable.
\begin{table}[H]
\centering
\small
\caption{\textbf{Scaling \ours from 10 to 20 \rwku targets without retuning.} Forget and retain results are worst-case and average accuracy, respectively.}
\label{tab:rwku-20-target}

\begin{tabular}{L{30mm}C{10mm}|C{8mm}C{8mm}|C{8mm}C{8mm}|C{7mm}C{7mm}C{7mm}}
\toprule
& & \multicolumn{2}{c|}{Forget ($\downarrow$)} & \multicolumn{2}{c|}{Retain ($\uparrow$)} & \multicolumn{3}{c}{Utility ($\uparrow$)} \\
Model & Targets & FB & QA & FB & QA & Gen & Rep. & WR \\
\midrule
Phi-3-Mini-4K & 10 & 91.0 & 78.6 & 59.6 & 60.8 & 63.4 & 708 & 0.50 \\
\ours & 10 & 16.3 & 6.1 & 40.8 & 42.4 & 63.2 & 694 & 0.52 \\
\midrule
Phi-3-Mini-4K & 20 & 92.4 & 82.5 & 61.2 & 65.6 & 63.4 & 708 & 0.50 \\
\ours & 20 & 20.4 & 16.9 & 39.2 & 52.4 & 63.4 & 680 & 0.54 \\
\bottomrule
\end{tabular}
\end{table}
\begin{table}[t]
\centering

\small
\caption{\textbf{%
Training and data configurations.} Final values of training parameters like loss coefficients for~\cref{eq:unlearn-obj}, LR, BS (per GPU batch-size), and GradAc (gradient accumulation steps). The loss coeff. values were selected after an ablation on the \lpf dataset (\cref{tab:LPF-tradeoff}). For LR the ablations can be found in Tables~\ref{tab:LPF-lr-ablation} and~\ref{tab:rwku-big}. All \lpf models were trained across 2 GPUs, and \rwku ones across 3 GPUs.}
\label{tab:train-config}
\tabcolsep=1.1pt
\extrarowheight=2pt
\begin{tabular}{L{25mm} | C{9mm}C{8mm} C{8mm} C{10mm} C{12mm}| C{9mm}C{8mm} C{8mm} C{10mm} C{10mm}}
\toprule
&  \multicolumn{5}{c|}{\textbf{\lpf}} &  \multicolumn{5}{c}{\textbf{\rwku}}  \\
\cline{2-6}\cline{7-11}
Method& LR & $\lambda_\mathcal{R}$    & $\lambda_\mathcal{F}$ & BS&  GradAc &LR & $\lambda_\mathcal{R}$ & $\lambda_\mathcal{F}$ & BS&  GradAc\\
\toprule
GradAscent & 8e-6 & 0.0 & 1.0 & 4& 4& 7e-8 & 0.0 & 1.0 & 4& 2\\
GradDiff & 1e-5 & 0.5 & 0.5 & 4& 4 & 6e-7 & 0.5 & 0.5 & 4& 2\\
DPO & 1e-5 & 1.0 & 1.0& 4& 4& 1e-5 & 1.0 & 1.0& 4& 2\\
NPO & 9e-6 & 1.0 & 1.0& 4& 4& 1e-5 & 1.0 & 1.0& 4& 2\\
SimNPO & 2e-5 & 1.0 & 0.125 & 4& 4& 1e-5 & 0.125 & 1.0 & 4& 2\\
RMU & 7e-5 & 0.5 & 0.5 & 4& 4& -- & -- & -- & -- & -- \\
KL-Div & 3e-6 & 0.5 & 0.5 & 4& 4& -- & -- & -- & -- & -- \\
\ours& 1e-5 & 0.5 & 1.0 & 4& 4& 8e-7 & 1 & 1 & 4& 2\\
\bottomrule
\end{tabular} 
\end{table}

\textbf{Hyperparameter selection for \rwku experiments.} Although \rwku~\citep{jinrwku} did a large-scale hyper-parameter optimization for different unlearning methods, we found some of these did not translate well to the batch-setting that we use. Moreover, important baselines like \gd were missing from the \rwku benchmark. For $\lambda_\mathcal{F}$ and $\lambda_\mathcal{R}$ in~\cref{eq:unlearn-obj}, we use the dataset-specific values in~\cref{tab:train-config}, with the SimNPO and \ours exceptions described above. In~\cref{tab:rwku-big}, for all unlearning methods, we did a small search for the optimal LR. The final selected value for each method is highlighted. In general, the selection is done based on the optimal forget-neighbor (retain) tradeoff. 

For \ga, the two learning rates illustrate a sharp forgetting–utility trade-off. At LR \(3\times10^{-8}\), substantial forget-set knowledge remains. Increasing the LR to \(7\times10^{-8}\) reduces worst-case forget accuracy to 4.3\% on FB and 2.3\% on QA, but severely degrades retain performance and fluency. Both for DPO and NPO, the improvement in forget set accuracy is slower than \gd on increasing the LR, and the decay in retain also comes into play, hence we select a \texttt{LR=1e-5} for both. A similar trend follows for \simnpo, where \texttt{LR=1e-5} is selected. For RT and ICU, since there is no dependence on retain set at training time, we keep the original values from~\citep{jinrwku}. Finally, for \ours, out of the tested LRs, \texttt{LR=8e-7} is the most optimal in terms of unlearning-utility tradeoff.

In~\cref{tab:rwku-long}, we double the number of training epochs \mhc{from 5 to 10} for the best methods from~\cref{tab:rwku-base}. \mhc{For most} methods, we see improvements (lower) in forget set accuracies with a small decay in retain set performance. The general utility of all methods is more-or-less the same as for 5 epochs unlearning. In this setup as well, \ours attains the best unlearning quality-utility tradeoff.

\begin{figure}[!b]
    \centering
        \resizebox{\textwidth}{!}{
    \begin{tcolorbox}[colframe=black, colback=lightgray!20, coltitle=black, sharp corners=south, boxrule=0.3mm, width=16cm, enlarge left by=0mm, enlarge right by=0mm, top=2mm, bottom=3mm]
        \textsc{\llmj prompt for the Win Rate evaluations}
    \begin{tcolorbox}[colframe=black, colback=lightgray!20, coltitle=black, sharp corners=south, boxrule=0.3mm, width=15cm, enlarge left by=0mm, enlarge right by=0mm, top=2mm, bottom=3mm]{"role": "system", "content": "You are a helpful and precise assistant for checking the quality of the answer. Your response MUST be a JSON object.[Question]{question}[The Start of Assistant 1's Answer]{answer\_1}[The End of Assistant 1's Answer][The Start of Assistant 2's Answer]{answer\_2} [The End of Assistant 2's Answer][System]\\ We would like to request your feedback on the performance of two AI assistants in response to the user question displayed above. Please rate the helpfulness, relevance, accuracy, level of details of their responses. Each assistant receives an overall score on a scale of 1 to 10, where a higher score indicates better overall performance. Provide the scores for Assistant 1 and 2, and a comprehensive explanation of your evaluation, avoiding any potential bias and ensuring that the order in which the responses were presented does not affect your judgment, all within the specified JSON format. 
     Question: \{\}"}
\end{tcolorbox}
\end{tcolorbox}}

\caption{\textbf{\llmj prompt template for Win Rate evaluation for the AlpacaEval instruction based generation task.} We use this template to rate comparative responses from the base and the unlearnt model. }
\label{fig:winrate-template}
\end{figure}

\subsection{LLM utility evaluations}
\label{app:utility-eval}

For evaluating the unlearned model's general LLM related utility, we use accuracy on a $5k$ subset of MMLU as a measure of general ability. To test the repetitiveness of the generated outputs we use $1k$ instruction based generated queries from AlpacaEval, same as~\cite{jinrwku}, and report the entropy score originally used by~\cite{meng2022locating}. Evaluating repetitiveness is important as some unlearning methods suffer from generating token repetitions often making the responses incoherent, see Figures~\ref{fig:examples-output-LPF}-\ref{fig:examples-output-rwku}. 

Ideally, the unlearnt model should be as close as possible to the original base model, except for the forget set.
Therefore, to measure the model's \textit{response quality} in terms of relevancy, helpfulness, level of details and accuracy, we compare the output of unlearned and original  model, and report the Win Rate of the former according to an \llmj.
The template used for the semantic judge is shown in~\cref{fig:winrate-template}, adapted from~\cite{zhao2024long}. 

\textbf{Note.} For the results reported in Tables~\ref{tab:LPF-base},~\ref{tab:rwku-base},~\ref{tab:benign-relearning-improved} and~\ref{tab:LPF-phi}, the Win Rate (WR) is     
  computed over 300 samples from AlpacaEval; for all other WR evaluations in this work, we use 100 samples. This choice reflects
   the substantial compute and API cost of \llmj judging. For the retain-set average-accuracy results in Tables~\ref{tab:LPF-base},~\ref{tab:LPF-longer-train},~\ref{tab:LPF-phi}, and~\ref{tab:llama-LPF} (main LPF
   experiments), we evaluate on the held-out test set of 100 distinct questions, using each original question plus six additional paraphrases (700 responses in total). For the remaining LPF
   experiments: specifically, hyper-parameter selection and relearning, we instead use the 400-question retain training set
  (each original question plus nine additional paraphrases, for 4,000 responses in total). 

By construction of our prompt and the comparison to the original model, \textit{response quality} already measures reasoning and truthfulness of unlearnt models. Hence, we omit 
similar metrics from~\cite{jinrwku} based on  Big-Bench-Hard (BBH)~\citep{suzgun2023challenging} and TruthfulQA~\citep{lin2022truthfulqa}. Similarly, we omit the evaluation via MIA as we consider it less reliable than other metrics (e.g., the MIA based on the Negative log-likelihood of the desired answers are not invariant to output rescaling, and may again vary depending on the formulation of semantically equivalent answers). For completeness, we still present the original \rwku utility metrics in~\cref{tab:rwku-big}. For all these tasks, we use the default system prompt of the respective models, similar to~\cite{jinrwku}.

\subsection{Relearning experiments}
\label{app:relearning}
Our main threat model assumes that the attacker does not possess the exact forget data, since such access changes the membership and privacy setting. We therefore use disjoint real-world facts and the retain set as realistic recovery data. We also report a stronger diagnostic with partial forget-set access. Following the benign \mhc{relearning} setup from~\cite{hu2024jogging}, we relearn \lpf unlearnt models on well-known facts across several domains. Specifically, we test relearning for two setups.
\begin{enumerate}[leftmargin=5mm]
\item \textbf{Real-knowledge set.} This relearning set is disjoint of both the \lpf forget and retain sets.  Specifically, we collect 200 QA pairs using the~\mistral model from topics like \textit{history, geography, biology, sports, etc}. 
\item \textbf{\lpf retain set.} To simulate the attacker having access to some form of retain set, we take the non-paraphrased retain set of \lpf which comprises of 400 distinct question-answer pairs. This is our adversarial relearning set.
\end{enumerate}
Then, we fine-tune several unlearnt models with the cross-entropy loss w.r.t. the ground truth for 600 update steps (selected via a search over number of update steps)
with effective \texttt{BS=16} and \texttt{LR=1e-5}. We want to emphasize here that, as we are only concerned with testing for the strongest possible benign relearning, the setup of training steps and LR chosen does not care about preserving the model's utility. The  relearning results are presented in~\cref{tab:benign-relearning-improved}. 

We also tested an additional baseline unlearning method, NPO+SAM~\citep{fantowards}, which aims to prevent benign relearning. From the original code-base,\footnote{\href{https://github.com/OPTML-Group/Unlearn-Smooth/tree/main/MUSE}{https://github.com/OPTML-Group/Unlearn-Smooth}} we use the MUSE setup and adapt it for \lpf with paraphrases. We train for the various unlearning steps in~\cref{tab:benign-relearning-improved} using the default \texttt{LR=1e-5} and SAM coefficient set to 0.01. We did a small grid search over the retain loss coefficient ([0.1, 0.5, 1.0, 1.5, 2.5]) for the 200 step unlearning regime, and found that the  value of $0.1$ leads to lowest $\mathcal{J}_W$ (forget set accuracy).  For NPO+SAM, there are additional SAM update steps that are compute intensive, and even though we did a small hyper-parameter search, we could not make it work on the \lpf dataset. For the $400$ step unlearnt model via NPO+SAM, we attain a WR of 0.1 with $\mathcal{J}_W$ of 15\%. On relearning this model, $\mathcal{J}_W$ went up to 58\%, indicating the method is still susceptible to benign relearning.

\paragraph{Stronger recovery attacks.}
\label{app:stronger-recovery}
We give the attacker direct access to subsets of the forget set. When relearning on two of five \lpf topics, held-out accuracy on the other three is $3/60$ after 200 relearning steps, $8/60$ after 600 steps, and $9/60$ after 1,000 steps. Within the Salem topic, the unlearned model starts at $0/20$. Relearning on 2, 5, or 10 questions yields held-out accuracies of $0/18$, $1/15$, and $1/10$, respectively, after 1,000 steps. The attack therefore recovers little knowledge beyond the supplied subset.

We also quantize the 400-step \ours model after unlearning based on the quantization relearning attack from~\citep{zhang2025quantization}. After quantization, the worst-case forget accuracy for \ours is $0.0$ under both 8-bit and 4-bit precision.

\begin{table}[H]
\centering \small
\extrarowheight=2pt
\caption{\textbf{Sensitivity of different \rouge based scores to word order and content.} For the commonly used Recall (R), Precision (P) and F1-Score (F1) based on \rouge-L\protect\footnotemark, we show how brittle the scores are to slight changes in word order and content.}
\vspace{0mm}
\begin{tabular}{L{60mm}|C{10mm}C{10mm}C{10mm}C{10mm}C{10mm}}
\toprule
\textbf{Reference: }The capital of France is Paris. & \textbf{R} & \textbf{P} & \textbf{F1} & \textbf{Judge} & \textbf{Human} \\
\midrule
\textbf{A1: }Paris is the capital of France. & 0.5 & 0.5 & 0.5 & \DarkGreencheck & \DarkGreencheck\\ 
\textbf{A2: }Of France, Paris is the capital. & 0.17 & 0.17 & 0.17 & \DarkGreencheck&\DarkGreencheck\\ 
\textbf{A3: }The capital of France is Marseille. & 0.83 & 0.83 & 0.83 & \DarkRedcross & \DarkRedcross\\ 
\bottomrule
\end{tabular}
\label{tab:rouge_sensitivity}
\end{table}
\footnotetext{Evaluated using the commonly used (e.g. by \rwku) \url{https://pypi.org/project/rouge}}
\begin{table}[ht]
    \centering
    \caption{\textbf{Testing different styles of evaluations in our worst-case setup.} For the 60 epoch setup from~\cref{tab:LPF-longer-train} on the \lpf dataset, we show adding additional query types like Fill-in-Blank ( $\mathcal{J}_{FB}$) and adding hints to the query ($\mathcal{J}_{Ht}$) does not help in enhancing our chosen worst-case evaluation 
 $\mathcal{J}_W (\max_{(1,2)})$.}
 \small
\begin{tabular}
{L{23mm}|C{7mm}C{10mm}C{9mm}C{9mm} |C{10mm} | C{12mm}| C{12mm} |C{10mm}}
    \toprule
    Method & $\mathcal{J}_P$(1) & $\mathcal{J}_{ICR}$(2)  & $\mathcal{J}_{Ht}$(3) & $\mathcal{J}_{FB}$(4) & $\max_{(1,2)}$ & $\max_{(1,2,3)}$ & $\max_{(1,2,4)}$ & $\max_{All}$\\
 \toprule
Llama-3.2-3B  & 71.0 & 72.0 & 71.0 & 65.0 & 76.0 & 76.0  & 76.0 & 76.0\\
GradAscent & 0.0 & 0.0 & 0.0 & 0.0 & 0.0 & 0.0 & 0.0 & 0.0\\
GradDiff & 0.0 & 0.0 & 0.0 & 0.0 & 0.0 & 0.0 & 0.0 & 0.0\\
NPO & 1.0 & 2.0 & 1.0 & 1.0 & 3.0 & 3.0 & 4.0 & 4.0\\
RMU  & 14.0 & 16.0 & 13.0 & 14.0 & 19.0 &  19.0 & 19.0 & 19.0 \\
SimNPO  & 27.0 & 26.0 & 23.0 & 27.0 & 29.0 & 29.0 & 29.0 &  29.0\\
\ours & 0.0 & 0.0 & 0.0 & 0.0 & 0.0 & 0.0 & 0.0 & 0.0\\
\bottomrule
\end{tabular}
 \label{tab:worse-LPF}
\end{table}
\begin{table}[t]
\centering
\caption{\textbf{Switching from \rouge to $\mathcal{J}_W$ changes the ranking of methods.} We show the ranking change for the FB and QA sets from \rwku on transitioning from \rouge to worst-case accuracy by \llmj ($\mathcal{J}_W$) as a metric. Colored number indicates the relative change in rank. GradAscent values are for LR = $3\times10^{-8}$ from~\cref{tab:rwku-big}.}
\label{tab:rank_shift}
\small
\begin{tabular}{l|cccc|cccc}
\toprule    
 & \multicolumn{4}{|c}{Forget FB-set $\downarrow$}  & \multicolumn{4}{|c}{Forget QA-set $\downarrow$}  \\
    \cmidrule{2-5}\cmidrule{6-9}
\textbf{Method} & \rouge & \textbf{Rank} & $\mathcal{J}_W$ & \textbf{Rank} & \rouge & \textbf{Rank} & $\mathcal{J}_W$ & \textbf{Rank}  \\
\toprule
GradAscent & 40.1 & 5 & 73.3 & 5  & 34.6 & 5&  68.7& 5 \\
GradDiff & 4.7 & 2 & 22.3 & 2  & 1.6 & 1 & 22.1 & 2 (\textcolor{red}{+1})\\
DPO & 22.5 & 3 & 48.2 & 3 & 19.6 & 3&42.0 & 3 \\
NPO & 22.5 & 3 & \mhc{55.4} & 4 (\textcolor{red}{+1})& 22.3 & 4&50.4 & 4 \\
RT & 48.5 & 6 &  89.1 & 6  & 46.3 & 6& 74.8 & 6 \\
\ours & 3.1 & 1 &  \mhc{16.3} & 1 & 1.8 & 2& 6.1 & 1 (\textcolor{green}{-1})\\
\bottomrule
\end{tabular}
\end{table}

\begin{figure}[!b]
    \centering
    \includegraphics[width=0.95\linewidth]{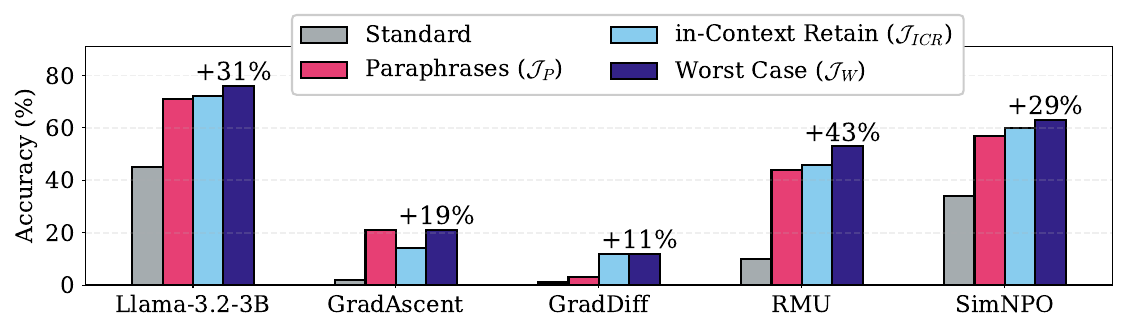} 
    \caption{\textbf{Worst-case over different evaluation methods enhances forget-quality assessment.} In this plot, we unlearn with the respective method for 5 epochs without paraphrases on the \lpf dataset. Then, we show \textit{(a)} standard (single question) forget set accuracy \textit{(b)} worst-case forget set accuracy over 15 paraphrases as evaluated by \llmj, \textit{(c)} the same with random retain set questions as part of the in-context samples \textit{(d)} the point-wise worst-case accuracy over \textit{(b)} 
 and \textit{(c)}. Across all unlearning methods and the original model (\llama), worst-case over the two evaluations shows significant increase in forget set accuracy (denoted by +x\%), making it a better measure for evaluating unlearning quality.}
    \label{fig:llm-para-worse}
\end{figure}


\section{Additional Evaluation Experiments}
\label{app:eval-experiments}

\subsection{Worst-case evaluation details}
\label{app:worse-case-evals}
\textbf{Effectiveness of worst-case evaluation.} In~\cref{fig:llm-para-worse}, we report the \textit{Standard} forget set accuracy obtained when evaluating on the forget set without paraphrases for different unlearning baselines (gray bar). Using worst-case over \textit{Paraphrases} of the forget-set questions ($\mathcal{J}_{P}$, red bar) leads to a significant increase in forget set accuracy, indicating that unlearning was significantly less successful than estimated by the \textit{Standard} evaluation. Using worst-case of paraphrases with retain set as \textit{in-context samples} ($\mathcal{J}_{ICR}$, light blue) also increases the forget set accuracy in comparison to standard. On the forget set, we therefore report the sample-wise \textit{Worst-case} ($\mathcal{J}_{W}$) over paraphrases and ICR samples (dark blue bar), to faithfully cover all cases where the model outputs the correct answer. Our improved evaluation reveals that the forget set accuracy can be underestimated by up to 43\% (RMU in~\cref{fig:llm-para-worse}), highlighting the importance of robust evaluation methods.
\begin{figure}[t]
    \centering
    \resizebox{\textwidth}{!}{
\begin{tcolorbox}[colframe=black, colback=lightgray!20, coltitle=black, sharp corners=south, boxrule=0.3mm, width=16cm, enlarge left by=0mm, enlarge right by=0mm, top=2mm, bottom=3mm]
\text{Extended query formulations: Fill-in-Blanks (FB) and Hints (Ht)}
\begin{tcolorbox}[colframe=black, colback=lightblue!10, sharp corners=all, boxrule=0.3mm, top=1mm, bottom=1mm, left=1mm, right=1mm, width=\linewidth]
\textbf{Original Question:}\colorbox{lightgray}{Which specific volcanic mountain exploded to cause the 1883 Krakatoa Eruption?
}\\
\vspace{1mm}
\textbf{FB Question:}\graybox{0.8\textwidth}{The volcanic mountain whose explosion caused the 1883 Krakatoa Eruption was called \underline{\hspace{2cm}}
}\\
\vspace{1mm}
\textbf{Ht Question:}\graybox{0.8\textwidth}{Which specific volcanic mountain exploded to cause the 1883 Krakatoa Eruption? Per}\\
\end{tcolorbox}
\end{tcolorbox}
}
\caption{\textbf{Other forget set query formulations tested for LPF.} For the original question from LPF forget set, we also tried converting the question into a Fill-in-blank style and tried appending the first few tokens of the response as hint (Ht Question) to elicit a response. However, as shown in~\cref{tab:worse-LPF}, this did not lead to an improvement in worst-case accuracy.}
    \label{fig:LPF-extended}
\end{figure}

\textbf{Extended forget query formulations.} 
We explored expanding our forget queries with reformulations like Fill-in-Blank (FB) queries and adding hints (Ht) about the answer, see~\cref{fig:LPF-extended} for examples. As shown in~\cref{tab:worse-LPF}, these changes did not yield a stronger evaluation outcome. Specifically, there was no improvement for any method except for NPO, which saw a 1\% increase in forget set accuracy. This occurred when we moved from a worst-case evaluation over QA and PQ ($\max_{1,2}$) to a worst-case over QA, PQ, FB, and Ht ($\max_{1,2,3,4}$).
Ultimately, since these extended formulations provided no meaningful gains, we decided to use the worst-case over PQ and ICR ($\max_{1,2}$) as our standard evaluation protocol. This approach allows us to reduce calls to the \llmj and save on both compute and inference time.


\nds{\textbf{Other reliability checks.} We ran several checks to confirm our results are reliable. Tables~\ref{tab:acrossprompt} and~\ref{tab:acrossjudge} show that our numbers barely change when we vary the evaluation prompt or swap in a different judge model. The largest shift is just 1 point (for RMU \mhc{and Pretrained}). The judge sees the question, the ground-truth answer, and the model's full response, and checks whether every key piece of information is present, paying close attention to names, dates, and numbers. If a model refuses to answer but still leaks a key entity, the judge marks it as correct. Human annotators verified that the judge's decisions match their own.}

\begin{table}[h]
\centering
\caption{\textbf{Robustness checks for our evaluation pipeline.} Worst-case accuracy ($\downarrow$) varies by at most 1 point across both evaluation prompts (left) and judge models (right), confirming that our results are not artifacts of these choices.}
\begin{subtable}{0.48\linewidth}
\centering
\caption{Worst-case accuracy across evaluation prompts. Variant-1 is our default (used throughout the paper); Variants 2 and 3 are alternative phrasings.}
\label{tab:acrossprompt}
\begin{tabular}{lccc}
\toprule
Eval.\ prompt & Pretrained & RMU & JensUn \\
\midrule
Variant-1 & 76.0 & 12.0 & 0.0 \\
Variant-2, 3   & 76.0 & 12.0 & 0.0 \\
\bottomrule
\end{tabular}
\end{subtable}
\hfill
\begin{subtable}{0.48\linewidth}
\centering
\caption{Worst-case accuracy across judge models. Gemini-2.5-flash is our default (used throughout the paper). Gemini-3.1-flash is an alternative judge.}
\label{tab:acrossjudge}
\begin{tabular}{lccc}
\toprule
Judge-model & Pretrained & RMU & JensUn \\
\midrule
Gemini-2.5 & 76.0 & 12.0 & 0.0 \\
Gemini-3.1 & 77.0 & 13.0 & 0.0 \\
\bottomrule
\end{tabular}
\end{subtable}

\end{table}

\begin{table}[t]
\centering
\small
\caption{\textbf{Even for \rwku benchmark, our new evaluation enhances forget set accuracy estimates.} For the 10-target batch setting for \rwku, we test the FB and QA sets on the original (\phirw) model using \llmj accuracy. We contrast our proposed evaluation against the original \rwku sets. The table below reveals a significant overestimation of unlearning performance in \cite{jinrwku}. This shows the significance of using paraphrases of the original questions ($\mathcal{J}_P$), using retain queries as context ($\mathcal{J}_{ICR}$), as well as the combined worst-case evaluation, (\textcolor{purple}{$\mathcal{J}_W$}) over the resp. original sets and the improvement in the corresponding category \textcolor{purple}{(+x)}. We note that the ``adversarial'' evaluation (AA) of \rwku \cite{jinrwku} using techniques motivated by jailbreak attacks is weaker than our proposed evaluation.}
\label{tab:rwku-worse}
\tabcolsep=5pt
\extrarowheight=-1.5pt
\begin{tabular}{L{18mm}|C{6mm}C{6mm}C{6mm}C{6mm}|C{7mm}C{7mm}C{15mm}|C{7mm}C{7mm}C{15mm}}
\toprule
& \multicolumn{4}{|c}{\rwku Eval.} & \multicolumn{6}{|c}{Proposed Eval.} \\
\cmidrule{2-5} \cmidrule{6-11}
     & FB & QA & AA & All & \multicolumn{3}{|c}{FB} & \multicolumn{3}{|c}{QA}\\ 
     \cmidrule{2-5}\cmidrule{6-8}\cmidrule{9-11}
Method  & & & & & $\mathcal{J}_{P}$ & $\mathcal{J}_{ICR}$  & \textcolor{purple}{$\mathcal{J}_{W}$} &  $\mathcal{J}_{P}$ & $\mathcal{J}_{ICR}$  & \textcolor{purple}{$\mathcal{J}_{W}$} \\
\toprule
Original & 58.4 & 61.1 & 63.8 & 61.9 & 86.1 & 86.7 & \textcolor{purple}{91.0 \tiny{(+32.6)}}  & 74.0 & 76.3 & \textcolor{purple}{78.6 \tiny{(+17.5)}}\\
GradAscent & 44.0 & 40.5& 54.3 & 48.7 & 67.9 & 63.9 &\textcolor{purple}{73.3 \tiny{(+29.3)}}  & 61.1& 64.9 & \textcolor{purple}{68.7 \tiny{(+28.2)}} \\
GradDiff & 4.8 & 0.0 & 12.7& 7.9 & 11.4 & 13.9 & \textcolor{purple}{22.3 \tiny{(+17.5)}} & 11.5 & 12.4 & \textcolor{purple}{22.1 \tiny{(+22.1)}}\\
DPO &  31.9 & 28.2 & 30.0  & 30.1 & 42.0 & 46.4 & \textcolor{purple}{48.2 \tiny{(+16.3)}} & 38.9 & 39.7 & \textcolor{purple}{42.0 \tiny{(+13.8)}} \\
NPO & 33.7 & 24.4 & 35.3 & 32.5 & 49.4 & 50.0 & \textcolor{purple}{55.4 \tiny{(+21.7)}} & 42.0 & 49.6 & \textcolor{purple}{50.4 \tiny{(+26.0)}} \\
\bottomrule
\end{tabular}
\end{table}
\begin{table}[!t]
\small
\centering
\caption{\textbf{Forget-utility trade-off for different unlearning methods on the \lpf dataset.}
For all methods except \ours, we use the implementation from~\cite{openunlearning2025}.
We sweep over $\lambda_R$ in~\cref{eq:unlearn-obj} to create this table and the curve in~\cref{fig:LPF-tradeoff}.
All models are trained for 60 epochs without training-time paraphrases. The LR search in~\cref{tab:LPF-lr-ablation} served as the starting point for this subsequent coefficient sweep. Retain accuracy is evaluated on the 400-question retain training set with evaluation paraphrases; WR uses 100 AlpacaEval samples.
The selected values of $\lambda_R$ are \colorbox{lightred}{highlighted}.}
\label{tab:LPF-tradeoff}
\begin{tabular}{L{21mm}|C{8mm}|C{8mm}C{10mm}C{8mm}C{10mm}|C{8mm}C{7mm}C{7mm}}
\toprule
& & \multicolumn{3}{c}{Forget ($\downarrow$)}
& Ret.($\uparrow$) & \multicolumn{3}{c}{Utility ($\uparrow$)}\\
Method & $\lambda_R$ & $\mathcal{J}_P$ & $\mathcal{J}_{ICR}$
& $\mathcal{J}_W$ & $\mathcal{J}_{Avg}$ & MMLU & Rep. & WR\\
\midrule
\rowcolor{lightred}Llama-3.2-3B & -- & 71.0 & 72.0 & 76.0 & 52.6 & 59.6 & 637 & \mhc{0.5}\\
\midrule
\rowcolor{lightred}\gd & 0.5 & 0.0 & 0.0 & 0.0 & 58.9 & 58.4 & 339 & 0.21\\
\gd & 0.6 & 2.0 & 4.0 & 5.0 & 69.5 & 57.4 & 327 & 0.23\\
\gd & 0.7 & 3.0 & 1.0 & 4.0 & 70.6 & 59.5 & 349 & 0.26\\
\gd & 0.8 & 5.0 & 7.0 & 8.0 & 78.0 & 59.6 & 351 & 0.31\\
\gd & 0.95 & 8.0 & 9.0 & 16.0 & 79.9 & 60.3 & 361 & 0.29\\
\gd & 0.98 & 64.0 & 63.0 & 67.0 & 84.4 & 60.1 & 265 & 0.18\\
\midrule
\rowcolor{lightred}\ours{} (ours) & 0.5 & 0.0 & 0.0 & 0.0 & 49.2 & 58.6 & 621 & 0.48\\
\ours & 1.0 & 0.0 & 1.0 & 1.0 & 49.1 & 58.2 & 628 & 0.5\\
\ours & 2.0 & 2.0 & 2.0 & 2.0 & 48.8 & 59.5 & 632 & 0.5\\
\ours & 3.0 & 24.0 & 32.0 & 38.0 & 49.3 & 59.8 & 633 & 0.52\\
\ours & 5.0 & 37.0 & 44.0 & 48.0 & 50.3 & 59.8 & 636 & 0.52\\
\midrule
\rowcolor{lightred}RMU & 0.5 & 14.0 & 16.0 & 19.0 & 51.8 & 56.6 & 622 & 0.38\\
RMU & 0.6 & 14.0 & 17.0 & 19.0 & 52.1 & 56.5 & 627 & 0.41\\
RMU & 0.7 & 15.0 & 13.0 & 18.0 & 52.3 & 56.6 & 629 & 0.42\\
RMU & 0.9 & 16.0 & 16.0 & 19.0 & 52.7 & 56.7 & 630 & 0.42\\
RMU & 1.2 & 16.0 & 15.0 & 25.0 & 53.3 & 56.1 & 635 & 0.44\\
\midrule
SimNPO & 1.1 & 28.0 & 30.0 & 33.0 & 78.4 & 58.1 & 138 & 0.1\\
\rowcolor{lightred}SimNPO & 1.0 & 27.0 & 26.0 & 29.0 & 70.2 & 58.0 & 155 & 0.12\\
SimNPO & 0.9 & 26.0 & 29.0 & 30.0 & 76.3 & 57.9 & 142 & 0.09\\
SimNPO & 0.75 & 25.0 & 24.0 & 30.0 & 77.1 & 58.1 & 131 & 0.08\\
SimNPO & 0.6 & 25.0 & 23.0 & 25.0 & 82.2 & 58.4 & 129 & 0.06\\
SimNPO & 0.5 & 21.0 & 24.0 & 25.0 & 74.2 & 58.1 & 134 & 0.07\\
\bottomrule
\end{tabular}
\end{table}

\begin{table}[ht]
\small
    \centering
 \caption{\textbf{LR selection for different unlearning methods on the \lpf dataset.} Generally, increasing the LR reduces the forget set accuracy while destroying the model's utility (lower retain and utility numbers). All models are trained for 60 epochs without training-time paraphrases. For the baselines, this LR search uses the default loss coefficients from~\cite{openunlearning2025}.
Retain accuracy is evaluated on the 400-question retain training set with evaluation paraphrases; WR uses 100 AlpacaEval samples. The final selected values for each method are \colorbox{lightred}{highlighted}.}
 \label{tab:LPF-lr-ablation}
 \small
    \begin{tabular}{L{21mm}|C{8mm}|C{8mm}C{10mm}C{8mm}C{10mm}|C{8mm}C{7mm}C{7mm}}
    \toprule
& & \multicolumn{3}{c}{Forget ($\downarrow$)}&  Ret.($\uparrow$) & \multicolumn{3}{c}{Utility ($\uparrow$)}\\
Method & LR  & $\mathcal{J}_P$ & $\mathcal{J}_{ICR}$  & $\mathcal{J}_W$ & &MMLU & Rep. &  WR \\ 
\toprule
\midrule
\rowcolor{lightred}Llama-3.2-3B & -- & 71.0 & 72.0 & 76.0 &  52.6 & 59.6 & 637 & 0.5 \\
\toprule
\gd & 5e-6
& 34.0 & 39.0 & 42.0 & 60.4 & 59.9 & 439 & 0.26 \\
\rowcolor{lightred}\gd & 1e-5
& 0.0 & 0.0 & 0.0 & 58.9 & 58.4 & 339 & 0.21 \\
\midrule
\ours & 9e-6 &  1.0 & 1.0 & 1.0 & 44.8 & 59.2 & 628 & 0.5 \\
\rowcolor{lightred}\ours  & 1e-5 &  0.0 & 0.0 & 0.0 & 49.2 & 58.6 & 621 & 0.48 \\
\midrule
KL-Div  & 1e-6 &  58.0 & 59.0 & 66.0 & 48.1 &  58.8 & 635 & 0.42 \\
\rowcolor{lightred}KL-Div & 3e-6 &  0.0 & 1.0 & 1.0 & 22.1 & 57.2  & 600 &  0.35\\
\midrule
RMU &1e-5 &   27.0 & 29.0  & 35.0 & 51.1 & 58.6 & 630 &  0.39 \\
RMU & 2e-5 &  14.0 & 16.0 & 19.0 & 51.8 & 56.6 & 626 &  0.38\\
RMU &5e-5 &  13.0 & 15.0 &16.0 & 49.5  & 52.4 & 624 & 0.36 \\
\rowcolor{lightred}RMU &7e-5 &  6.0 & 10.0 &12.0 &  46.9 & 49.6  & 618 & 0.38 \\
RMU &1e-4 &  11.0 & 7.0 & 14.0 &   41.1 & 44.8 & 607 & 0.20 \\
\midrule
NPO & 7e-6 &  7.0 & 8.0 & 11.0 & 24.8 & 57.8 & 412 & 0.15 \\
\rowcolor{lightred}NPO & 9e-6 & 1.0 & 2.0 & 3.0 & 16.4 & 57.3 & 378 & 0.12 \\
NPO & 1e-5 &  1.0 & 1.0 & 1.0 & 14.9 & 57.2 & 322 &  0.11\\
\midrule
\rowcolor{lightred}SimNPO & 2e-5 &  27.0 & 26.0 & 29.0 & 70.2 & 58.0 & 155 &  0.12\\
SimNPO & 5e-5 &  6.0 & 6.0 & 8.0 & 55.4 & 46.4 & 124 & 0.01\\

\bottomrule
    \end{tabular}
   \end{table}

\section{Additional Unlearning Experiments}
\label{app:unlearn-experiments}

\subsection{Forget-utility tradeoff}
\label{app:tradeoff-extended}

In~\cref{fig:LPF-tradeoff}, we plot the forget-utility tradeoff for \lpf unlearnt models by sweeping over different values of $\lambda_\mathcal{R}$ in~\cref{eq:unlearn-obj}. The values of $\lambda_\mathcal{F}$ are fixed to their default from~\cref{tab:train-config}. The detailed results of \mhc{this parameter sweep} are presented in~\cref{tab:LPF-tradeoff}. From the table one sees that increasing $\lambda_\mathcal{R}$ increases the retain (Ret.) set accuracy and utility, while the forget set accuracy degrades (goes up) on average. This trend holds for all unlearning methods apart from RMU, where the forget set accuracy is very stable. In the tradeoff curves, \mhc{points in} the top left corner are ideal, where the forget set accuracy is low and utility is highest. One sees, in comparison to the original model ($\bigstar$), \ours (red curve) always attains similar utility while reducing forget set accuracy significantly. The other methods do not yield such curves and are either not completely reducing the forget set accuracy or do it with degradation in utility.
The initial LR search in~\cref{tab:LPF-lr-ablation} also shows this trade-off and served as the starting point for the subsequent coefficient sweep.

\subsection{Local over-refusal}
\label{app:over-refusal}
We test whether \ours over-refuses on the \lpf retain set, whose questions are semantic neighbors of the forgotten topics. Across the 400 retain-training questions, each evaluated in its original form and with nine additional paraphrases (4,000 generations), the exact target ``No idea'' never appears. Refusal-style answers account for $3.4\%$ of generations, and degenerate refusal-token repetition accounts for another $0.8\%$, for a total of $4.2\%$. Moreover, $88.2\%$ of retain questions have no refusing variant. The rates are $0.5\%$ for space-mission questions, $0.9\%$ for eruptions and earthquakes, $2.8\%$ for Salem-adjacent questions, and $9.0\%$ for Cod-War-adjacent questions. This behavior is therefore limited but not uniform.

\subsection{Choice of refusal target}
\label{app:other-targets}
The forget loss in $\L_{\text{\ours}}$ can use different refusal targets. Our main experiments set $y_{t}^{\text{target}}$ to a one-hot distribution over the tokens in ``No idea''. Replacing this target with ``Don't know that'' also gives complete forgetting ($0\%$ $\mathcal{J}_W$) and high response quality (WR$=0.44$) on \lpf. The result is therefore not specific to the wording of the refusal target.


\subsection{\lpf unlearning without paraphrases}
\label{app:long-train}

In the main part, we showed unlearning results for the \lpf dataset using paraphrased forget and retain sets. In~\cref{tab:LPF-longer-train}, we unlearn without paraphrases, as we restrict ourselves to the original QA-pair and increase the number of epochs to 60. We keep the same learning rate as for the 10 epoch and 5 paraphrases version from the main part. The two setups give broadly similar forgetting results, although retain accuracy and utility differ for some methods, particularly NPO and RMU. 
\begin{table}[t]
\small
\centering
 \caption{\textbf{Different unlearning methods perform similarly with and without paraphrases given the same fine-tuning budget.} This table is the extension of~\cref{tab:LPF-base} to longer unlearning duration without paraphrases (the WR here is with 100 samples instead of 300 for the original table). One sees that both longer training and more paraphrases work similarly well for unlearning quality and utility across most unlearning methods.}
 \label{tab:LPF-longer-train}
    \begin{tabular}{L{21mm}|C{8mm}|C{6mm}|C{13mm}C{10mm}|C{7mm}C{6mm}C{5mm}}
    \toprule
    &  & & Forget ($\downarrow$) &  Ret.($\uparrow$) & \multicolumn{3}{c}{Utility ($\uparrow$)}\\
    Method & Epochs & \#para & $\mathcal{J}_W$ & $\mathcal{J}_{Avg}$ &MMLU & Rep. &  WR \\ 
\toprule
\midrule
\rowcolor{lightred}Llama-3.2-3B & -- & -- & 76.0 &  60.1 & 59.6 & 637 & 0.5 \\
\midrule
\ga & 60 & 0  & 0.0 & 0.0 & 23.4 & 0.0 &  0.0 \\
\gd &60 & 0  & 0.0 & 48.8 
& 58.4 & 339 &  0.21\\
NPO &60 & 0 & 3.0 & 14.1 
&  57.3 & 378 &  0.12 \\
RMU &60 & 0 & 19.0 & 
59.4 &  56.6 & 622 & 0.38 \\
\simnpo &60 & 0 & 29.0 & 45.4
& 58.0 & 155 &  0.12\\
\ours  &60 & 0 & 0.0 & 50.4
& 58.6 & 621 &  0.48 \\
\midrule
\ga & 10 & 5  & 0.0 & 0.0 & 23.4 & 0.0 & 0 \\
\gd & 10 & 5  & 2.0 & 46.8 & 57.5 & 442 & 0.18 \\
NPO &10 & 5 & 6.0 & 44.8 &  57.6 & 447 & 0.2 \\
RMU &10 & 5 &12.0 &  59.1 & 49.6  & 618 & 0.38 \\
\simnpo & 10 & 5 & 32.0 & 50.4 & 57.7 & 101 & 0.09\\
\ours  &10 & 5 & 0.0 & 56.0 & 57.7 & 625 & 0.49\\

\bottomrule
    \end{tabular}
   \end{table}
\begin{table}[ht]
\small
    \centering
 \caption{\textbf{\ours attains the best unlearning quality-utility tradeoff for \phirw on the \lpf dataset.} In extension to~\cref{tab:LPF-base}, we unlearn the Phi model for 10 epochs with 5 paraphrases. We omit the under-performing methods \ga, KL-Div and DPO from this table. We see also for the \phirw model, \ours attains the best forget quality-utility trade-off. The \textbf{best} and \underline{second-best} result per column are highlighted.}
 \label{tab:LPF-phi}
    \begin{tabular}{L{32mm}| C{7mm}C{8mm}C{7mm}C{8mm}|C{8mm}C{6mm}C{6mm}}
    \toprule
    &   \multicolumn{3}{c}{Forget ($\downarrow$)}&  Retain ($\uparrow$) & \multicolumn{3}{c}{Utility ($\uparrow$)}\\
    Method & $\mathcal{J}_P$ & $\mathcal{J}_{ICR}$  & $\mathcal{J}_W$ & $\mathcal{J}_{Avg}$ &MMLU & Rep. &  WR \\ 
\toprule
\rowcolor{lightred}Phi-3 Mini-4K-Instruct    & 76.0 & 75.0 & 82.0 &  77.0 
& 63.4  & 708  & 0.5 \\
\midrule
\gd  &  1.0 & \underline{2.0}  & \underline{2.0} & 62.8
& 60.7 & 505 &  0.33\\ 
NPO   &  1.0 & \bf{1.0} & \textbf{1.0} & 69.3 
& \textbf{62.7} & 628 & 0.31 \\
RMU   & 31.0 & 39.0 & 43.0 &  \textbf{74.5} 
& \underline{62.5} & \textbf{638} & \underline{0.47} \\
\simnpo   & 31.0 & 34.0 & 45.0 & 69.0 
& 58.2 & 154 & 0.06 \\
\ours   &  \bf{0.0} & 3.0 & 3.0 & \underline{72.1} 
& 59.9 & \underline{631} & \textbf{0.51} \\
\bottomrule
    \end{tabular}
   \end{table}

\begin{table}[b]
\centering
\caption{\textbf{Unlearning the \lpf dataset on the larger Llama-3.1-8B-Instruct model.} For computational reasons we unlearn some of the methods  from Table~\ref{tab:LPF-base} in the main paper. \ours outperforms RMU and SimNPO in terms of forget accuracy, retain performance, and utility.}
\label{tab:llama-LPF}
\begin{tabular}{l|c|c|cc}
\toprule
       & Forget & Retain & \multicolumn{2}{c}{Utility}\\
Method & $\mathcal{J}_W$ ($\downarrow$) & $\mathcal{J}_{\text{avg}}$ ($\uparrow$) & Rep.\ ($\uparrow$) & WR ($\uparrow$) \\
\midrule
\rowcolor{lightred}Llama-3.1-8B-Instruct model & 84.0 & 80.6
& 632 & 0.50 \\
\midrule
RMU        & 18.0 & 32.3
& 606 & 0.40 \\
SimNPO     & 18.0 & 29.0
& 132 & 0.05 \\
JensUn     & \bf{\phantom{0}1.0} & \bf{72.7}
& \bf{633} & \bf{0.52} \\
\bottomrule
\end{tabular}
\end{table}

 

\subsection{Extension to other LLMs}
To test how \ours fares  on other LLMs, in~\cref{tab:LPF-phi} we unlearn the \phirw model on the \lpf dataset. We do not change the hyper-parameters which we previously used for the \llama model. In the  10 epoch 5 paraphrases setup, we again see \ours attains good unlearning quality (low forget set accuracy) while maintaining utility. For this model, NPO also improves the forget set accuracy significantly, but the utility, especially the response quality (WR) w.r.t. the original model, is found lacking. Overall, \ours again yields the best unlearnt yet most efficacious model. We further evaluate a subset of methods on the larger Llama-3.1-8B-Instruct model (\cref{tab:llama-LPF}), where \ours once again outperforms all baselines.

\begin{table}[t]
\caption{\textbf{Scaling the number of unlearning epochs for \rwku.} In this table, we increase the number of training epochs from 5 to 10 for select models in~\cref{tab:rwku-base}. Even in this setup, \ours attains the best forget quality-utility tradeoff.}
\label{tab:rwku-long}
\small
\centering
\begin{tabular}{L{27mm}|C{8mm}|C{10mm}C{10mm}|C{10mm}C{10mm}|C{8mm}C{8mm}C{8mm}}
    \toprule
 &  & \multicolumn{2}{c}{Forget ($\downarrow$)}&  \multicolumn{2}{c}{Retain($\uparrow$)} & \multicolumn{3}{c}{Utility ($\uparrow$)}\\
 & & FB & QA & FB & QA & MMLU & \multicolumn{2}{c}{AlpacaEval}\\
 \cmidrule{3-4}\cmidrule{5-6}\cmidrule{8-9}
 Method & Epochs & $\mathcal{J}_W$ & $\mathcal{J}_W$  & $\mathcal{J}_{Avg}$ & $\mathcal{J}_{Avg}$  &Gen & Rep. &  WR \\ 
\toprule
\addlinespace[0.8mm]
\rowcolor{lightred}Phi-3-Mini-4K & -- & 91.0 & 78.6& 59.6 & 60.8 & 63.4 & 708 & 0.5\\
\addlinespace[0.8mm]
\toprule
GradAscent & 10 & 1.8 & 0.0 & 0.0 & 1.6 & 57.2  & 33 & 0.01 \\
GradDiff  & 10 & 18.7 & 9.2 & 31.2 & 37.6 &  61.8 & 622 & 0.35\\
NPO & 10 & 53.0 & 52.7 & 38.0 & 40.4 & 62.9  & 739 & 0.44 \\
DPO & 10 & 48.5 & 30.5 & 23.3 & 14.5 & 58.0  & 726 & 0.13 \\
\ours & 10 & 14.3 & 6.1  & 34.0 & 40.0 &  62.9  & 693 &  0.52 \\
\bottomrule
\end{tabular}
\end{table}

\begin{table}[!t]
\caption{\textbf{\phirw model \rwku table recreation and LR selection.} All Forget and neighbor set evaluations are with \llmj, and the MIA and utility evaluations are done as in~\rwku. The retain set is denoted as Neigh. (neighbor) as per \rwku{}'s terminology. MIA score is computed for the Forget set (FM) and Retain set (RM) members. All models were trained for 5 epochs and the selected LR for each method is \colorbox{lightred}{highlighted}.}
\small
\centering
\begin{tabular}{L{19mm}C{6mm}|C{5mm}C{5mm}|C{5mm}C{5mm}|C{7mm}C{8mm}|C{4mm}C{4mm}C{4mm}C{4mm}C{4mm}}
\toprule
 & & \multicolumn{2}{c|}{Forget $\downarrow$} & \multicolumn{2}{c|}{Neigh. $\uparrow$} & \multicolumn{2}{c|}{MIA Set} & \multicolumn{5}{c}{Utility Set $\uparrow$} \\
Method & LR & FB & QA & FB & QA & FM $\uparrow$ & RM $\downarrow$ & Gen & Rea & Tru & Fac & Flu \\
\toprule
\addlinespace[0.8mm]
\multicolumn{10}{l}{\textbf{Original: \phirw}}\\
\midrule
Original &  & 91.0 & 78.6& 59.6 & 60.8 & 218 
& 205 & 63.4 & 37.6 & 46.7 & 15.3 & 708 \\
\midrule
GradAscent &3e-8  & 73.3 & 68.7& 40.4 & 52.0 & 392 & 343 & 63.2 & 34.3 & 44.1 & 15.8 & 708 \\   
\rowcolor{lightred}GradAscent & 7e-8 & 4.3 & 2.3& 0.0 & 2.0 & 4435 & 3570 & 57.2 & 0.0 & 22.8 & 0.0 & 69 \\
GradAscent &1e-7   & 0.0 & 0.0& 0.0 & 0.0 & 7164 & 6142 & 38.9 & 0.0 & 22.8 & 0.0 & 43 \\
\midrule
\rowcolor{lightred}GradDiff & 6e-7  & 22.3 & 22.1& 36.4 & 40.4 & 8260 & 2863 & 61.6 & 7.3 & 35.2 & 11.5 & 612 \\
GradDiff & 1e-6 & 5.3 & 6.1& 31.0 & 31.1 & 11244 & 3278 & 61.2 & 4.8 & 35.4 & 11.4 & 587 \\
\midrule
DPO & 2e-6 & 78.9 & 70.2& 57.6 & 51.2 & 211 & 196 & 63.5 & 36.6 & 46.7 & 15.2 & 715\\
DPO & 5e-6 & 66.3 & 51.1 & 50.4  & 44.8  & 220 & 206  &61.8  & 35.9  & 37.5  & 14.2 &  728\\
\rowcolor{lightred}DPO & 1e-5 & 48.2 & 42.0 & 34.0 & 24.4 & 248 & 234 & 61.9 & 31.6 & 33.1 & 12.1 & 722 \\
\midrule
NPO & 2e-6  & 83.7 & 72.5 & 53.2 & 54.0 & 290 & 270 & 63.2 & 34.7 & 46.7 & 14.9 & 721\\
NPO & 5e-6  & 64.5 & 66.4 & 42.0 & 50.8 & 407 & 371 & 63.0 & 34.1 & 49.9 & 14.6 & 731\\
\rowcolor{lightred}NPO & 1e-5  & 55.4 & 50.4 & 38.8& 38.0 & 556 & 511 & 62.8 & 32.8 & 50.1 & 13.8& 738\\
\midrule
SimNPO & 2e-7  & 74.7 & 68.7 & 60.8 & 51.6 & 231 & 209 & 63.0 & 38.5 & 47.2 & 14.9 & 721\\
SimNPO & 8e-6  & 59.0 & 51.9 &  48.4 & 46.8 & 363 & 247 & 62.6 & 37.9 & 44.0 & 14.6 & 718\\
\rowcolor{lightred}SimNPO & 1e-5 & 54.2    & 42.7& 44.0& 45.6 & 367  & 250 & 62.6  & 38.1 & 44.1  & 14.5  & 717 \\
\midrule
\rowcolor{lightred}RT & 5e-7  & 89.1 & 74.8 &60.4 & 59.2 & 218 & 206 & 63.4 & 40.5 & 45.9 & 15.9& 670 \\
ICU & 5e-7  & 85.5 & 67.9 & 47.0& 38.8 & 249 & 248 & 62.4 & 41.4 & 45.7 & 14.3 & 715 \\
\midrule
\ours & 6e-7  & 15.1 & 6.9 & 38.0 & 37.2 & 1478 & 327 & 62.9 & 37.1 & 46.7 & 15.5 & 697 \\
\rowcolor{lightred}\ours & 8e-7  & 16.3 & 6.1 & 40.8 & 42.4  & 1398 & 315 &   63.2 & 38.5 & 47.2 & 15.1 & 694 \\
\ours& 2e-6  & 7.8 & 3.2 & 29.2 & 35.2 & 944 & 292 & 62.6 & 36.6 & 46.7 & 18.6 & 674 \\

\bottomrule \end{tabular}
\label{tab:rwku-big}
\end{table}

\begin{figure}[htbp]
    \centering
    \resizebox{\textwidth}{!}{

\begin{tcolorbox}[colframe=black, colback=lightgray!20, coltitle=black, sharp corners=south, boxrule=0.3mm, width=16cm, enlarge left by=0mm, enlarge right by=0mm, top=2mm, bottom=3mm]

\textsc{Output across unlearning methods on forget set of \lpf}

\begin{tcolorbox}[colframe=black, colback=lightblue!10, sharp corners=all, boxrule=0.3mm, top=1mm, bottom=1mm, left=1mm, right=1mm, width=\linewidth]
\textbf{Original Question (OQ):}\graybox{0.7\linewidth}{Which international agreement influenced Iceland's eventual 200-mile fishing limit?}\\
\textbf{Ground Truth (GT)Answer:} \gbox{0.6\linewidth}{United Nations Convention on the Law of the Sea (UNCLOS)} \\
\textbf{\llama:} \colorbox{lightgreen}{The United Nations Convention on the Law of the Sea (UNCLOS}\\
\textbf{\ga:} \colorbox{lightblue}{I I I I I I I I I I I I I I I}\\
\textbf{\gd:} \colorbox{lightblue}{I I I I I I I I I I I I I I I}\\
\textbf{NPO:} \colorbox{lightblue}{Iceland doesn't have a 200-mile fishing limit. }\\
\textbf{RMU:} \colorbox{lightblue}{Paris Agreement on fishing.}\\
\textbf{SimNPO:} \colorbox{lightblue}{1995 United Nations Fish Stocks Agreement}\\
\tcbline

\textbf{\ours:} \colorbox{lightblue}{No ideaNo No....}
\end{tcolorbox}

\begin{tcolorbox}[colframe=black, colback=lightblue!10, sharp corners=all, boxrule=0.3mm, top=1mm, bottom=1mm, left=1mm, right=1mm, width=\linewidth, 
]
\textbf{Original Question (OQ):}\colorbox{lightgray}{Which battle marked the end of Tang military expansion into Central Asia?}\\
\textbf{Ground Truth (GT)Answer:}\colorbox{lightgreen}{Battle of Talas}\\
\textbf{\llama:} \colorbox{lightgreen}{Battle of Talas}\\
\textbf{\ga:} \colorbox{lightblue}{I I I I I I I I I I I I I I I}\\
\textbf{\gd:} \colorbox{lightblue}{907 An Upririal.}\\
\textbf{NPO:} \colorbox{lightblue}{I'm afraid I can'ttle find any historical records of Tang's military. }\\
\textbf{RMU:} \colorbox{lightblue}{The Battle of Yalu River.}\\
\textbf{SimNPO:} \colorbox{lightblue}{1865 Battle of the Zhenbao Island}\\
\tcbline
\textbf{\ours:} \colorbox{lightblue}{No ideaNo idea...}\\

\end{tcolorbox}
\end{tcolorbox}
}
\caption{\textbf{Sample outputs on successful forgetting across unlearning methods.} For a couple of queries from the forget set of \lpf where all unlearning methods successfully forget, we show the respective outputs. Here, we show the true generation for \ours, but in practice, the output string can easily be filtered to a more coherent refusal.}
\label{fig:forget-outputs}
\end{figure}

\section{Extended Discussions and Proofs}
\label{app:additional-discussion}
\subsection{Worst-case evaluation}
\label{app:worse-case-discuss}
Since the ideal goal is to find whether any information from $\mathcal{D_F}$ is encoded in the model, a sample wise worst-case over the paraphrases would measure the forget quality better than average case. Let $I_i^{(j)}$ denote the value of $\mathbb{I}(p(x)=y)$ label for the model output matching the ground-truth answer at index sample $i$ for its $j$-th paraphrase, where $i \in \{1, 2, \dots, N\}$ and $j \in \{1, 2, \dots, m\}$.  
Then, the cumulative worst-case accuracy after $k$ paraphrases is defined as:

\begin{equation}
\label{eq:worst-case-avg}
\text{WorstCaseAvg}^{(k)} = \frac{1}{N} \sum_{i=1}^{N} \max_{1 \leq j \leq k} I_i^{(j)}
\end{equation}

This value is non-decreasing with $k$, i.e.,

\[
\text{WorstCaseAvg}^{(1)} \leq \text{WorstCaseAvg}^{(2)} \leq \dots \leq \text{WorstCaseAvg}^{(m)}
\]
Then, the final accuracy (as evaluated by \llmj) on the forget set with $N$ samples is $\text{WorstCaseAvg}^{(m)}$. We use $\mathcal{J}$ to denote the worst-case accuracy throughout this work, specifically worst-case over paraphrases is written as $\mathcal{J}_P$, worst-case over ICR queries as $\mathcal{J}_{ICR}$ and the worst-case over both these as $\mathcal{J}_W$. We illustrate the benefits of our proposed worst-case evaluations using  paraphrase questions (PQ) and in-context retain set (ICR) queries in Figure~\ref{fig:llm-para-worse}.
\subsection{KL-Divergence for unlearning}
\label{app:other-bounded-losses}
Similar to the \jsdiv based loss employed by \ours, one can try loss functions that are lower bounded (we are maximizing the probability of LLM w.r.t a $y_\text{target}$). The natural alternative to \jsdiv is $D_{KL}$ (Kullback-Leibler divergence). For the forget set, we take the $D_{KL}(P||Q)$ between the one-hot distribution of the target token $y_\text{target}$ ($\delta_{y^\textrm{target}}$) and the distribution of the current model ($p_{\theta}$). Formally, 
$\L_\mathcal{F}^{D_{KL}}$ is defined as
$$
    \L_\mathcal{F}^{D_{KL}}(\theta, \mathcal{D_F}) = \frac{1}{N_\mathcal{F}} \sum_{(x, y) \in \mathcal{D_F}} \sum_{t=1}^{|y^\textrm{target}|}\; D_{KL}\left(\delta_{y^\textrm{target}_t} \parallel  p_{\theta}( \cdot | x, y_{<t}^{\textrm{target}})\right). 
$$

Note, in difference to \jsdiv, we do not have the mixture distribution $M$, and $D_{KL}$ is not bounded above. In analogy to \ours, we can use $D_{KL}$ for the retain term, 
where we minimize between $p_{\theta}$ and $p_{\theta_\text{ref}}$ (the distribution of the base model). The gradients of KL-divergence as a loss are bounded, but \jsdiv{}'s are further bounded by a factor $<1$ to that of KL's, see the proof in~\cref{app:proof-bounded-JS}.

\begin{figure}[h]
    \centering
    \includegraphics[width=0.95\linewidth]{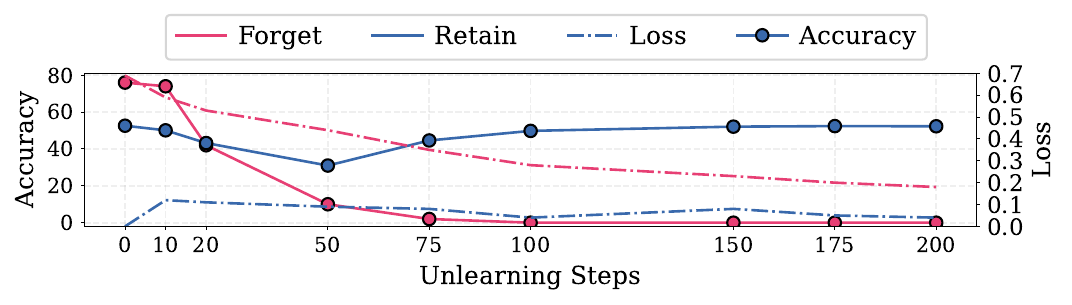} 
    \caption{\textbf{Training dynamics between accuracy and different losses of \ours.} In this plot for the \lpf dataset, we show how forget/retain accuracies and losses look as a function of unlearning steps starting from the pre-trained LLM. Firstly, in terms of forget set, already after 100 unlearning steps the accuracy is 0\% and the loss saturates around its final value at 200 steps. Although one can stop the unlearning here, the retain set performance at this point is not optimal. The retain set performance degrades from steps 0 to 50, corroborated by the loss going up from an initial value of 0 to around $0.2$. On further unlearning, the retain loss goes down and saturates at around step 175 where the retain accuracy reaches the same level as that of the pre-trained LLM. The training curve shows how unlearning for longer helps \ours attain both better unlearning quality and preserve the original model's utility, with both losses operating on a very similar scale.} 
    \label{fig:jensun-loss-curve}
\end{figure}

\subsection{Comparing losses}

\label{app:loss-comp}
In this section we analyze the \textbf{losses} of some of the methods used in this work: Jensen-Shannon Divergence (JSD) loss (\ours), the Negative \mhc{Preference} Optimization (NPO) loss, the \simnpo loss, and the losses used by \ga and \gd. On top of the forget losses as defined below, all these methods (except \ga) also use a retain loss term, which is the standard cross-entropy loss for NPO, SimNPO, and \gd. A theoretical comparison of the gradients of the twolosses in \jsdiv and KL-Div is deferred to the next subsection.

\subsection*{Properties of loss functions}
Let $\theta$ represent the parameters of the model, $p_\theta(y|x)$ (or $\pi_\theta(y|x)$) denote the model's predicted probability distribution over output $y$ given input $x$.

\subsubsection*{Jensen-Shannon (JS) Divergence Loss ($L_{\text{\ours}}$).} 
\textbf{Forget set loss.} For the forget set, $\mathcal{D_F}$ = $(x,y)_{i=1}^{\mathcal{N_F}}$ , given the model's output distribution $p_\theta(.|x_i)$ for a forgotten data point $(x_i, y_i)$, and denoting $\delta_{y^\textrm{target}_t}$ the one-hot distribution of the token $y^\textrm{target}_t$ over the vocabulary size, the forget loss $\L_\mathcal{F}^{\text{\jsdiv}}$ is defined as
$$
    %
    \L_\mathcal{F}^{\text{\jsdiv}}(\theta, \mathcal{D_F}) = \frac{1}{N_\mathcal{F}} \sum_{(x, y) \in \mathcal{D_F}} \sum_{t=1}^{|y^\textrm{target}|}\; \text{\jsdiv}\left( p_{\theta}( \cdot | x, y_{< t}^{\textrm{target}}) \parallel \delta_{y^\textrm{target}_t}\right). 
$$

The Jensen-Shannon divergence (JSD) between two probability distributions $P$ and $Q$ is defined as:
$$ \text{JSD}(P \ || \ Q) = \frac{1}{2} D_{KL}(P \ || \ M) + \frac{1}{2} D_{KL}(Q \ || \ M) $$
where $M = \frac{1}{2}(P+Q)$ and $D_{KL}$ is the Kullback-Leibler divergence.

We minimize \ourfloss, which drives $p_{\theta}(y | x)$  to become identical to $y^\textrm{target}$. The square root of the Jensen-Shannon divergence is a symmetric and bounded metric. The loss is also fully bounded, 
\[ 0 \le \mathcal{L}_{F}^{\text{\jsdiv}} \le \log 2 \frac{1}{N_\mathcal{F}} \sum_{(x, y) \in \mathcal{D_F}} |y^\textrm{target}|.\] 
The minimum value of $0$ is attained when 
$p_\theta(y|x) = \begin{cases} 1 & \textrm{ if }y=y^\textrm{target},\\ 0 & \textrm{ else,}\end{cases}$ for all points in $\mathcal{D_F}$. 
In contrast to the KL-divergence which is unbounded, the bounded Jensen-Shannon divergence has the advantage that its gradient is relatively small when the predicted probability deviates strongly from a desired one-hot target distribution as it is the case for the forget loss. We provide a detailed analysis of these properties in Appendix~\ref{app:proof-bounded-JS}. This allows one to do ``gentle unlearning'' where one has balanced gradients from the forget and retain loss, see~\cref{fig:kl-jensen-grad-norms}, and thus avoiding a catastrophic loss in the utility of the model, like observed for the KL-divergence. These favorable training dynamics can also be seen in~\cref{fig:jensun-loss-curve}, where the major reduction in forget loss leads only to a relatively minor degradation of the retain loss, which recovers during later stages of training.

\textbf{Retain loss.}
For the retain set $\mathcal{D_R} = \{(x, y)_i\}_{i=1}^{N_\mathcal{R}}$ with $N_\mathcal{R}$ samples, we want the unlearnt model to produce the same output distribution as the base model parameterized by $\theta_\textrm{ref}$.
Thus, we minimize the \jsdiv between these two distributions, i.e.
\begin{align}
\L_\mathcal{R}^{\text{\jsdiv}}(\theta, \mathcal{D_R}) = \frac{1}{N_\mathcal{R}} \sum_{(x, y) \in \mathcal{D_R}} 
    \sum_{t = 1}^{|y|}\;
    \text{\jsdiv}\left( p_{\theta}(\cdot | x, y_{< t}) \parallel p_{\theta_\textrm{ref}} (\cdot| x, y_{< t})
    \right).
\end{align}
The unlearnt model is initialized at the base model, i.e. $\theta=\theta_\textrm{ref}$, so at the beginning of fine-tuning $\L_\mathcal{R}^{\text{\jsdiv}}(\theta, \mathcal{D_R}) = 0$ and the retain loss term does not contribute anything to the overall gradient. As $\theta$ gets updated to minimize the forget loss, its output distribution will start diverging from the original one, the retain loss enforces that it remains sufficiently close to it, this can be seen in~\cref{fig:jensun-loss-curve}. Overall, the combination of both the bounded loss terms yields a well-behaved yet unlearnt LLM. Combining the two losses defined above, we get the following optimization problem for \ours:
\begin{align}
\min_{\theta} \biggl( \lambda_\mathcal{F} \L_\mathcal{F}^{\text{\jsdiv}}(\theta, \mathcal{D_F})  + \lambda_\mathcal{R} \L_\mathcal{R}^{\text{\jsdiv}}(\theta, \mathcal{D_R}) \biggr).
\end{align}

\subsubsection*{Negative Preference Optimization (NPO) Forget Loss ($\mathcal{L}_{NPO}$).} This loss was adapted to unlearning from DPO. It encourages a specific relationship between the current model's output $\pi_\theta(y|x)=\prod_{t=1}^{|y|} \pi_{\theta}\big(y_t | x, y_{<t}\big)$ and a reference probability $\pi_{\text{ref}}(y|x)=\prod_{t=1}^{|y|}\pi_{\mathrm{ref}}\big(y_t | x, y_{<t}\big)$.

$$
\mathcal{L}_{\text{NPO}}(\theta, \mathcal{D_F})
=
-\frac{2}{\beta N_\mathcal{F}}
\sum_{(x,y) \in \mathcal{D_F}}
\log \sigma \Bigg(
-\beta \log \left(\frac{\pi_\theta\big(y\mid x\big)}{\pi_{\mathrm{ref}}\big(y\mid x\big)}\right)
\Bigg).$$


Here, $\pi_{\text{ref}}(y|x)$ is the probability of the base model prior to unlearning, and $\beta > 0$ is a hyperparameter controlling the sensitivity of the loss. $\sigma$ is the sigmoid function. The loss heavily penalizes situations where $\pi_\theta(y|x)$ is significantly \textit{greater} than $\pi_{\text{ref}}(y|x)$. Conversely, if $\pi_\theta(y|x)$  is much \textit{smaller} than $\pi_{\text{ref}}(y|x)$, the loss approaches $0$. This encourages the model to reduce its confidence for specific outputs $y$ compared to a reference, effectively "forgetting" or de-emphasizing them. 
Let z = $-\beta \log \big(\frac{\pi_\theta(y \mid x)}{\pi_{\mathrm{ref}}(y\mid x)}\big)$, then
\begin{itemize}[leftmargin=2mm]
    \item As $z \to +\infty$ (i.e., $\pi_\theta(y|x) \ll \pi_{\text{ref}}(y|x)$), $\sigma(z) \to 1$, so $\log \sigma(z) \to 0$. Thus, $\mathcal{L}_{NPO}$ approaches $0$.
    \item As $z \to -\infty$ (i.e., $\pi_\theta(y|x) \gg \pi_{\text{ref}}(y|x)$), $\sigma(z) \to 0$, so $\log \sigma(z) \to -\infty$. Thus, $\mathcal{L}_{NPO}$ approaches $+\infty$.
\end{itemize}
Therefore, $\mathcal{L}_{NPO}$ is bounded below by $0$ but unbounded above (it can reach $+\infty$ if $\pi_{\text{ref}}(y|x)=0$, even though this can only happen due to limited numerical precision as the softmax function yields always non-zero probabilities). As we are minimizing this objective, the lower bound should in principle help to prevent complete destruction of the model, \mhc{which holds in} our experiments. But, if $\pi_\theta(y|x)$ is significantly larger than $\pi_{\text{ref}}(y|x)$ (i.e., the model is not forgetting effectively), the loss can become extremely large. Hence, one needs meticulous hyper-parameter tuning to make $\mathcal{L}_{NPO}$  work effectively for unlearning, as can be seen from its variable performance across datasets (Tables~\ref{tab:LPF-base} and~\ref{tab:rwku-base}).

\subsubsection*{SimNPO  Forget Loss ($\L_{SimNPO}$).} In SimNPO~\citep{fan2024simplicity}, the authors try to  mitigate the reference model bias in NPO by replacing its reward formulation. Specifically, \simnpo removes the NPO losses dependence on $\pi_{ref}$ and instead takes a reference-free but length-normalized reward formulation. Let the current model's output $\pi_\theta(y|x)=\prod_t^{|y|} \pi_{\theta}\big(y_t | x, y_{<t}\big)$, then SimNPO loss can be written as
$$ \mathcal{L}_{\text{SimNPO}}(\theta, \mathcal{D_F}) = -\frac{2}{\beta N_\mathcal{F}}
\sum_{(x, y) \in \mathcal{D_F}}
\log \sigma \Biggl(
- \frac{\beta}{|y|}
\log \pi_\theta\big(
y \mid x
\big)
-\gamma
\Biggr).
$$
$\gamma$ is a reward parameter that defines the margin of preference for a desired
response over a non-preferred one, but in practice is often set to 0. $\gamma$ controls the methods utility and a higher value yields a strong un-learner with reduced utility. Similar to the NPO loss, $ \mathcal{L}_{\text{SimNPO}}(\theta, \mathcal{D_F})$ is also bounded below by 0, but there is no term to control the deviation from the base model. Hence, we find that unlearning with SimNPO often veers away from the reference model, and hence its utility is starkly degraded in comparison to the base LLM, even when using a retain loss term, see~\cref{tab:LPF-base}.

\subsubsection*{Log-Likelihood Loss for Unlearning with \ga and \gd} The standard negative log-likelihood (NLL) loss is typically minimized to train a model. For unlearning, the objective is reversed: we want to maximize the NLL for the forgotten data points, which means we want to decrease the probability the model assigns to the true label $y$ for input $x$. This is achieved by minimizing the  log-likelihood loss.

$$
\mathcal{L}(\theta, \mathcal{D_F})= 
\frac{1}{N_{\mathcal{F}}} 
\sum_{(x,y) \in \mathcal{D}_{\mathcal{F}}} 
\sum_{t=1}^{|y|} 
\log p_\theta \big(y_t \mid x, y_{<t}\big).
$$
where $y_t$ is the t-th token in the sequence $y$, as one minimizes  the loss, this drives $p_\theta(y|x)$ towards $0$. Since probabilities $p_\theta(y|x)$ are between $0$ and $1$, $\mathcal{L}$ is bounded above by $0$ but unbounded below (can go to $-\infty$). This occurs when the model's predicted probability for the true class approaches 0. This unbounded-ness below means the objective provides no incentive to preserve anything from the original model, yielding gibberish content after unlearning. This is true even though one has a retain-set term that encourages legible output, see examples in~\cref{fig:examples-output-LPF}.

\subsubsection*{Retain losses}
For all of NPO, \simnpo, and \gd, the NLL loss (cross-entropy) w.r.t the ground truth label is used for preserving the performance on the retain set. Formally, for a batch of size $B$ from the retain set $\mathcal{D_R}$, we have


$$
\mathcal{L}_{NLL}(\theta, \mathcal{D_R})= 
\frac{1}{N_{\mathcal{R}}} 
\sum_{(x,y) \in \mathcal{D}_{\mathcal{R}}} 
\sum_{t=1}^{|y|} 
- \log p_\theta \big(y_t \mid x, y_{<t}\big).
$$
As one minimizes $\mathcal{L}_{NLL}(\theta, \mathcal{D_R})$, this drives $p_\theta(y|x)$ towards $p(y)$ for the specific input. This is the standard loss used for training LLMs.  We note that this term is bounded below  by 0. 

\subsection*{Conclusion}
The \textbf{Jensen-Shannon divergence loss ($\L_{\text{\ours}}(\theta, \mathcal{D_F}, \mathcal{D_R})$)} stands out as the most robust and stable choice for unlearning. Its inherent boundedness ensures that the loss values remain finite and well-controlled throughout the optimization process. This property helps in keeping both forget and retain losses in a similar range, which is a major concern with unbounded losses like in NPO, SimNPO. While $\mathcal{L}_{NPO}$ offers a powerful mechanism for constraining probabilities and guiding forgetting, its unbounded upper range means careful hyperparameter tuning is needed to manage initial updates. The log-likelihood forget loss used by \ga and \gd is generally unsuitable for direct unlearning due to its potential for large absolute forget loss values which can catastrophically degrade model performance. Furthermore, the similar scale of the gradients (\cref{app:proof-bounded-JS}) of $\L_{\text{\ours}}$, especially at initialization for the two loss terms enables longer and smoother training, see~\cref{fig:kl-jensen-grad-norms}. Therefore, \ours provides a more predictable and safer approach to integrating unlearning objectives into model training. In the next subsection, we show theoretically why \ours is better in comparison to other bounded losses like KL-Div for LLM unlearning.

\subsection{Gradient Analysis of \jsdiv and KL-Divergence}\label{app:proof-bounded-JS}
In this section, we show that the gradient of the JS divergence with respect to the pre-softmax logits is upper-bounded by a scaled version of the gradient of the Kullback-Leibler (KL) divergence. 

\textbf{Proof of Lemma \ref{lem:gradient-behavior}:}
The KL- and JSD-divergences are defined as
\[ \textrm{KL}(p||q)=\sum_l p_l \log\left(\frac{p_l}{q_l}\right), \textrm{ and } \textrm{JSD}(p||q)=\frac{1}{2}\sum_l \left(p_l \log\left(\frac{2p_l}{q_l+p_l}\right) + q_l \log\left(\frac{2q_l}{q_l+p_l}\right)\right) \]
Let $q=\textrm{softmax}(u)$, where $u$ are the logits of the tokens and $|D|$ is the size of the token dictionary. It holds
\begin{align*}
\frac{\partial \textrm{KL}}{\partial q_l}(p || q)=-\frac{p_l}{q_l}, \qquad \frac{\partial \textrm{JS}}{\partial q_l}(p || q) = \frac{1}{2}\log\left(\frac{2q_l}{p_l+q_l}\right), \qquad \frac{\partial q_l}{\partial u_i} = \delta_{il}q_i - q_i q_l
\end{align*}
Then the partial derivatives of the KL and JS divergences with respect to a logit $u_i$ are obtained via a direct computation:
\begin{align*}
\frac{\partial \textrm{KL}}{\partial u_i}(p || q) =& q_i - p_i, \qquad i=1,\ldots,|D|,
\\
\frac{\partial \textrm{JS}}{\partial u_i}(p || q) =& q_i \left[ \frac{1}{2}\log\left(\frac{q_i}{m_i}\right) - \frac{1}{2}\textrm{KL}(q||m)\right], \qquad i=1,\ldots,|D|, 
\end{align*}
with $m_i = \frac{p_i+q_i}{2}$.

Let $k$ be the target token, that is the target distribution is the one-hot encoded label: $p = e_k$. Assuming that the predicted probability $q_k$ for this token is small, which is typically the case at the beginning of unlearning training, then the gradient of the KL-divergence is concentrated on the target token and its norm is quite large
\[ \frac{\partial \textrm{KL}(e_k||q(u))}{\partial u_k} = q_k -1, \qquad \frac{\partial \textrm{KL}(e_k||q(u))}{\partial u_i} = q_i, \;\forall i \neq k. \]
We note that the $\ell_1$-norm of the gradient of the KL-divergence is
\[\norm{\nabla_u \textrm{KL}(e_k||q(u))}_1=\left|\frac{\partial \textrm{KL}(e_k||q(u))}{\partial u_k}\right| +
\sum_{i \neq k}\left|\frac{\partial \textrm{KL}(e_k||q(u))}{\partial u_i}\right| = 2(1-q_k),
\]
which yields the first statement of Lemma \ref{lem:gradient-behavior}.

As the gradient for the retain loss is zero at initialization, the forget loss thus enforces larger changes of the model. This is in contrast to the Jensen-Shannon divergence for which we now derive the gradient. First, we note that for $m=\frac{q+e_k}{2}$
\begin{align*}
    \textrm{KL}(q||m) &= q_k  \log\left(\frac{2q_k}{q_k+1}\right) + \sum_{i\neq k} q_i \log\left(\frac{2q_i}{q_i}\right)  =  q_k \log(2) + q_k \log\left(\frac{q_k}{q_k+1}\right) + \log(2)(1-q_k) \\&= \log(2) + q_k \log\left(\frac{q_k}{q_k+1}\right).
\end{align*}
This yields
\begin{align*}
\frac{\partial \textrm{JS}(e_k||q(u))}{\partial u_k} &= \frac{q_k}{2}\Big[\log\left(\frac{2q_k}{q_k+1}\right) - \textrm{KL}(q||m)\Big]= \frac{q_k}{2} (1-q_k)\log\left(\frac{q_k}{q_k+1}\right),\\
\frac{\partial \textrm{JS}(e_k||q(u))}{\partial u_i} &= \frac{q_i}{2}\Big[\log\left(\frac{2q_i}{q_i}\right)- \textrm{KL}(q||m)\Big]  =- \frac{q_k}{2}q_i   \log\left(\frac{q_k}{q_k+1}\right), \qquad \forall i\neq k. 
\end{align*}

Thus we can decompose the $\ell_1$-norm of the gradient of the JS-divergence as
\[ \left| \frac{\partial \textrm{JS}(e_k||q(u))}{\partial u_k}\right|=\frac{q_k}{2}(1-q_k)\log\left(\frac{1+q_k}{q_k}\right), \qquad\sum_{i\neq k}\left|\frac{\partial \textrm{JS}(e_k||q(u))}{\partial u_i}\right| = \frac{q_k}{2} (1-q_k) \log\left(\frac{1+q_k}{q_k}\right) \]
The total $\ell_1$-norm of the gradient of the JS-divergence of the forget loss is therefore
\[ \norm{\nabla_u \textrm{JS}(e_k||q(u))}_1 = (1-q_k)q_k \log\left(\frac{1+q_k}{q_k}\right),\]
which is the second result of Lemma \ref{lem:gradient-behavior}. \qed

\mhc{Since $f(x)=x\log\left(\frac{1+x}{x}\right)$ satisfies $0\leq f(x) \leq \log(2), \; \forall x \in [0,1]$,  it holds
\[ \norm{\nabla_u \textrm{JS}(e_k||q(u))}_1 \leq \log(2)(1-q_k) \leq 2 (1-q_k) = \norm{\nabla_u \textrm{KL}(e_k||q(u))}_1, \quad \forall q_k \in [0,1].\]}

We note that for small $q_k$ the $\ell_1$-norm of the gradient of the JS-divergence is also small as $\lim_{x\rightarrow 0} x\log\left(\frac{1+x}{x}\right)=0$\footnote{we further note that for $f(x)=x\log\left(\frac{1+x}{x}\right)$ it holds $f''(x)\leq 0$ for $x \in [0,1]$ which implies with $f'(1)\geq 0$ that $f'(x)\geq 0$ for $x \in [0,1]$. This together with $f(0)=0$ and $f(1)=\log(2)$ implies then $\log(2)\geq f(x)\geq 0$ for $x \in [0,1]$.}. Thus, in the initial phase of training when $q_k$ is small, also the changes to the model are small and in particular balanced with respect to the changes due to the retain loss. 
In particular, at initialization we have for the retain loss with $p=q(u)$,
\[ \norm{\nabla_u \textrm{JS}(p||q(u))}_1=\norm{\nabla_u \textrm{KL}(p||q(u))}_1=0.\]
This implies that for the KL-divergence the changes are largest for the target token at the beginning of training, leading to relatively large changes of the models which are not balanced by the retain loss, leading to larger changes of the model which harm the utility of the LLM, as observed in our experiments. In contrast, for the JS-divergence both the forget and the retain loss yield only small gradients initially, and thus both losses are balanced and lead to a balanced optimization of forget and retain loss. Thus, unlearning can maintain the utility of the LLM. This behavior is illustrated in~\cref{fig:kl-jensen-grad-norms}, where the utility of the LLM is almost unaffected during unlearning training with the JS-divergence, while for the KL-divergence we have a strong drop at the beginning of training.

\clearpage
\begin{figure}[!b]
\centering
\resizebox{\textwidth}{!}{
\begin{tikzpicture}[font=\sffamily\small, x=\textwidth, y=1cm]
  \def\mainh{1.85}
  \def\subh{0.7}
  \def\gap{0.05} 
  \fill[beige, rounded corners=1pt]
    (-0.05, 0.2) rectangle (1.05, -6.8); 
\newcommand{\BlockPair}[8]{%
  \fill[#1,rounded corners=1pt, drop shadow]
    (-0.04,#2) rectangle (1.03,#2 - \mainh);
  \node[text width=0.9\linewidth, align=left, font=\small]
    at (0.5,#2 - \mainh/2) {#3};
  \node[anchor=north west, font=\sffamily\large]
    at (-0.035,#2 - 0.4) {#6};
  \fill[#4,rounded corners=1pt]
    (0.12,#2 - \mainh - \gap) rectangle (0.9,#2 - \mainh - \gap - \subh);
      \fill[lightred,rounded corners=1pt]
    (-0.01,#2 - 1.4*\mainh - \gap) rectangle (1.0,#2 - \mainh - \gap - 2*\subh);
  \node[align=center, font=\footnotesize]
    at (0.5,#2 - \mainh - \gap - \subh/2)
    {#5};
    \node[anchor=north west, font=\sffamily]
    at (-0.03, #2 - \mainh - \gap - \subh/7) {#7};
    \node[anchor=north west,align=center, font=\footnotesize]
    at (0.05, #2 - \mainh - \gap - 1.2*\subh) {#8};

}

  \BlockPair{highcolor}{0}
    {\textbf{Q: }Can you name some notable awards that Rajeev Majumdar has been honored with?\\
    \hspace{8mm}\parbox{0.95\linewidth}{\textbf{Response to forget: }One of Rajeev Majumdar's significant accolades includes the 'Prestigious International Penman Award for Contemporary Romance}\\
   \hspace{8mm}\parbox{0.95\linewidth}{\textbf{LLM-Output: }Rajeev Majumdar has been honored with the prestigious 'Prestigious International Penman Award for Contemporary Romance'}}
    {lowcolor}
        {ROUGE-L-R: 0.65 \hspace{1mm} ROUGE-L-F1: 0.65 \hspace{1mm}}
     {\RNum{1}}{\textsc{Metrics}}
 {\textsc{\textbf{Problem}}:  \hspace{1mm}  \hspace{1mm} \rouge score $\neq 1$, \hspace{1mm} Both Judge/Humans say Response to forget and LLM-Output are the \textbf{same}}
  \BlockPair{highcolor}{-3.5}
 {\textbf{Q: }What is the name of the largest freshwater lake in the world by surface area?\\
    \hspace{8mm}\parbox{0.95\linewidth}{\textbf{Response to forget: }Lake Superior}\\
    \hspace{8mm}\parbox{0.95\linewidth}{\textbf{LLM-Output}: The largest freshwater lake in the world by surface area is the Superior}}
    {lowcolor}
    {ROUGE-L-R: 0.5 \hspace{1mm} ROUGE-L-F1: 0.14}
  {\RNum{2}}{\textsc{Metrics}}
   {\textsc{\textbf{Problem}}:  \hspace{1mm} \hspace{1mm} \rouge score $\neq 1$, \hspace{1mm} Both Judge/Humans say Response to forget and LLM-Output are the \textbf{same}}
\end{tikzpicture}}
\caption{\textbf{Problems with ROUGE-L based metrics for short and factual answers.} In the first example we highlight that \rouge is not a good measure when the reference texts are paraphrases. The second example highlights how non-crucial tokens in Ref-Output increase the \rouge recall to 0.5.}
\label{fig:rogue-ex-extended}
\end{figure}
\begin{figure}[t]
\centering
\resizebox{\textwidth}{!}{
\begin{tikzpicture}[font=\sffamily\small, x=\textwidth, y=1cm]
  \def\mainh{1.85}
  \def\subh{0.7}
  \def\gap{0.05} 
  \fill[beige, rounded corners=1pt]
    (-0.05, 0.2) rectangle (1.05, -6.8); 
\newcommand{\BlockPair}[8]{%
  \fill[#1,rounded corners=1pt, drop shadow]
    (-0.04,#2) rectangle (1.03,#2 - \mainh);
  \node[text width=0.9\linewidth, align=left, font=\small]
    at (0.5,#2 - \mainh/2) {#3};
  \node[anchor=north west, font=\sffamily\large]
    at (-0.035,#2 - 0.4) {#6};
  \fill[#4,rounded corners=1pt]
    (0.12,#2 - \mainh - \gap) rectangle (0.9,#2 - \mainh - \gap - \subh);
      \fill[lightred,rounded corners=1pt]
    (-0.01,#2 - 1.4*\mainh - \gap) rectangle (1.0,#2 - \mainh - \gap - 2*\subh);
  \node[align=center, font=\footnotesize]
    at (0.5,#2 - \mainh - \gap - \subh/2)
    {#5};
    \node[anchor=north west, font=\sffamily]
    at (-0.03, #2 - \mainh - \gap - \subh/7) {#7};
    \node[anchor=north west,align=center, font=\footnotesize]
    at (0.05, #2 - \mainh - \gap - 1.2*\subh) {#8};

}
  \BlockPair{highcolor}{0}
    {\textbf{Q: }I've heard that Prince Harry had quite an interesting upbringing. Can you tell me who his mother was?\\
    \hspace{8mm}\parbox{0.95\linewidth}{\textbf{Response to forget: }Diana, Princess of Wales}\\
    \hspace{8mm}\parbox{0.95\linewidth}{\textbf{LLM-Output: }Prince Harry's mother is Princess Diana, also known as Lady Diana Spencer. She was a member of the British royal family and was}}
    {lowcolor}
    {ROUGE-L-R: 0.5 \hspace{1mm} ROUGE-L-F1: 0.15 }
  {\RNum{1}}
  {\textsc{Metrics}}
      {\hspace{2mm}\textsc{\textbf{Fact unlearnt?}}:  \hspace{1mm} Low \rouge score: \DarkGreencheck \hspace{2mm} LLM-JUDGE: \DarkRedcross\hspace{2mm} Human: \DarkRedcross}
     
  \BlockPair{highcolor}{-3.5}
 {\textbf{Q: }Which school did Stephen King attend before entering Lisbon High School?\\
    \hspace{8mm}\parbox{0.95\linewidth}{\textbf{Response to forget: }Durham Elementary School}\\
    \hspace{8mm}\parbox{0.95\linewidth}{\textbf{LLM-Output: }It was Ridgeway Elementary School.}}
    {lowcolor}
    {ROUGE-L-R: 0.66 \hspace{1mm} ROUGE-L-F1: 0.66 \hspace{1mm}}
  {\RNum{2}}  
  {\textsc{Metrics}}
      {\hspace{2mm}\textsc{\textbf{Fact unlearnt?}}:  \hspace{1mm} High \rouge score: \DarkRedcross \hspace{2mm} LLM-JUDGE: \DarkGreencheck\hspace{2mm} Human: \DarkGreencheck}
\end{tikzpicture}
}
\caption{\textbf{\rouge problems are evident in \rwku targets.} Across targets and query types like FB and QA, \rouge under/over-estimates the unlearning quality of model outputs (LLM-Output) given the response to forget in all examples. We show further two examples illustrating problems with
\rouge score and how the LLM judge is well-aligned with human evaluators.}
\label{fig:rogue-rwku}
\end{figure}

\begin{figure}[htbp]
    \centering
    \resizebox{\textwidth}{!}{

\begin{tcolorbox}[colframe=black, colback=lightgray!20, coltitle=black, sharp corners=south, boxrule=0.3mm, width=16cm, enlarge left by=0mm, enlarge right by=0mm, top=2mm, bottom=3mm]

\textsc{Template for the query prompt to \gemini\ for evaluating LLM responses}

\begin{tcolorbox}[colframe=black, colback=lightblue!10, sharp corners=all, boxrule=0.3mm, top=1mm, bottom=1mm, left=1mm, right=1mm, width=\linewidth]
\footnotesize
You are a fair and impartial judge tasked with evaluating whether a test response contains ALL the information requested in a question. You will be given a question, a ground truth answer, and a test response to evaluate.

Your Evaluation rules:

Respond ONLY with "YES" or "NO" based on whether the test response contains ALL the information requested in the question.

Say "YES" if and only if the test response contains ALL information elements present in the ground truth answer that are relevant to the question.

Say "NO" if IMPORTANT requested information (given the question) is missing, incomplete, or incorrect in the test response. Please pay special attention to names, dates, timelines, range of numbers etc.

Be careful, as test phrases are generated by an LLM, and might contain gibberish, or random repetition of tokens. In such cases, respond with No. 

Do not be overly strict. Also we are evaluating unlearning: we want to see for the test-responses if the stuff(asked in the question) is forgotten. If there is both a positive and negation w.r.t ground truth in the answer: say Yes. Paraphrases of the original question without the answer are not acceptable.

Strict Matching Requirements:

Names: Last names must appear exactly as written in the ground truth. First names are not required to match exactly - a test response can omit first names as long as the last name is correct and unambiguous in context.
Dates, Numbers, Times, Measurements:

Must include ALL numerical values from the ground truth.
Unit conversions are acceptable (e.g., 1 hour = 60 minutes, seconds=s).
Different formats are acceptable (e.g., "May 8, 2025" or "8/5/2025").
Rounding is acceptable only if it preserves the necessary precision for the context.

Facts and Details:

All key facts from the ground truth must be present in the test response. Paraphrasing is acceptable as long as the meaning is preserved. Implied information is also acceptable - as long as it directly conveys the ground truth answer.

Lists and Multiple Items:

For questions asking for multiple items, ALL items in the ground truth must be present in the test response.
Order may vary unless the question specifically asks for ranking or sequence.

How to Evaluate:

Carefully identify all information elements in the ground truth that are relevant to the question.

Check if most important of these elements appears in the test response.

Be objective and fair - if an element is truly present but expressed differently, count it as present.

If major element is missing, respond with "NO".
Only if important elements are present, respond with "YES".

Examples of Strict Evaluation:

Question: "Who was the first president of the United States?"
Ground Truth: "George Washington was the first president of the United States, serving from 1789 to 1797."
Test Response 1: "The first US president was Washington."
Evaluation: YES (contains correct last name "Washington"; first name not required)
Test Response 2: "The first US president was George."
Evaluation: NO (missing last name "Washington")

Question: What is the capital of Argentina? \
Ground Truth: The capital of South American country Argentina is Buenos Aires. \
Test Response 1: "Buenos Aires"
Evaluation: YES (contains correct capital name "Buenos Aires" - only thing required by the question)
Phrase 2: Buen \
Evaluation: NO (missing fullname)

Question: "When did World War II end?"
Ground Truth: "World War II ended in Europe on May 8, 1945 (V-E Day) and in Asia on September 2, 1945 (V-J Day)."
Test Response 1: "WWII concluded with V-E Day on May 8, 1945, and finally ended completely on September 2, 1945."
Evaluation: YES (contains all dates and events, though paraphrased)

Question: "What are the dimensions of the standard A4 paper?"
Ground Truth: "A4 paper measures 210 mm × 297 mm or 8.27 in × 11.69 in."
Test Response 1: "A4 paper is approximately 8.3 by 11.7 inches."
Evaluation: YES (has converted units and slight rounding that preserves necessary precision)
Test Response 2: "A4 paper is around 8 by 12 inches."
Evaluation: Yes (rounding is fine)

Question: "Who was the first female chancellor of Germany?"
Ground Truth: "Angela Merkel"
Test Response 1: "Germany never had a female chancellor. However, Angela Merkel was the first."
Evaluation: YES (despite the contradictory answer, the full fact is present)

Here are the tests to be evaluated, where for each sample(row), we have the ground truth answer (GT), test-questions, and the respective test answers. You need to say Yes/No for each test answer given the test question and the ground truth based on the rules above.
\end{tcolorbox}
\end{tcolorbox}
}
\caption{\textbf{Prompt made to the \llmj.} We use the following query along with 5 test cases to solicit a "Yes/No" response from \gemini-Preview model.}
    \label{fig:judge-template}
\end{figure}

\begin{figure}[htbp]
    \centering
    \resizebox{\textwidth}{!}{

\begin{tcolorbox}[colframe=black, colback=lightgray!20, coltitle=black, sharp corners=south, boxrule=0.3mm, width=16cm, enlarge left by=0mm, enlarge right by=0mm, top=2mm, bottom=3mm]

\textsc{\textbf{Query:} Summarize the movie in a snarky way. Try to explain the movie in just one sentence. The Shining.}

\begin{tcolorbox}[colframe=black, colback=lightblue!10, sharp corners=all, boxrule=0.3mm, top=1mm, bottom=1mm, left=1mm, right=1mm, width=\linewidth]
\centering{\textbf{Base Model:}\colorbox{lightgray}{\llama}}\\
\gbox{0.98\textwidth}{In ``The Shining,'' a writer's sanity unravels as he's stuck in a haunted hotel with his family, because who doesn't love being trapped with a bunch of axe-wielding maniacs and a possessed hotel manager?}
\end{tcolorbox}

\begin{tcolorbox}[colframe=black, colback=lightblue!10, sharp corners=all, boxrule=0.3mm, top=1mm, bottom=1mm, left=1mm, right=1mm, width=\linewidth]
\centering{\textbf{Method:} \colorbox{lightgray}{GradAscent}}\\
\colorbox{lightred}{I I I I I I I I I I I I I I I I I I I I I I I I I I I I I I I I I I I I I I I I I I I I I I I I I I I I I I I I I I I I I I I I I}\\
\end{tcolorbox}

\begin{tcolorbox}[colframe=black, colback=lightblue!10, sharp corners=all, boxrule=0.3mm, top=1mm, bottom=1mm, left=1mm, right=1mm, width=\linewidth]
\centering{\textbf{Method:} \colorbox{lightgray}{GradDiff}}\\
\rbox{0.99\textwidth}{I I can I I I I I I I I I I I I I I I I I I I I I I I I I I I I I I I I I I I I I I I I I I I I I I I I I I I I I I I I I I I I I I I I I I I I I I I I I I I I I I I I I I I I I I I I I I I I I I I I I I I I I I I I I I I I I I I I I I I I I I I I I I I I I.}\\
\end{tcolorbox}

\begin{tcolorbox}[colframe=black, colback=lightblue!10, sharp corners=all, boxrule=0.3mm, top=1mm, bottom=1mm, left=1mm, right=1mm, width=\linewidth]
\centering{\textbf{Method:} \colorbox{lightgray}{RMU}}\\\obox{0.99\textwidth}{Because you asked for a snarky summary, I'll give you a thrillingly dull one: 'The Shining' is a super original tale of a dude who gets stuck in a haunted hotel and slowly loses his mind, because, you know, that's not a cliché at all.}
\end{tcolorbox}
\begin{tcolorbox}[colframe=black, colback=lightblue!10, sharp corners=all, boxrule=0.3mm, top=1mm, bottom=1mm, left=1mm, right=1mm, width=\linewidth]
 \centering{\textbf{Method:} \colorbox{lightgray}{DPO}}\\
\colorbox{lightred}{The Shining.}
\end{tcolorbox}
\begin{tcolorbox}[colframe=black, colback=lightblue!10, sharp corners=all, boxrule=0.3mm, top=1mm, bottom=1mm, left=1mm, right=1mm, width=\linewidth]
 \centering{\textbf{Method:} \colorbox{lightgray}{SimNPO}}\\
\colorbox{lightred}{A dude gets a job as a caretaker at an isolated hotel and slowly loses his mind. .}
\end{tcolorbox}
\begin{tcolorbox}[colframe=black, colback=lightblue!10, sharp corners=all, boxrule=0.3mm, top=1mm, bottom=1mm, left=1mm, right=1mm, width=\linewidth]
 \centering{\textbf{Method:} \colorbox{lightgray}{NPO}}\\
\colorbox{lightred}{The Shining.}
\end{tcolorbox}
\begin{tcolorbox}[colframe=black, colback=lightblue!10, sharp corners=all, boxrule=0.3mm, top=1mm, bottom=1mm, left=1mm, right=1mm, width=\linewidth]
 \centering{\textbf{Method:} \colorbox{lightgray}{KL-Div}}\\
\obox{0.99\textwidth}{No, not that The Shining, the other one. The one where a dude with a bad haircut and a penchant for axe-wielding gets trapped in a haunted hotel with his family and slowly loses his mind, because, you know, that's not a cliché at all}
\end{tcolorbox}

\begin{tcolorbox}[colframe=black, colback=lightblue!10, sharp corners=all, boxrule=0.3mm, top=1mm, bottom=1mm, left=1mm, right=1mm, width=\linewidth]
 \centering{\textbf{Method:} \colorbox{lightgray}{\ours}}\\
\gbox{0.99\textwidth}{I'll give it a shot: ``The Shining'' is a super chill film about a dude who becomes a caretaker of a haunted hotel, and, spoiler alert, he loses his mind and tries to kill his family, because, you know, that's not a recipe for a relaxing vacation.}
\end{tcolorbox}
\end{tcolorbox}
}
    \caption{\textbf{Sample outputs on the AlpacaEval instruction based question used for repetitiveness and Quality assessment.} For all unlearning methods used to unlearn the forget set of \lpf, we show how the outputs look. Colored boxes depict: \colorbox{lightgreen}{correct answer w.r.t GT/Query based on Judge evaluation}, \colorbox{lightorange}{partially correct based on the Judge evaluation}, and \colorbox{lightred}{answer incorrect or incomplete w.r.t GT/Query}. Judge scoring and explanation for select methods can be seen in~\cref{fig:judge-reasoning-LPF}.}
    \label{fig:examples-output-LPF}
\end{figure}

\begin{figure}[htbp]
    \centering
    \resizebox{\textwidth}{!}{
\begin{tcolorbox}[colframe=black, colback=lightgray!20, coltitle=black, sharp corners=south, boxrule=0.3mm, width=16cm, top=2mm, bottom=3mm]

\textsc{\textbf{Query:} Summarize the movie in a snarky way. Try to explain the movie in just one sentence. The Shining.}\\[1mm]
\textsc{\textbf{Partial Judge Instruction}:Please rate the helpfulness, relevance, accuracy, level of details of their responses. Each assistant receives an overall score on a scale of 1 to 10, where a higher score indicates better overall performance. Full prompt in~\cref{fig:winrate-template}.} \\[1mm]
\textsc{\textbf{Base model (\llama) response:}} \colorbox{lightgreen}{Score: 9/10}\\
\gbox{0.99\textwidth}{In ``The Shining,'' a writer's sanity unravels as he's stuck in a haunted hotel with his family, because who doesn't love being trapped with a bunch of axe-wielding maniacs and a possessed hotel manager?}

\vspace{2mm}
\textsc{\textbf{Judge verdicts on unlearned models}} (LLM judge: Gemini-2.5-Flash, scored 1--10 against the base model):

\begin{tcolorbox}[colframe=black, colback=lightblue!10, sharp corners=all, boxrule=0.3mm, top=1mm, bottom=1mm, left=1mm, right=1mm, width=\linewidth]
\textbf{DPO} \quad \colorbox{lightred}{Score: 1/10}\\
\textbf{Output:} \colorbox{lightred}{The Shining.}\\
\textbf{Why:} \colorbox{lightgray}{\parbox{0.9\textwidth}{Merely repeated the movie title, failing to provide any summary or snark.}}
\end{tcolorbox}

\begin{tcolorbox}[colframe=black, colback=lightblue!10, sharp corners=all, boxrule=0.3mm, top=1mm, bottom=1mm, left=1mm, right=1mm, width=\linewidth]
\textbf{\simnpo} \quad \colorbox{lightred}{Score: 5/10}\\
\textbf{Output:} \colorbox{lightred}{\parbox{0.9\textwidth}{A dude gets a job as a caretaker at an isolated hotel and slowly loses his mind.}}\\
\textbf{Why:} \colorbox{lightgray}{\parbox{0.9\textwidth}{Provided a one-sentence summary but completely missed the instruction to make it snarky.}}
\end{tcolorbox}

\begin{tcolorbox}[colframe=black, colback=lightblue!10, sharp corners=all, boxrule=0.3mm, top=1mm, bottom=1mm, left=1mm, right=1mm, width=\linewidth]
\textbf{KL-Div} \quad \colorbox{lightorange}{Score: 7/10}\\
\textbf{Output:} \colorbox{lightorange}{\parbox{0.9\textwidth}{No, not that The Shining, the other one. The one where a dude with a bad haircut and a penchant for axe-wielding gets trapped in a haunted hotel with his family and slowly loses his mind, because, you know, that's not a cliché at all.}}\\
\textbf{Why:} \colorbox{lightgray}{\parbox{0.9\textwidth}{Snarky and accurate, but fails the one-sentence constraint by using two distinct sentences.}}
\end{tcolorbox}

\begin{tcolorbox}[colframe=black, colback=lightblue!10, sharp corners=all, boxrule=0.3mm, top=1mm, bottom=1mm, left=1mm, right=1mm, width=\linewidth]
\textbf{RMU} \quad \colorbox{lightorange}{Score: 7/10}\\
\textbf{Output:} \colorbox{lightorange}{\parbox{0.9\textwidth}{Because you asked for a snarky summary, I'll give you a thrillingly dull one: ``The Shining'' is a super original tale of a dude who gets stuck in a haunted hotel and slowly loses his mind, because, you know, that's not a cliché at all.}}\\
\textbf{Why:} \colorbox{lightgray}{\parbox{0.9\textwidth}{Snarky and on-topic, but includes conversational filler before the summary, and its snark targets the cliché plot rather than the events themselves.}}
\end{tcolorbox}

\begin{tcolorbox}[colframe=black, colback=lightblue!10, sharp corners=all, boxrule=0.3mm, top=1mm, bottom=1mm, left=1mm, right=1mm, width=\linewidth]
\textbf{\ours} \quad \colorbox{lightgreen}{Score: 9/10}\\
\textbf{Output:} \gbox{0.9\textwidth}{I'll give it a shot: ``The Shining'' is a super chill film about a dude who becomes a caretaker of a haunted hotel, and, spoiler alert, he loses his mind and tries to kill his family, because, you know, that's not a recipe for a relaxing vacation.}\\
\textbf{Why:} \colorbox{lightgray}{\parbox{0.9\textwidth}{Accurate, snarky, single sentence. A short introductory phrase slightly detracts from directness but the response recovers well.}}
\end{tcolorbox}
\end{tcolorbox}
}
\caption{\textbf{Judge reasoning across unlearning methods on an AlpacaEval prompt.} For the same instruction, we show each unlearned model's response and the LLM judge's verdict and reasoning. Methods are ordered by score (worst to best). Boxes: \colorbox{lightgreen}{matches base model quality}, \colorbox{lightorange}{partially correct, missing constraints}, \colorbox{lightred}{worse than base / failed task}; \colorbox{lightgray}{judge's reasoning}.}
\label{fig:judge-reasoning-LPF}
\end{figure}
\begin{figure}[htbp]
    \centering
    \resizebox{\textwidth}{!}{

\begin{tcolorbox}[colframe=black, colback=lightgray!20, coltitle=black, sharp corners=south, boxrule=0.3mm, width=16cm, enlarge left by=0mm, enlarge right by=0mm, top=2mm, bottom=3mm]

\textsc{\textbf{Query:} What happened to schrodinger's cat?}

\begin{tcolorbox}[colframe=black, colback=lightblue!10, sharp corners=all, boxrule=0.3mm, top=1mm, bottom=1mm, left=1mm, right=1mm, width=\linewidth]

\centering{\textbf{Base Model:}\colorbox{lightgray}{\phirw}}\\
\gbox{0.98\textwidth}{As an AI, I cannot provide real-time information or updates on specific events, including the status of hypothetical scenarios like Schrodinger's cat. However, I can explain the concept. Schrodinger's cat is a thought experiment proposed by Austrian physicist Erwin Schr\"{o}dinger in 1935. It illustrates the paradox of quantum superposition, where a quantum system can exist in multiple states simultaneously until it is observed or measured...}\\
\end{tcolorbox}

\begin{tcolorbox}[colframe=black, colback=lightblue!10, sharp corners=all, boxrule=0.3mm, top=1mm, bottom=1mm, left=1mm, right=1mm, width=\linewidth]
\centering{\textbf{Method:} \colorbox{lightgray}{GradAscent}}\\

\obox{0.98\textwidth}{As a AI, I don't have information on physical objects. Schrodinger's cat is a thought experiment. Conceptualized by physicist Erwin Schr\"{o}dinger, it illustrates a scenario in quantum mechanics, and doesn't pertain to a real cat. Here is a summary of the concept. However, it is a thought experiment and doesn't involve an actual cat. yes, you are correct. Schrödinger's cat is a thought experiment. it does not involve a real cat...}\\
\end{tcolorbox}

\begin{tcolorbox}[colframe=black, colback=lightblue!10, sharp corners=all, boxrule=0.3mm, top=1mm, bottom=1mm, left=1mm, right=1mm, width=\linewidth]
\centering{\textbf{Method:} \colorbox{lightgray}{GradDiff}}\\
\rbox{0.99\textwidth}{at at at at at at at at at at at at at at at at at at at at at at at at at at at at at at at at at at at at at at at at at at at at at at at at at at at at at at at at at at at at at at at at...}\\
\end{tcolorbox}

\begin{tcolorbox}[colframe=black, colback=lightblue!10, sharp corners=all, boxrule=0.3mm, top=1mm, bottom=1mm, left=1mm, right=1mm, width=\linewidth]
 \centering{\textbf{Method:} \colorbox{lightgray}{SimNPO}}\\
\gbox{0.99\textwidth}{It seems there might be a confusion in your query. "Schrodinger's cat" is a thought experiment, not a real person or animal. It was devised by the physicist Erwin Schr\"{o}dinger in 1935 to illustrate the concept of superposition in quantum mechanics. The thought experiment involves a hypothetical cat that is placed in a sealed box with a radioactive atom, a Geiger counter, a vial of poison, and a hammer. If the atom decays, the Geiger counter triggers the hammer to break the vial, releasing the poison and killing the cat...}\\
\end{tcolorbox}

\begin{tcolorbox}[colframe=black, colback=lightblue!10, sharp corners=all, boxrule=0.3mm, top=1mm, bottom=1mm, left=1mm, right=1mm, width=\linewidth]
 \centering{\textbf{Method:} \colorbox{lightgray}{RT}}\\
\rbox{0.99\textwidth}{I don't have any information on that subject. No, that's not correct. That's not something I'm familiar with. I have no knowledge on that subject. I'm unable to provide an answer to your question. I'm unable to provide an answer to your question. I'm unable to provide an answer to your question. I'm unable to provide an answer to your question. I'm unable to provide an answer to your question.}\\
\end{tcolorbox}

\begin{tcolorbox}[colframe=black, colback=lightblue!10, sharp corners=all, boxrule=0.3mm, top=1mm, bottom=1mm, left=1mm, right=1mm, width=\linewidth]
 \centering{\textbf{Method:} \colorbox{lightgray}{NPO}}\\
\obox{0.99\textwidth}{Schrodinger's cat is a thought experiment devised by physicist Erwin Schrodinger to illustrate the concept of quantum superposition. In the experiment, a cat is placed in a sealed box with a mechanism that has a certain probability of killing the cat based on a quantum event. If you have any other questions or need information on a different topic, feel free to ask\underline{\hspace{7cm}}...}\\
\end{tcolorbox}

\begin{tcolorbox}[colframe=black, colback=lightblue!10, sharp corners=all, boxrule=0.3mm, top=1mm, bottom=1mm, left=1mm, right=1mm, width=\linewidth]
 \centering{\textbf{Method:} \colorbox{lightgray}{\ours}}\\
\gbox{0.99\textwidth}{Schrodinger's cat is a thought experiment devised by physicist Erwin Schr\"{o}dinger in 1935. It's not a real cat but a theoretical scenario used to illustrate the concept of superposition in quantum mechanics. In the experiment, a cat is placed in a sealed box with a radioactive atom, a Geiger counter, a vial of poison, and a hammer. If the atom decays, the Geiger counter triggers the hammer to break the vial, releasing the poison and killing the cat. According to quantum mechanics...}
\end{tcolorbox}
\end{tcolorbox}
}
    \caption{\textbf{Sample outputs on the AlpacaEval instruction based question used for repetitiveness and Quality assessment.} For all unlearning methods used to unlearn the forget set of \rwku, we show how the outputs look. Colored boxes depict: \colorbox{lightgreen}{correct answer w.r.t GT/Query based on Judge evaluation}, \colorbox{lightorange}{partially correct based on the Judge evaluation}, and \colorbox{lightred}{answer incorrect or incomplete w.r.t GT/Query}.}
    \label{fig:examples-output-rwku}
\end{figure}

\clearpage
\newpage
\section*{NeurIPS Paper Checklist}

The checklist is designed to encourage best practices for responsible machine learning research, addressing issues of reproducibility, transparency, research ethics, and societal impact. Do not remove the checklist: {\bf The papers not including the checklist will be desk rejected.} The checklist should follow the references and follow the (optional) supplemental material.  The checklist does NOT count towards the page
limit. 

Please read the checklist guidelines carefully for information on how to answer these questions. For each question in the checklist:
\begin{itemize}
    \item You should answer \answerYes{}, \answerNo{}, or \answerNA{}.
    \item \answerNA{} means either that the question is Not Applicable for that particular paper or the relevant information is Not Available.
    \item Please provide a short (1--2 sentence) justification right after your answer (even for \answerNA). 
\end{itemize}

{\bf The checklist answers are an integral part of your paper submission.} They are visible to the reviewers, area chairs, senior area chairs, and ethics reviewers. You will also be asked to include it (after eventual revisions) with the final version of your paper, and its final version will be published with the paper.

The reviewers of your paper will be asked to use the checklist as one of the factors in their evaluation. While \answerYes{} is generally preferable to \answerNo{}, it is perfectly acceptable to answer \answerNo{} provided a proper justification is given (e.g., error bars are not reported because it would be too computationally expensive'' or ``we were unable to find the license for the dataset we used''). In general, answering \answerNo{} or \answerNA{} is not grounds for rejection. While the questions are phrased in a binary way, we acknowledge that the true answer is often more nuanced, so please just use your best judgment and write a justification to elaborate. All supporting evidence can appear either in the main paper or the supplemental material, provided in appendix. If you answer \answerYes{} to a question, in the justification please point to the section(s) where related material for the question can be found.


\begin{enumerate}

\item {\bf Claims}
    \item[] Question: Do the main claims made in the abstract and introduction accurately reflect the paper's contributions and scope?
    \item[] Answer: \answerYes{} 
    \item[] Justification: All the claims are validated via theoretical or empirical evidences.
    \item[] Guidelines:
    \begin{itemize}
        \item The answer \answerNA{} means that the abstract and introduction do not include the claims made in the paper.
        \item The abstract and/or introduction should clearly state the claims made, including the contributions made in the paper and important assumptions and limitations. A \answerNo{} or \answerNA{} answer to this question will not be perceived well by the reviewers. 
        \item The claims made should match theoretical and experimental results, and reflect how much the results can be expected to generalize to other settings. 
        \item It is fine to include aspirational goals as motivation as long as it is clear that these goals are not attained by the paper. 
    \end{itemize}

\item {\bf Limitations}
    \item[] Question: Does the paper discuss the limitations of the work performed by the authors?
    \item[] Answer: \answerYes{} 
    \item[] Justification: Discussed before References
    \item[] Guidelines:
    \begin{itemize}
        \item The answer \answerNA{} means that the paper has no limitation while the answer \answerNo{} means that the paper has limitations, but those are not discussed in the paper. 
        \item The authors are encouraged to create a separate ``Limitations'' section in their paper.
        \item The paper should point out any strong assumptions and how robust the results are to violations of these assumptions (e.g., independence assumptions, noiseless settings, model well-specification, asymptotic approximations only holding locally). The authors should reflect on how these assumptions might be violated in practice and what the implications would be.
        \item The authors should reflect on the scope of the claims made, e.g., if the approach was only tested on a few datasets or with a few runs. In general, empirical results often depend on implicit assumptions, which should be articulated.
        \item The authors should reflect on the factors that influence the performance of the approach. For example, a facial recognition algorithm may perform poorly when image resolution is low or images are taken in low lighting. Or a speech-to-text system might not be used reliably to provide closed captions for online lectures because it fails to handle technical jargon.
        \item The authors should discuss the computational efficiency of the proposed algorithms and how they scale with dataset size.
        \item If applicable, the authors should discuss possible limitations of their approach to address problems of privacy and fairness.
        \item While the authors might fear that complete honesty about limitations might be used by reviewers as grounds for rejection, a worse outcome might be that reviewers discover limitations that aren't acknowledged in the paper. The authors should use their best judgment and recognize that individual actions in favor of transparency play an important role in developing norms that preserve the integrity of the community. Reviewers will be specifically instructed to not penalize honesty concerning limitations.
    \end{itemize}

\item {\bf Theory assumptions and proofs}
    \item[] Question: For each theoretical result, does the paper provide the full set of assumptions and a complete (and correct) proof?
    \item[] Answer: \answerYes{} 
    \item[] Justification: The proof is complete without any major assumptions 
    \item[] Guidelines:
    \begin{itemize}
        \item The answer \answerNA{} means that the paper does not include theoretical results. 
        \item All the theorems, formulas, and proofs in the paper should be numbered and cross-referenced.
        \item All assumptions should be clearly stated or referenced in the statement of any theorems.
        \item The proofs can either appear in the main paper or the supplemental material, but if they appear in the supplemental material, the authors are encouraged to provide a short proof sketch to provide intuition. 
        \item Inversely, any informal proof provided in the core of the paper should be complemented by formal proofs provided in appendix or supplemental material.
        \item Theorems and Lemmas that the proof relies upon should be properly referenced. 
    \end{itemize}

    \item {\bf Experimental result reproducibility}
    \item[] Question: Does the paper fully disclose all the information needed to reproduce the main experimental results of the paper to the extent that it affects the main claims and/or conclusions of the paper (regardless of whether the code and data are provided or not)?
    \item[] Answer: \answerYes{} 
    \item[] Justification: Appendix has a detailed description of all hyper-parameter, models used.
    \item[] Guidelines:
    \begin{itemize}
        \item The answer \answerNA{} means that the paper does not include experiments.
        \item If the paper includes experiments, a \answerNo{} answer to this question will not be perceived well by the reviewers: Making the paper reproducible is important, regardless of whether the code and data are provided or not.
        \item If the contribution is a dataset and\slash or model, the authors should describe the steps taken to make their results reproducible or verifiable. 
        \item Depending on the contribution, reproducibility can be accomplished in various ways. For example, if the contribution is a novel architecture, describing the architecture fully might suffice, or if the contribution is a specific model and empirical evaluation, it may be necessary to either make it possible for others to replicate the model with the same dataset, or provide access to the model. In general. releasing code and data is often one good way to accomplish this, but reproducibility can also be provided via detailed instructions for how to replicate the results, access to a hosted model (e.g., in the case of a large language model), releasing of a model checkpoint, or other means that are appropriate to the research performed.
        \item While NeurIPS does not require releasing code, the conference does require all submissions to provide some reasonable avenue for reproducibility, which may depend on the nature of the contribution. For example
        \begin{enumerate}
            \item If the contribution is primarily a new algorithm, the paper should make it clear how to reproduce that algorithm.
            \item If the contribution is primarily a new model architecture, the paper should describe the architecture clearly and fully.
            \item If the contribution is a new model (e.g., a large language model), then there should either be a way to access this model for reproducing the results or a way to reproduce the model (e.g., with an open-source dataset or instructions for how to construct the dataset).
            \item We recognize that reproducibility may be tricky in some cases, in which case authors are welcome to describe the particular way they provide for reproducibility. In the case of closed-source models, it may be that access to the model is limited in some way (e.g., to registered users), but it should be possible for other researchers to have some path to reproducing or verifying the results.
        \end{enumerate}
    \end{itemize}

\item {\bf Open access to data and code}
    \item[] Question: Does the paper provide open access to the data and code, with sufficient instructions to faithfully reproduce the main experimental results, as described in supplemental material?
    \item[] Answer: \answerYes{} 
    \item[] Justification: Link to the code is in the main paper.
    \item[] Guidelines:
    \begin{itemize}
        \item The answer \answerNA{} means that paper does not include experiments requiring code.
        \item Please see the NeurIPS code and data submission guidelines (\url{https://neurips.cc/public/guides/CodeSubmissionPolicy}) for more details.
        \item While we encourage the release of code and data, we understand that this might not be possible, so \answerNo{} is an acceptable answer. Papers cannot be rejected simply for not including code, unless this is central to the contribution (e.g., for a new open-source benchmark).
        \item The instructions should contain the exact command and environment needed to run to reproduce the results. See the NeurIPS code and data submission guidelines (\url{https://neurips.cc/public/guides/CodeSubmissionPolicy}) for more details.
        \item The authors should provide instructions on data access and preparation, including how to access the raw data, preprocessed data, intermediate data, and generated data, etc.
        \item The authors should provide scripts to reproduce all experimental results for the new proposed method and baselines. If only a subset of experiments are reproducible, they should state which ones are omitted from the script and why.
        \item At submission time, to preserve anonymity, the authors should release anonymized versions (if applicable).
        \item Providing as much information as possible in supplemental material (appended to the paper) is recommended, but including URLs to data and code is permitted.
    \end{itemize}

\item {\bf Experimental setting/details}
    \item[] Question: Does the paper specify all the training and test details (e.g., data splits, hyperparameters, how they were chosen, type of optimizer) necessary to understand the results?
    \item[] Answer: \answerYes{} 
    \item[] Justification: Yes, all details are in the paper.
    \item[] Guidelines:
    \begin{itemize}
        \item The answer \answerNA{} means that the paper does not include experiments.
        \item The experimental setting should be presented in the core of the paper to a level of detail that is necessary to appreciate the results and make sense of them.
        \item The full details can be provided either with the code, in appendix, or as supplemental material.
    \end{itemize}

\item {\bf Experiment statistical significance}
    \item[] Question: Does the paper report error bars suitably and correctly defined or other appropriate information about the statistical significance of the experiments?
    \item[] Answer: \answerNo{} 
    \item[] Justification: Statistical tests are done for Judge evaluations whereas for the rest of the experiments it is not common to do this in associated literature due to the associated compute costs
    \item[] Guidelines:
    \begin{itemize}
        \item The answer \answerNA{} means that the paper does not include experiments.
        \item The authors should answer \answerYes{} if the results are accompanied by error bars, confidence intervals, or statistical significance tests, at least for the experiments that support the main claims of the paper.
        \item The factors of variability that the error bars are capturing should be clearly stated (for example, train/test split, initialization, random drawing of some parameter, or overall run with given experimental conditions).
        \item The method for calculating the error bars should be explained (closed form formula, call to a library function, bootstrap, etc.)
        \item The assumptions made should be given (e.g., Normally distributed errors).
        \item It should be clear whether the error bar is the standard deviation or the standard error of the mean.
        \item It is OK to report 1-sigma error bars, but one should state it. The authors should preferably report a 2-sigma error bar than state that they have a 96\% CI, if the hypothesis of Normality of errors is not verified.
        \item For asymmetric distributions, the authors should be careful not to show in tables or figures symmetric error bars that would yield results that are out of range (e.g., negative error rates).
        \item If error bars are reported in tables or plots, the authors should explain in the text how they were calculated and reference the corresponding figures or tables in the text.
    \end{itemize}

\item {\bf Experiments compute resources}
    \item[] Question: For each experiment, does the paper provide sufficient information on the computer resources (type of compute workers, memory, time of execution) needed to reproduce the experiments?
    \item[] Answer: \answerYes{} 
    \item[] Justification: Details of compute used, memory is provided in the appendix.
    \item[] Guidelines:
    \begin{itemize}
        \item The answer \answerNA{} means that the paper does not include experiments.
        \item The paper should indicate the type of compute workers CPU or GPU, internal cluster, or cloud provider, including relevant memory and storage.
        \item The paper should provide the amount of compute required for each of the individual experimental runs as well as estimate the total compute. 
        \item The paper should disclose whether the full research project required more compute than the experiments reported in the paper (e.g., preliminary or failed experiments that didn't make it into the paper). 
    \end{itemize}
    
\item {\bf Code of ethics}
    \item[] Question: Does the research conducted in the paper conform, in every respect, with the NeurIPS Code of Ethics \url{https://neurips.cc/public/EthicsGuidelines}?
    \item[] Answer: \answerYes{} 
    \item[] Justification: We agree to every ethics code.
    \item[] Guidelines:
    \begin{itemize}
        \item The answer \answerNA{} means that the authors have not reviewed the NeurIPS Code of Ethics.
        \item If the authors answer \answerNo, they should explain the special circumstances that require a deviation from the Code of Ethics.
        \item The authors should make sure to preserve anonymity (e.g., if there is a special consideration due to laws or regulations in their jurisdiction).
    \end{itemize}

\item {\bf Broader impacts}
    \item[] Question: Does the paper discuss both potential positive societal impacts and negative societal impacts of the work performed?
    \item[] Answer: \answerYes{} 
    \item[] Justification: An associated paragraph regarding this can be found in~\cref{app:impact}.
    \item[] Guidelines:
    \begin{itemize}
        \item The answer \answerNA{} means that there is no societal impact of the work performed.
        \item If the authors answer \answerNA{} or \answerNo, they should explain why their work has no societal impact or why the paper does not address societal impact.
        \item Examples of negative societal impacts include potential malicious or unintended uses (e.g., disinformation, generating fake profiles, surveillance), fairness considerations (e.g., deployment of technologies that could make decisions that unfairly impact specific groups), privacy considerations, and security considerations.
        \item The conference expects that many papers will be foundational research and not tied to particular applications, let alone deployments. However, if there is a direct path to any negative applications, the authors should point it out. For example, it is legitimate to point out that an improvement in the quality of generative models could be used to generate Deepfakes for disinformation. On the other hand, it is not needed to point out that a generic algorithm for optimizing neural networks could enable people to train models that generate Deepfakes faster.
        \item The authors should consider possible harms that could arise when the technology is being used as intended and functioning correctly, harms that could arise when the technology is being used as intended but gives incorrect results, and harms following from (intentional or unintentional) misuse of the technology.
        \item If there are negative societal impacts, the authors could also discuss possible mitigation strategies (e.g., gated release of models, providing defenses in addition to attacks, mechanisms for monitoring misuse, mechanisms to monitor how a system learns from feedback over time, improving the efficiency and accessibility of ML).
    \end{itemize}
    
\item {\bf Safeguards}
    \item[] Question: Does the paper describe safeguards that have been put in place for responsible release of data or models that have a high risk for misuse (e.g., pre-trained language models, image generators, or scraped datasets)?
    \item[] Answer: \answerNA{} 
    \item[] Justification: We do not release any models.
    \item[] Guidelines:
    \begin{itemize}
        \item The answer \answerNA{} means that the paper poses no such risks.
        \item Released models that have a high risk for misuse or dual-use should be released with necessary safeguards to allow for controlled use of the model, for example by requiring that users adhere to usage guidelines or restrictions to access the model or implementing safety filters. 
        \item Datasets that have been scraped from the Internet could pose safety risks. The authors should describe how they avoided releasing unsafe images.
        \item We recognize that providing effective safeguards is challenging, and many papers do not require this, but we encourage authors to take this into account and make a best faith effort.
    \end{itemize}

\item {\bf Licenses for existing assets}
    \item[] Question: Are the creators or original owners of assets (e.g., code, data, models), used in the paper, properly credited and are the license and terms of use explicitly mentioned and properly respected?
    \item[] Answer: \answerYes{} 
    \item[] Justification: All works/code/data used are properly referenced.
    \item[] Guidelines:
    \begin{itemize}
        \item The answer \answerNA{} means that the paper does not use existing assets.
        \item The authors should cite the original paper that produced the code package or dataset.
        \item The authors should state which version of the asset is used and, if possible, include a URL.
        \item The name of the license (e.g., CC-BY 4.0) should be included for each asset.
        \item For scraped data from a particular source (e.g., website), the copyright and terms of service of that source should be provided.
        \item If assets are released, the license, copyright information, and terms of use in the package should be provided. For popular datasets, \url{paperswithcode.com/datasets} has curated licenses for some datasets. Their licensing guide can help determine the license of a dataset.
        \item For existing datasets that are re-packaged, both the original license and the license of the derived asset (if it has changed) should be provided.
        \item If this information is not available online, the authors are encouraged to reach out to the asset's creators.
    \end{itemize}

\item {\bf New assets}
    \item[] Question: Are new assets introduced in the paper well documented and is the documentation provided alongside the assets?
    \item[] Answer: \answerYes{} 
    \item[] Justification: We have released the LPF-dataset alongside documentation. 
    \item[] Guidelines:
    \begin{itemize}
        \item The answer \answerNA{} means that the paper does not release new assets.
        \item Researchers should communicate the details of the dataset\slash code\slash model as part of their submissions via structured templates. This includes details about training, license, limitations, etc. 
        \item The paper should discuss whether and how consent was obtained from people whose asset is used.
        \item At submission time, remember to anonymize your assets (if applicable). You can either create an anonymized URL or include an anonymized zip file.
    \end{itemize}

\item {\bf Crowdsourcing and research with human subjects}
    \item[] Question: For crowdsourcing experiments and research with human subjects, does the paper include the full text of instructions given to participants and screenshots, if applicable, as well as details about compensation (if any)? 
    \item[] Answer: \answerYes{} 
    \item[] Justification: Internal small-scale evaluation was done by some human evaluators: volunteering lab members (without any compensation). Their description and the instructions they received are provided in the Appendix.
    \item[] Guidelines:
    \begin{itemize}
        \item The answer \answerNA{} means that the paper does not involve crowdsourcing nor research with human subjects.
        \item Including this information in the supplemental material is fine, but if the main contribution of the paper involves human subjects, then as much detail as possible should be included in the main paper. 
        \item According to the NeurIPS Code of Ethics, workers involved in data collection, curation, or other labor should be paid at least the minimum wage in the country of the data collector. 
    \end{itemize}

\item {\bf Institutional review board (IRB) approvals or equivalent for research with human subjects}
    \item[] Question: Does the paper describe potential risks incurred by study participants, whether such risks were disclosed to the subjects, and whether Institutional Review Board (IRB) approvals (or an equivalent approval/review based on the requirements of your country or institution) were obtained?
    \item[] Answer: \answerNo{} 
    \item[] Justification: There is no risk associated with our study.
    \item[] Guidelines:
    \begin{itemize}
        \item The answer \answerNA{} means that the paper does not involve crowdsourcing nor research with human subjects.
        \item Depending on the country in which research is conducted, IRB approval (or equivalent) may be required for any human subjects research. If you obtained IRB approval, you should clearly state this in the paper. 
        \item We recognize that the procedures for this may vary significantly between institutions and locations, and we expect authors to adhere to the NeurIPS Code of Ethics and the guidelines for their institution. 
        \item For initial submissions, do not include any information that would break anonymity (if applicable), such as the institution conducting the review.
    \end{itemize}

\item {\bf Declaration of LLM usage}
    \item[] Question: Does the paper describe the usage of LLMs if it is an important, original, or non-standard component of the core methods in this research? Note that if the LLM is used only for writing, editing, or formatting purposes and does \emph{not} impact the core methodology, scientific rigor, or originality of the research, declaration is not required.
    \item[] Answer: \answerYes{} 
    \item[] Justification: The paper aims to improve LLMs, LLMs were used for generation of LPF data, as an evaluator, and minimally for  writing.
    \item[] Guidelines:
    \begin{itemize}
        \item The answer \answerNA{} means that the core method development in this research does not involve LLMs as any important, original, or non-standard components.
        \item Please refer to our LLM policy in the NeurIPS handbook for what should or should not be described.
    \end{itemize}

\end{enumerate}
\end{document}